\documentclass[runningheads]{llncs}
\usepackage{graphicx}

\usepackage{tikz}
\usepackage{amsmath,amssymb} 
\usepackage{color}

\usepackage[accsupp]{axessibility}  

\usepackage[pagebackref=true,breaklinks=true,letterpaper=true,colorlinks,citecolor=blue,linkcolor=red,bookmarks=false]{hyperref}
\usepackage{siunitx}
\usepackage{amssymb}
\usepackage{amsmath}
\usepackage{tabularx}
\usepackage{booktabs}
\usepackage{algorithm}
\usepackage{algorithmic}
\usepackage[switch]{lineno}
\usepackage{multirow}

\begin{document}

\newcolumntype{Y}{>{\centering\arraybackslash}X}

\pagestyle{headings}
\mainmatter
\def\ECCVSubNumber{4891}  

\title{Spatio-Temporal Deformable Attention Network for Video Deblurring} 

\titlerunning{Spatio-Temporal Deformable Attention Network for Video Deblurring}
\author{%
Huicong Zhang\inst{1}\orcidID{0000-0001-5708-0855} \and
Haozhe Xie\inst{2}\orcidID{0000-0001-9596-5179} \and
Hongxun Yao\inst{1}\orcidID{0000-0003-3298-2574}}
\authorrunning{H. Zhang et al.}
\institute{
$^1$ Harbin Institute of Technology 
\hspace{0.1in}
$^2$ Tencent AI Lab
\url{https://vilab.hit.edu.cn/projects/stdan}}

\maketitle

\begin{abstract}
The key success factor of the video deblurring methods is to compensate for the blurry pixels of the mid-frame with the sharp pixels of the adjacent video frames. 
Therefore, mainstream methods align the adjacent frames based on the estimated optical flows and fuse the alignment frames for restoration. However, these methods sometimes generate unsatisfactory results because they rarely consider the blur levels of pixels, which may introduce blurry pixels from video frames. 
Actually, not all the pixels in the video frames are sharp and beneficial for deblurring. 
To address this problem, we propose the spatio-temporal deformable attention network (STDANet) for video delurring, which extracts the information of sharp pixels by considering the pixel-wise blur levels of the video frames.
Specifically, STDANet is an encoder-decoder network combined with the motion estimator and spatio-temporal deformable attention (STDA) module, where motion estimator predicts coarse optical flows that are used as base offsets to find the corresponding sharp pixels in STDA module.
Experimental results indicate that the proposed STDANet performs favorably against state-of-the-art methods on the GoPro, DVD, and BSD datasets.

\end{abstract}

\section{Introduction}

In the past few years, hand-held image capturing devices, such as smartphones and action cameras, have been pervasive in our daily life. 
The camera shake and high-speed movement in dynamic scenes often generate undesirable blur in the video.
The blurry video significantly reduces the visual quality and degrades performance in many subsequent vision tasks, including tracking~\cite{DBLP:conf/cvpr/MeiR08,DBLP:conf/cvpr/JinFC05}, video stabilization~\cite{DBLP:journals/pami/MatsushitaOGTS06}, and SLAM~\cite{DBLP:conf/iccv/LeeKL11}.  
Therefore, it is extremely attractive to develop an effective method to deblur videos for above mentioned human perception and high-level vision tasks.

Unlike image deblurring, video deblurring methods exploit additional information in the temporal domain. 
The key success factor of the video deblurring methods is to compensate for the blurry pixels of the mid-frame with the sharp pixels of the adjacent video frames. 
Traditional video deblurring methods~\cite{DBLP:conf/cvpr/KimL15a,DBLP:conf/iccv/BarBRS07,DBLP:conf/cvpr/DaiW08,DBLP:conf/eccv/WulffB14} often model motion blur by optical flow. 
Then those methods jointly estimate the optical flow and latent frames under the constraints by some hand-crafted priors. 


\begin{figure*}[!t]
\centering
  \includegraphics[width=\textwidth]{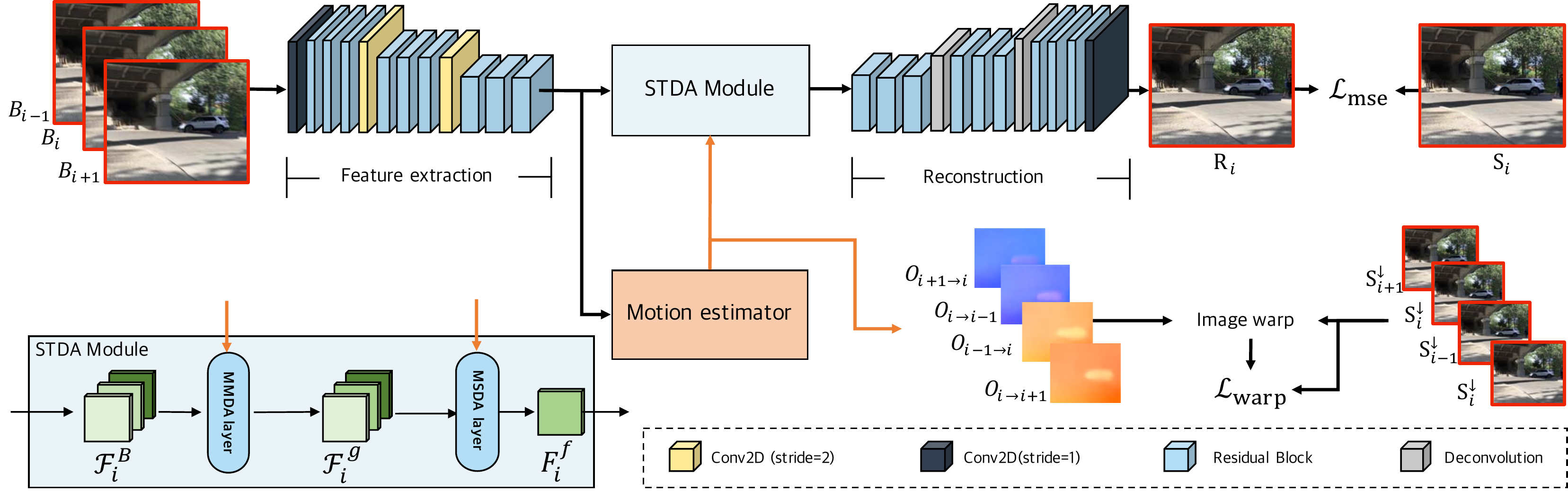} 
  \caption{The overview of STDANet, which takes three adjacent frames as input and restores the sharp mid-frame. Note that $\mathbf{S}_{i - 1}^\downarrow$, $\mathbf{S}_{i}^\downarrow$, and $\mathbf{S}_{i + 1}^\downarrow$ are the corresponding downsampled ground truth sharp frames of $\mathbf{S}_{i - 1}$, $\mathbf{S}_{i}$, and $\mathbf{S}_{i + 1}$, respectively.}
  \label{fig:overview}
\end{figure*}

Early deep learning methods~\cite{DBLP:conf/iccv/KimLSH17,DBLP:conf/cvpr/SuDWSHW17,DBLP:conf/cvpr/NahSL19,DBLP:conf/cvpr/WangCYDL19} directly concatenate the multi-frames features to restore the mid-frame based on the CNN.
However, those methods do not take full advantage of the information of the video frames because they explicitly considering the alignment of video frames.
The recent mainstream deep learning methods~\cite{DBLP:conf/cvpr/PanBT20,DBLP:conf/cvpr/LiXZ0ZRSL21} align the video frames by optical flows and directly generate the sharp frames by fusing aligned frames. 
However, they are less effective for the frames whose pixels contain large displacements because they may introduce blurry pixels that are not beneficial for blurring. 
EDVR~\cite{DBLP:conf/cvpr/WangCYDL19} computes the pixel-wise similarity in multiple frames and restores the pixels in the mid-frame with high-similarity pixels in the video frames.
However, the pixels of high similarity in the adjacent frames are also blurry for the blurry pixels in the mid-frame, which are not beneficial for deblurring. 

To solve these issues, we propose spatio-temporal deformable attention network (STDANet), 
which extracts the information of sharp pixels by considering the pixel-wise blur levels of the video frames. 
Specifically, STDANet is based on an encoder-decoder network combined with motion estimator and spatio-temporal deformable attention (STDA) module. 
First, the encoder extracts the multi-features from multiple input frames.
Then, the motion estimator predicts coarse optical flows between consecutive video frames given the multi-features generated by the encoder. 
After that, the estimated optical flows and the extracted features are fed to STDA module to generate the fused features by aggregating the information of the sharp pixels from the extracted multi-features.
Different from recent methods~\cite{DBLP:conf/cvpr/PanBT20,DBLP:conf/cvpr/LiXZ0ZRSL21}, where the optical flows are used to align the adjacent frames, the optical flows are used as base offsets in the STDA module, which reduces the degradation of deblurring results caused by inaccurate optical flows. 
Finally, the decoder restores the sharp mid-frame based on the fused features. 
The main contributions are summarized as follows:
\begin{itemize}
\item We propose a spatio-temporal deformable attention (STDA) module which aggregates the information of sharp pixels in the input consecutive video frames and eliminates the effects of blurry pixels introduced from input consecutive video frames. 
\item We present a spatio-temporal deformable attention network (STDANet) equipped with motion estimator and the proposed STDA module, where motion estimator predicts coarse optical flows and provides base offsets to find sharp pixels in adjacent frames. 
\item We quantitatively and qualitatively evaluate STDANet on the DVD, GoPro, and BSD datasets.
The experimental results indicate that STDANet performs favorably against state-of-the-art methods with comparable computational complexity. 
\end{itemize}

\section{Related Work}

\subsection{Single-Image Deblurring}

The traditional single image deblurring methods~\cite{DBLP:journals/tip/RenCPGZY16,DBLP:conf/iccp/SunCWH13,DBLP:conf/cvpr/KrishnanTF11,DBLP:conf/eccv/MichaeliI14,DBLP:conf/cvpr/LevinWDF11} assume a uniform blur kernel and design various natural image priors to compensate for the ill-posed blur removal process. 
However, these methods do not have the ability to handle the non-uniform blur. 
To solve the non-uniform blur problem, one group of methods~\cite{DBLP:conf/nips/HarmelingHS10,DBLP:conf/eccv/GuptaJZCC10,DBLP:conf/cvpr/WhyteSZP10,DBLP:conf/iccv/HirschSHS11,DBLP:conf/cvpr/XuZJ13} extends the degree of freedom of the blur model from uniform to non-uniform in a limited way compared to the dense matrix. 
Another group of methods~\cite{DBLP:conf/cvpr/Couzinie-DevySAP13,DBLP:conf/iccv/KimAL13,DBLP:conf/cvpr/KimL15a,DBLP:conf/cvpr/KimL14a} introduces additional segmentations into blur models or adopt motion estimation-based deblurs. 

With the development of deep learning, many CNN-based methods are proposed to solve dynamic scene deblurring. 
Gong~\cite{DBLP:conf/cvpr/GongYLZRSHS17} adopt a fully-convolutional deep neural network (FCN) to directly estimate the motion flow from the blurry image and restore the unblurred image from the estimated motion flow. 
Sun~\cite{DBLP:conf/cvpr/SunCXP15} use CNN to estimate the motion blur field. 
With the emergence of large datasets for single image deblurring, several works~\cite{DBLP:conf/cvpr/TaoGSWJ18,DBLP:conf/cvpr/NahKL17,DBLP:conf/cvpr/ZhangDLK19,DBLP:conf/iccv/KupynMWW19,DBLP:conf/eccv/ParkKKC20} use CNN to directly generate clear images from blurry images in an end-to-end manner. 
Nah~\cite{DBLP:conf/cvpr/NahKL17} use a multi-scale method for single image deblurring. 
However, the parameters between each scale are not shared, which leads to a huge amount of parameters.
To solve this problem, SRN~\cite{DBLP:conf/cvpr/TaoGSWJ18} introduces a deblur network with skip connections where the parameters are shared in each scale. 
DeblurGAN-v2~\cite{DBLP:conf/iccv/KupynMWW19} uses an end-to-end generative adversarial network (GAN) for single image motion deblurring and introduces the Feature Pyramid Network into single image deblurring. 
DMPHN~\cite{DBLP:conf/cvpr/ZhangDLK19} introduces the hierarchical multi-patch (MP) model for deblurring and improves deblur performance. 
MT-RNN~\cite{DBLP:conf/eccv/ParkKKC20} uses an RNN with recursive feature maps for progressive deblurring over iterations. 

\subsection{Multi-Image Deblurring}
Several methods utilize multiple images to solve dynamic scene deblurring from videos. 
The traditional methods~\cite{DBLP:conf/cvpr/KimL15a,DBLP:conf/iccv/BarBRS07,DBLP:conf/cvpr/DaiW08,DBLP:conf/eccv/WulffB14} jointly estimate the optical flow and blur kernel to restored frames with the some hand-crafted priors.  
However, the proposed priors usually lead to complex energy functions which are difficult to solve. 
In addition, Su~\cite{DBLP:conf/cvpr/SuDWSHW17} align the consecutive frames and then the Convolutional Neural Networks are used to restored images. 
Kim~\cite{DBLP:conf/iccv/KimLSH17} propose a recurrent neural network to fuse the concatenation of the multi-frames features. 
Wieschollek~\cite{DBLP:conf/iccv/WieschollekHSL17} develop a recurrent network to recurrently use the features from the previous frame in multiple scales. 
Wang~\cite{DBLP:conf/cvpr/WangCYDL19} achieve better alignment performance base on deformable convolution.  
Zhou~\cite{DBLP:conf/iccv/ZhouZPZXR19} use the dynamic filters to align the consecutive frames.
Pan~\cite{DBLP:conf/cvpr/PanBT20} introduce a temporal sharpness prior to improve the ability of the deblur network.
Zhang~\cite{DBLP:journals/tip/ZhangLZMLL19} develop a adversarial loss and spatial-temporal 3D convolutions to improve latent frame restoration. 
Recently, ARVo\cite{DBLP:conf/cvpr/LiXZ0ZRSL21} uses self-attention to capture the pixel correlation of the consecutive frames. 
However, those methods rarely consider the different blur levels of each frame, which make they do not take full advantage of the sharpness pixel information in the video frames.

\section{The Proposed Method}

The proposed STDANet aims to restore the sharp mid-frame $\mathbf{R}_i$ given three consecutive blurry frames $\mathcal{B}_i = \left\{\mathbf{B}_k \right\}_{k=i-1}^{i+1}$.
As shown in Figure~\ref{fig:overview}, it contains four components: the feature extraction network, the motion estimator, the STDA module, and the reconstruction network, where the feature extraction network and the reconstruction network follow the encoder-decoder architecture.
First, the feature extraction network generates the extracted features $\mathcal{F}_i^b = \left\{ \mathbf{F}_{k}^b \right\}_{k=i-1}^{i+1}$ for $\mathcal{B}_i$.
Then, the motion estimator predicts 
the optical flows $\mathcal{O}_i = \left\{\mathbf{O}_{k \rightarrow k + 1} | k = i - 1, i \right\} \cup \left\{\mathbf{O}_{k + 1 \rightarrow k} | k = i - 1, i \right\}$
between the two adjacent frames $\mathbf{B}_k$ and $\mathbf{B}_{k + 1}$. 
Next, the STDA module takes $\mathcal{F}_i^b$, $\mathcal{O}_i$ as input and generates the fused features $\mathbf{F}_i^f$ by aggregating the features of low-blur-level pixels in the consecutive frames.
Finally, the reconstruction network restores the sharp frame $\mathbf{R}_{i}$ for $\mathbf{B}_i$. 
Except STDANet, we also propose STDANet-Stack, which uses a cascaded strategy~\cite{DBLP:conf/cvpr/PanBT20} to stack STDANet and takes five adjacent blurry frames $\left\{\mathbf{B}_k \right\}_{k=i-2}^{i+2}$ as input.

\subsection{Motion Estimator}
\label{sec:motion_estimator}
Previous video deblurring methods~\cite{DBLP:conf/cvpr/PanBT20,DBLP:conf/cvpr/LiXZ0ZRSL21} that use optical flows to align two adjacent frames to the mid-frame, which requires accurate optical flows generated by heavyweight neural networks such as PWC-Net~\cite{DBLP:conf/cvpr/SunY0K18}.
In contrast, optical flows are used as the base offsets in the STDA module, which are more robust to the errors in estimated optical flows.
Therefore, we propose the motion estimator that predicts coarse optical flows between two adjacent frames with much smaller computational complexity.
To accelerate the computational complexity, the motion estimator generates the optical flows that are of $1/4$ sizes the input images. 
Consequently, the motion estimator is $1/70$ the size of PWC-Net. 
Compared to existing methods for optical flow estimation~\cite{DBLP:conf/iccv/DosovitskiyFIHH15,DBLP:conf/cvpr/SunY0K18,DBLP:conf/cvpr/XuRK17}, the motion estimator does not use any time-consuming layers such as correlation layer~\cite{DBLP:conf/iccv/DosovitskiyFIHH15}, cost volume layer~\cite{DBLP:conf/cvpr/SunY0K18,DBLP:conf/cvpr/XuRK17}.  

\begin{figure*}[!t]
\centering
  \includegraphics[width=0.8\textwidth]{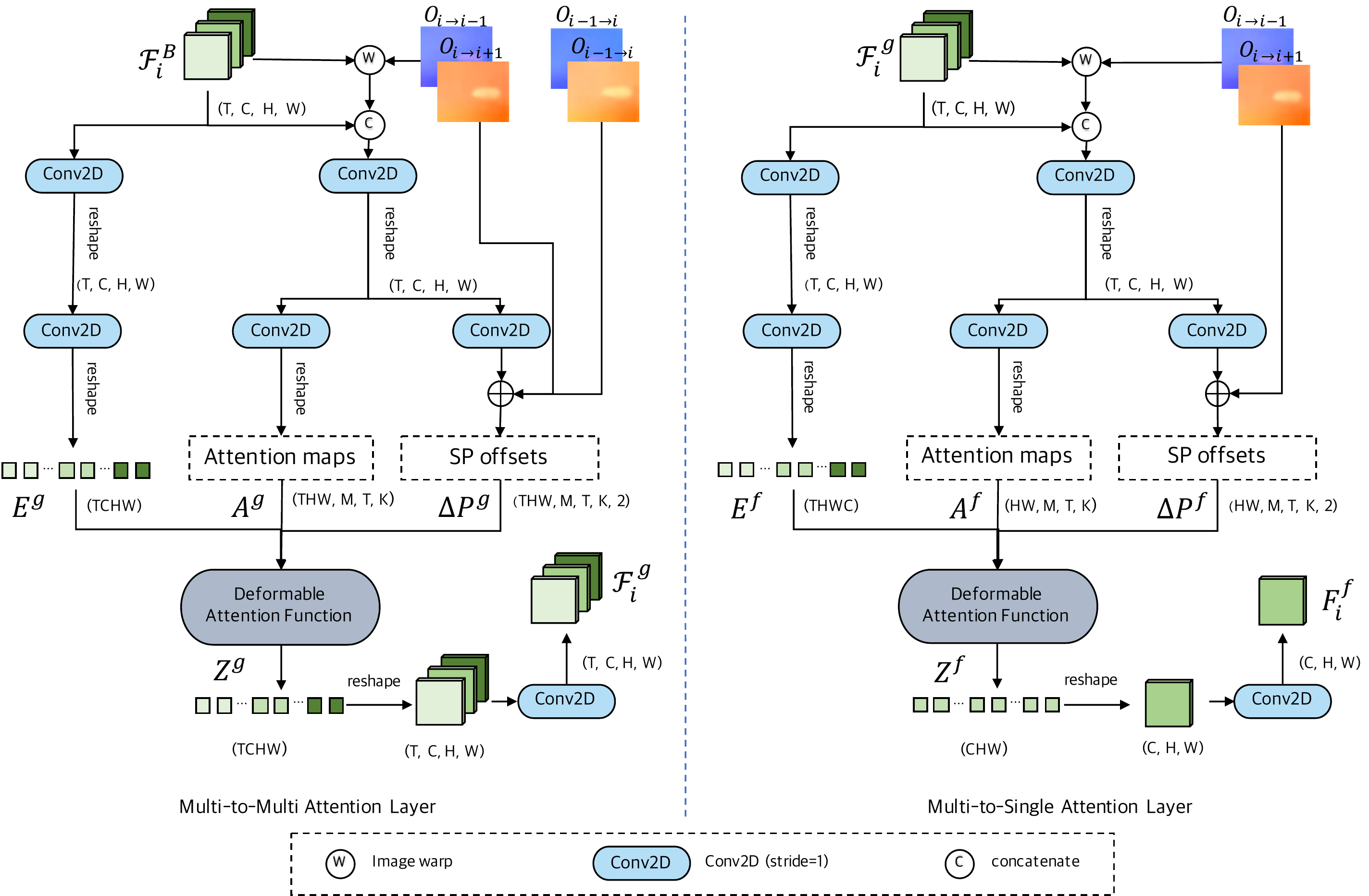} 
  \caption{The detailed network structure of the MMA and MSA layers. Note that ``SP Offsets'' denotes ``the offsets of sampling points''.}
\label{fig:MMMS}
\end{figure*} 

\begin{figure*}[!t]
  \centering
  \includegraphics[width=\textwidth]{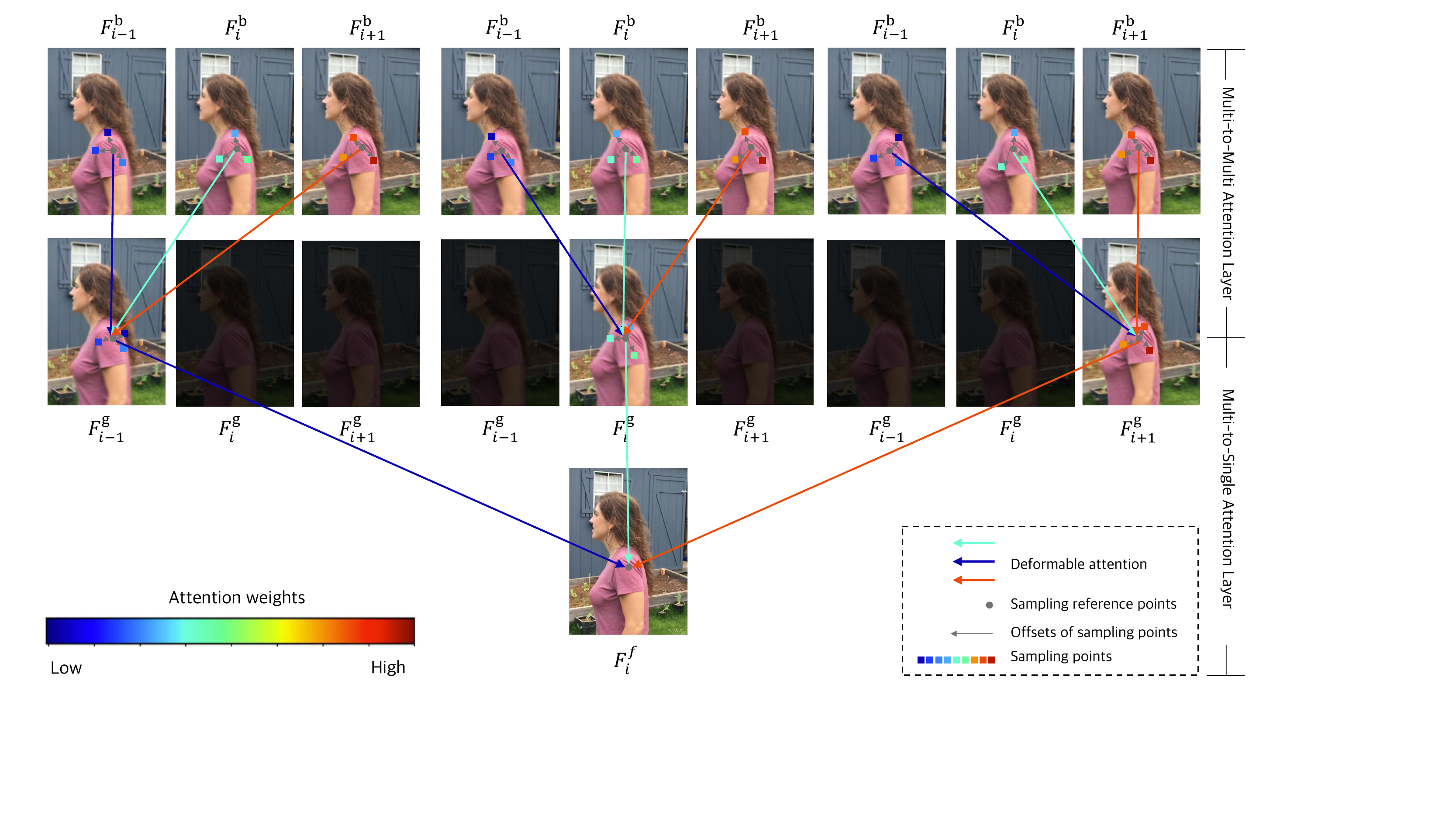} 
  \caption{The illustration of MMA and MSA layers.
  The colors of the sampling points denotes the corresponding attention weights, where higher attention weights indicate that the sampling points are sharper.
  First, the MMA layer extracts the information of sharp pixels from multi-features $\mathcal{F}^b_i$ and generates the features of adjacent frames $\mathcal{F}_i^g = \left\{ \mathbf{F}^{g}_{k} | \mathbf{F}^{g}_{k} \right\}_{k=i - 1}^{i + 1}$. 
  Second, the MSA layer generates the fused features $\mathbf{F}_i^f$ by aggregating the information of sharp pixels from $\mathcal{F}^g_i$.}
  \label{fig:MMMSoverview}
\end{figure*}

Specifically, the motion estimator consists of stacked four convolutional layers with kernel sizes of $3$ and strides of $1$. 
Given the three adjacent image features $\mathcal{F}_i^b$, the motion estimator generates four optical flows $\mathcal{O}_i = \left\{\mathbf{O}_{k \rightarrow k + 1} | k = i - 1, i \right\} \cup \left\{\mathbf{O}_{k + 1 \rightarrow k} | k = i - 1, i \right\}$, where $\mathbf{O}_{m \rightarrow n}$ represents the optical flow from the $m$-th frame to the $n$-th frame.

\subsection{Spatio-temporal Deformable Attention Module}
To extract the information of sharp pixels from consecutive video frames, we propose spatio-temporal deformable attention (STDA) module.
As shown in Figure~\ref{fig:MMMS}, there are two layers in the STDA module that aggregates features in a coarse-to-fine manner, named Multi-to-Multi attention (MMA) layer and Multi-to-Single attention (MSA) layer. 
Figure~\ref{fig:MMMSoverview} gives an illustration how the MMA and MSA layers extract image features of sharp pixels.

\subsubsection{Multi-to-Multi Attention Layer}

The multi-to-multi attention layer takes the image features of three consecutive frames $\mathcal{F}^b_i$ as input and 
generates the coarse aggregated image features $\mathcal{F}_i^g = \left\{ \mathbf{F}^{g}_{k} | \mathbf{F}^{g}_{k} \in \mathbb{R}^{C \times H \times W} \right\}_{k=i - 1}^{i + 1}$,
where $C$, $H$, and $W$ represent the number of channels, height, and width of the image features, respectively.

\textbf{Step 1.}
The image features $\mathcal{F}^b_i = \left\{ \mathbf{F}_{k}^b | \mathbf{F}_{k}^b \in \mathbb{R}^{C \times H \times W} \right\}_{k=i - 1}^{i + 1}$ of adjacent frames are aligned to the mid-frame with the estimated optical flows $\mathcal{O}_i$ and produces $\mathcal{F}^w_i = \left\{\mathbf{F}^{w}_{k} | \mathbf{F}_{k}^w \in \mathbb{R}^{C \times H \times W} \right\}$,
where $\mathbf{F}^{w}_{k}$ is
\begin{equation}
  \mathbf{F}^{w}_{k} = {\rm Warp}(\mathbf{F}^{b}_{k},\mathbf{O}_{i \rightarrow k}), k = i -1, i + 1
  \label{eq:warp}
\end{equation}
where ``Warp'' denotes the backward warp with operation. 

\textbf{Step 2.}
The concatenated features $\mathbf{F}_i^c \in \mathbb{R}^{(2T - 1) \times C \times H \times W}$ are generated by concatenating the aligned features $\mathcal{F}^w_i$ and image features $\mathcal{F}^b_i$, where $T$ denotes the number of frames in the sliding window. 

\textbf{Step 3.} 
Given $\mathbf{F}^c_i$ and $\mathcal{F}_i^b$ as input, the attention maps $\mathbf{A}^g \in \mathbb{R}^{Q \times M \times T \times K}$, the offsets of sampling points $\Delta \mathbf{P}^g \in \mathbb{R}^{Q \times M \times T \times K \times 2}$, and the flatten features $\mathbf{E}^g \in \mathbb{R}^{ THWC}$ are generated, 
where $Q = THW$.
$M$, $T$, and $K$ represent the number of attention heads, the number of sampling points, and the number of frames, respectively. 
The attention maps $\mathbf{A}^g$ are used to measure the sharpness of the pixels, which are normalized by $\sum_{t=1}^{T} \sum_{k=1}^{K} \mathbf{A}_{mtqk}^g = 1$.
The offsets of sampling points $\Delta \mathbf{P}^g$ and estimated optical flows $\mathcal{O}_i$ are used to find the corresponding pixels in the adjacent frames, where $\mathcal{O}_i$ provides the base offsets.
$\Delta \mathbf{P}^g$, $\mathbf{A}^g$, and $\mathcal{O}_i$ are generated as following

\begin{align}
\Delta \mathbf{P}^g &= \mathcal{C}_{\rm MMA}^{o}(\mathbf{F}_i^{c}) \nonumber \\
\mathbf{A}^g &= \mathcal{C}_{\rm MMA}^{m}(\mathbf{F}_i^{c}) \nonumber \\
\mathbf{E}^g &= {\rm Concat}(\mathcal{C}_{\rm MMA}^{l}(\mathbf{F}_{i-1}^b),\mathcal{C}_{\rm MMA}^{l}(\mathbf{F}_i^b),\mathcal{C}_{\rm MMA}^{l}(\mathbf{F}_{i+1}^b))
\label{eq:conv}
\end{align}
where ``Concat'' denotes the concatenation operation. 
$\mathcal{C}_{\rm MMA}^{m}, \mathcal{C}_{\rm MMA}^{o}, \mathcal{C}_{\rm MMA}^{l}$ represent different convolution layers. 
The attention map $\mathbf{A}^g $, offsets of sampling points $\Delta \mathbf{P}^g$ and flatten features $\mathbf{E}^g$ are fed to the deformable attention function $\mathcal{D}$~\cite{DBLP:conf/iclr/ZhuSLLWD21} and produces the fused features $\mathbf{Z}^g \in \mathbb{R}^{TCHW}$.

\begin{equation}
\mathbf{Z}^g = \mathcal{D}(\mathbf{A}^g, \phi( \Delta \mathbf{P}^g, \mathcal{O}_i), \mathbf{E}^g),
\label{eq:defatn}
\end{equation}
where $\phi$ represents the operation that adds the estimated optical flows to the offsets of sampling points $\Delta \mathbf{P}^g$. 
In ~\ref{eq:defatn}, the optical flows is used as based offsets, which reduces the degradation of deblurring results caused by inaccurate optical flows. 

\textbf{Step 4.}
$\mathbf{Z}^g$ is reshaped and splitted into $\left\{ \mathbf{F}^{h}_{k} | \mathbf{F}^{h}_{k} \in \mathbb{R}^{C \times H \times W} \right\}_{k = i - 1}^{i + 1}$. 
The final fused features $\mathcal{F}_i^g = \left\{ \mathbf{F}^{g}_{k} | \mathbf{F}^{g}_{k} \in \mathbb{R}^{C \times H \times W} \right\}_{k=i - 1}^{i + 1}$ are generated as following

\begin{equation}
  \mathbf{F}_k^g = \mathcal{C}_{\rm MMA}^{g}(\mathbf{F}^h_k)
  \label{eq:conv_out}
\end{equation}
where $\mathcal{C}_{\rm MMA}^{g}$ denotes a convolutional layer. 

\subsubsection{Multi-to-Single Attention Layer}

The multi-to-single attention layer takes the coarse aggregated image features $\mathcal{F}_i^g$ as input and generates the fused features $\mathbf{F}_i^f$ for the mid-frame.
Similar to the MMA layer, the MSA layer aggregates information of sharp pixels from the adjacent frames.
However, in the MSA layer, the aggregated features are only propagated to the mid-frame.
Therefore, in the MSA layer, the fused features $\mathbf{Z}^f \in \mathbb{R}^{CHW}$ is generated as following

\begin{equation}
    \mathbf{Z}^f = \mathcal{D}(\mathbf{A}^f, \phi(\Delta \mathbf{P}^f,\left\{\mathbf{O}_{k \rightarrow i} | k = i - 1, i + 1 \right\}), \mathbf{E}^f)
\end{equation}
where $\mathbf{A}^f \in \mathbb{R}^{HW \times M \times T \times K}$, $\Delta \mathbf{P}^f \in \mathbb{R}^{HW \times M \times T \times K \times 2}$, and $\mathbf{E}^f \in \mathbb{R}^{TCHW}$ are the attention maps, the offsets of sampling points, and flatten features obtained as in the MMA layer.
The fused features $\mathbf{F}_i^f$ is obtained as following

\begin{equation}
    \mathbf{F}_i^f = \mathcal{C}_{\rm MSA}^{f}(\mathbf{F}^n_i)
\end{equation}
where $\mathcal{C}_{\rm MSA}^{f}$ denotes a convolutional layer.
$\mathbf{F}^n_i \in \mathbb{R}^{C \times H \times W}$ is reshaped from $\mathbf{Z}^f$.

\subsection{Feature Extraction and Reconstruction Networks}

\noindent \textbf{Feature Extraction Network.}
The feature extraction network generates image features $\mathcal{F}^b_i$ from blurry images $\mathcal{B}_i$.
It consists of three convolutional blocks, two of which have a convolution layer with the stride of $2$ followed by three residual blocks with LeakyReLU as the activation function. 
The first convolutional block has a convolution layer with the stride of $1$ followed by three residual blocks with LeakyReLU as the activation function. 

\noindent \textbf{Reconstruction Network.}
The reconstruction network is used to restore the sharp mid-frame $\mathbf{R}_i$ by taking the fused features from STDA module as input. 
It consists of three convolutional blocks, two of which have one deconvolutional layer with the stride of $2$ and three residual blocks with LeakyReLU as the activation function. 
The last convolutional block has one convolutional layer with the stride of $1$ and three residual blocks with LeakyReLU as the activation function.

\subsection{Cascaded Progressive Deblurring} 
\label{sec:cpl}

Inspired by TSP~\cite{DBLP:conf/cvpr/PanBT20}, we propose STDANet-Stack by stacking STDANet in a cascaded manner~\cite{DBLP:conf/cvpr/PanBT20}.
It takes five adjacent blurry video frames $\left\{\mathbf{B}_k \right\}_{k=i-2}^{i+2}$ as input and restores the sharp mid-frame $\mathbf{R}_i$. 

Specifically, STDANet-Stack restores $\mathbf{R}_i$ in two steps.
First, it produces $\mathbf{\hat{R}}_{i - 1}$ by taking $\left\{\mathbf{B}_k\right\}_{k=i - 2}^i$ as input.
Similarly, $\mathbf{\hat{R}}_{i}$ and $\mathbf{\hat{R}}_{i + 1}$ are restored by taking $\left\{\mathbf{B}_k\right\}_{k=i - 1}^{i + 1}$ and $\left\{\mathbf{B}_k\right\}_{k=i}^{i + 2}$ as inputs, respectively.
Next, $\mathbf{R}_i$ is generated by taking $\left\{\mathbf{\hat{R}}_k\right\}_{k=i - 1}^{i + 1}$ as input.

\subsection{Loss Functions} 
\label{sec:lf}

We employ two loss functions to train STDANet and STDFANet-Stack.

\noindent \textbf{MSE Loss}
represents the distance between the restored frame $R$ and its corresponding ground truth sharp frame $S$, which is formulated as

\begin{equation}
\mathcal{L}_{mse} = \parallel \mathbf{R} - \mathbf{S} \parallel^{2}
\label{eq:mse}
\end{equation}

\noindent \textbf{Warp Loss}
is introduced to train the motion estimator in an unsupervised manner, which is computed as

\begin{equation}
\mathcal{L}_{warp} = \parallel \mathbf{S}_i^\downarrow - {\rm Warp}(\mathbf{S}_j^\downarrow, \mathbf{O}_{i \rightarrow j}) \parallel^{2} \label{eq:warp}
\end{equation}
where $\mathbf{S}_i^\downarrow$ and $\mathbf{S}_j^\downarrow$ are the two downsampled frames.
$\mathbf{O}_{i \rightarrow j}$ represents the estimated optical flow from $\mathbf{S}_j^\downarrow$ and $\mathbf{S}_i^\downarrow$.
``Warp'' denotes the backward warp operation.

\noindent \textbf{Total Loss}
are defined as

\begin{equation}
\mathcal{L}_{total} = \mathcal{L}_{mse} + \gamma \mathcal{L}_{warp}
\label{eq:totalloss}
\end{equation}
where $\gamma$ controls the weights of the two loss functions.

\section{Experiments}
\label{sec:Exp}

\subsection{Datasets}

\noindent \textbf{DVD}.
The DVD dataset~\cite{DBLP:conf/cvpr/SuDWSHW17} contains 71 videos (6,708 blurry-sharp pairs), splitting into 61 training videos (5,708 pairs) and 10 testing videos (1,000 pairs). 

\noindent \textbf{GoPro}.
The GoPro dataset~\cite{DBLP:conf/cvpr/NahKL17} contains 3,214 pairs of blurry images and sharp images at 1280$\times$720 resolution, where 2,103 and 1,111 pairs of blurry images and sharp images are used for training and testing, respectively.

\noindent \textbf{BSD}.
The BSD dataset~\cite{DBLP:conf/eccv/ZhongGZZ20} is a real-world video deblur dataset, which contains three sub-datasets with different sharp exposure time - blurry exposure time.

\subsection{Evaluation Metrics}

\begin{table*}[!t]
\centering
\caption{The quantitative results on the DVD dataset. Note that ``Ours$^*$'' denotes STDANet-Stack.}
    \begin{tabularx}{\linewidth}{XYcYYYYYYY}
		\toprule
		Method
		& SRN
		& IFI-RNN-L
		& STFAN
		& EDVR
		& TSP
		& PVDNet
		& ARVo
		& Ours
		& \textbf{Ours$^{*}$} \\
		\midrule
		PSNR     & 30.53      & 31.67      & 31.24      & 31.82      & 32.13     & 32.31  & 32.80 &  32.63  & \bf{33.05} \\
		SSIM      & 0.8940      & 0.9160      & 0.9340      & 0.9160    & 0.9270     & 0.9260   & 0.9352  &  0.9300 & \bf{0.9374}  \\
		\bottomrule
	\end{tabularx}
\label{tab:psnr_dvd}
\end{table*}

\begin{figure*}[!t]
    \renewcommand{\tabcolsep}{0.5pt}
    \renewcommand{\arraystretch}{1}
    \resizebox{\linewidth}{!}{
    	\begin{tabular}{cccccc}
    		\includegraphics[width=0.16\linewidth]{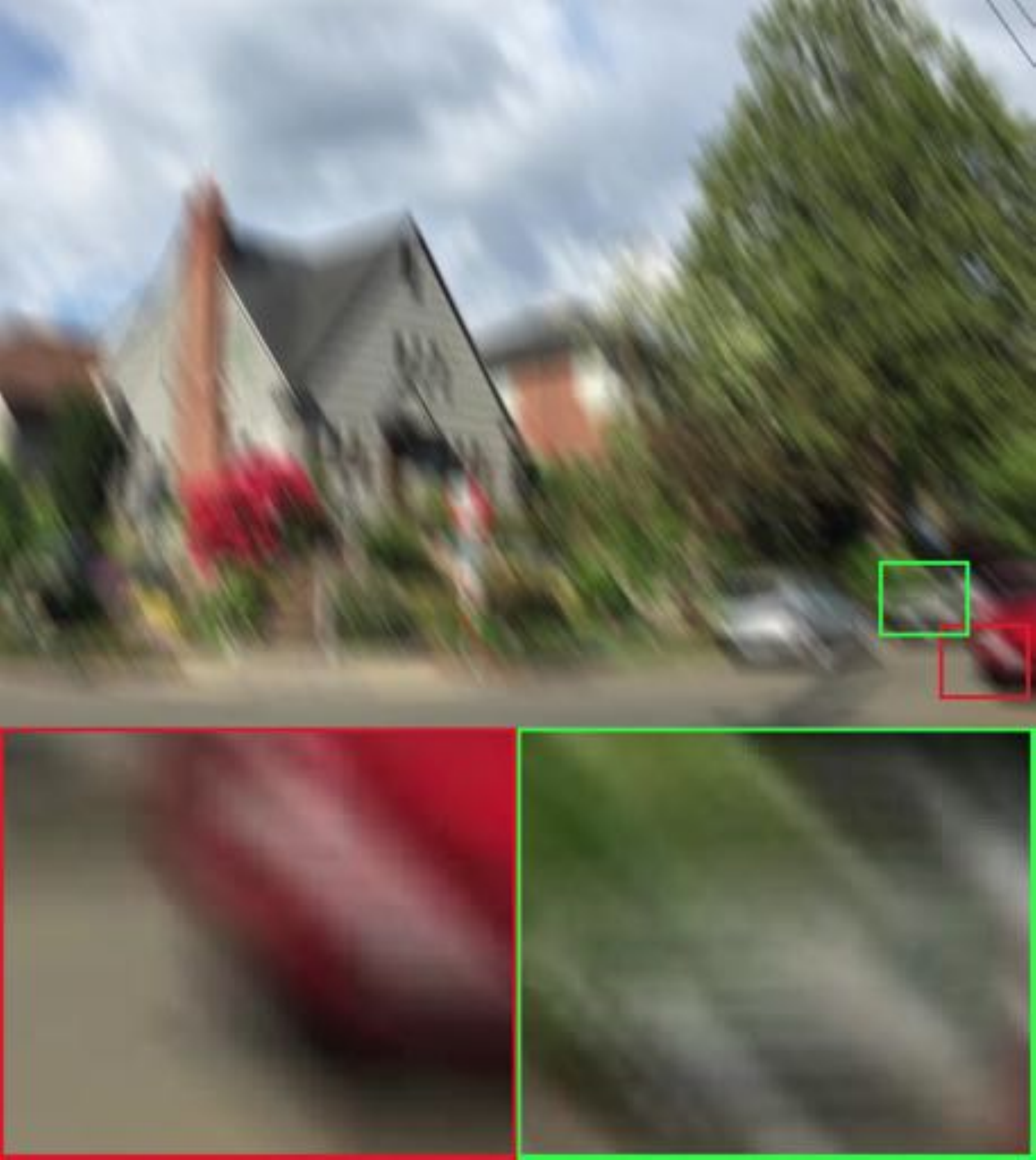} &
    		\includegraphics[width=0.16\linewidth]{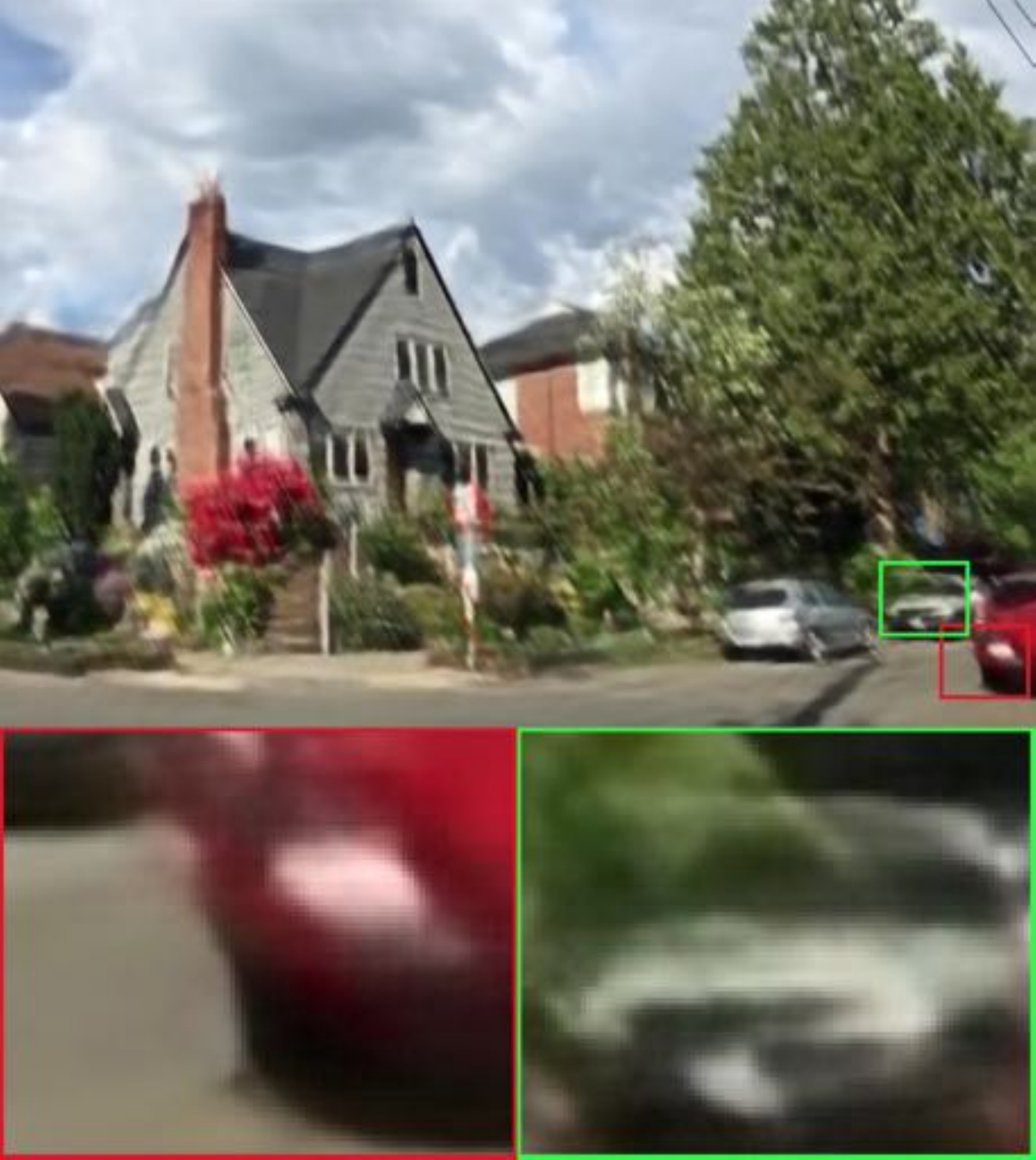} &
    		\includegraphics[width=0.16\linewidth]{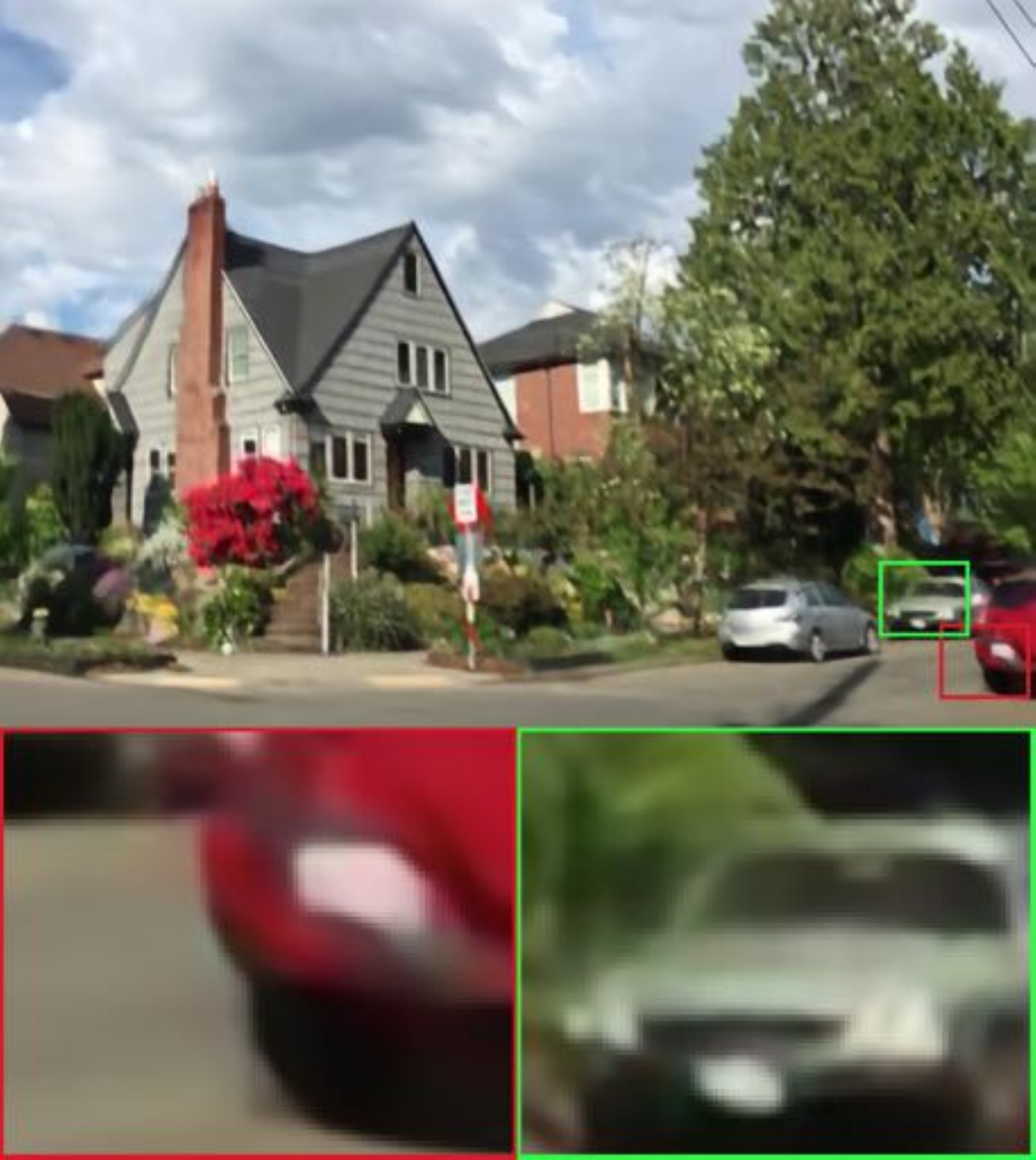} &
    		\includegraphics[width=0.16\linewidth]{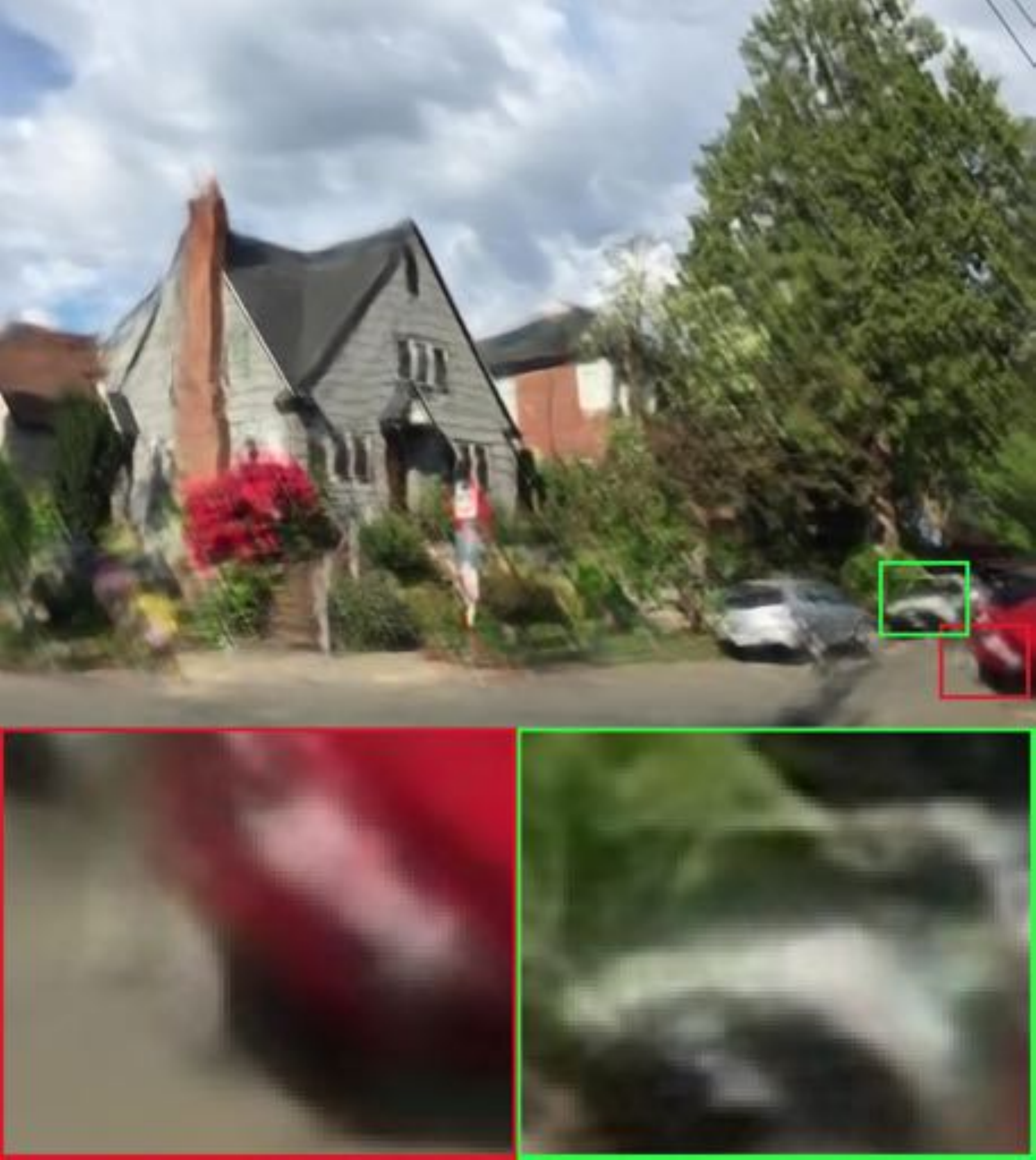} &	
    		\includegraphics[width=0.16\linewidth]{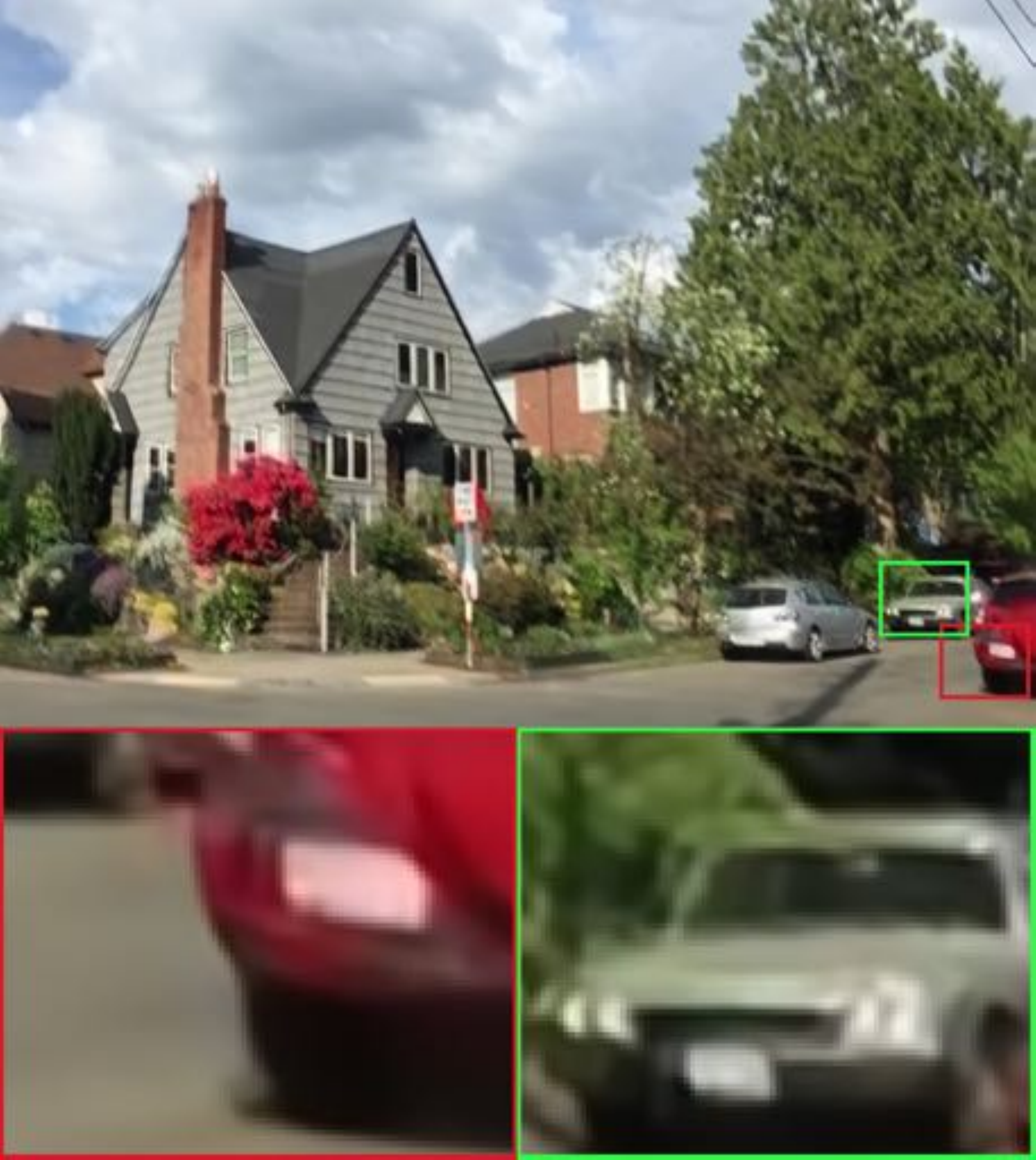} &
    		\includegraphics[width=0.16\linewidth]{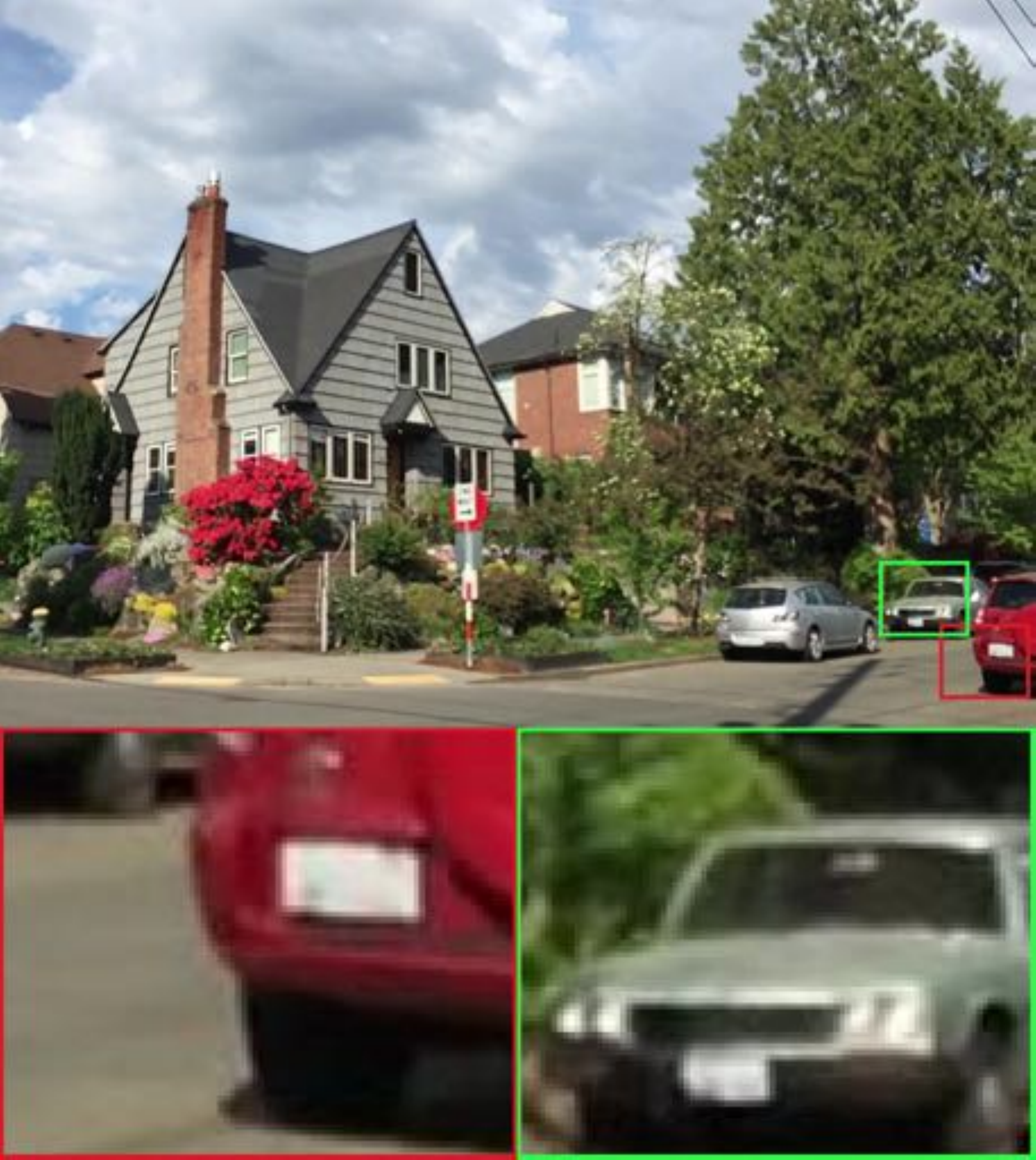} 
    		%
    		\\
    		\includegraphics[width=0.16\linewidth]{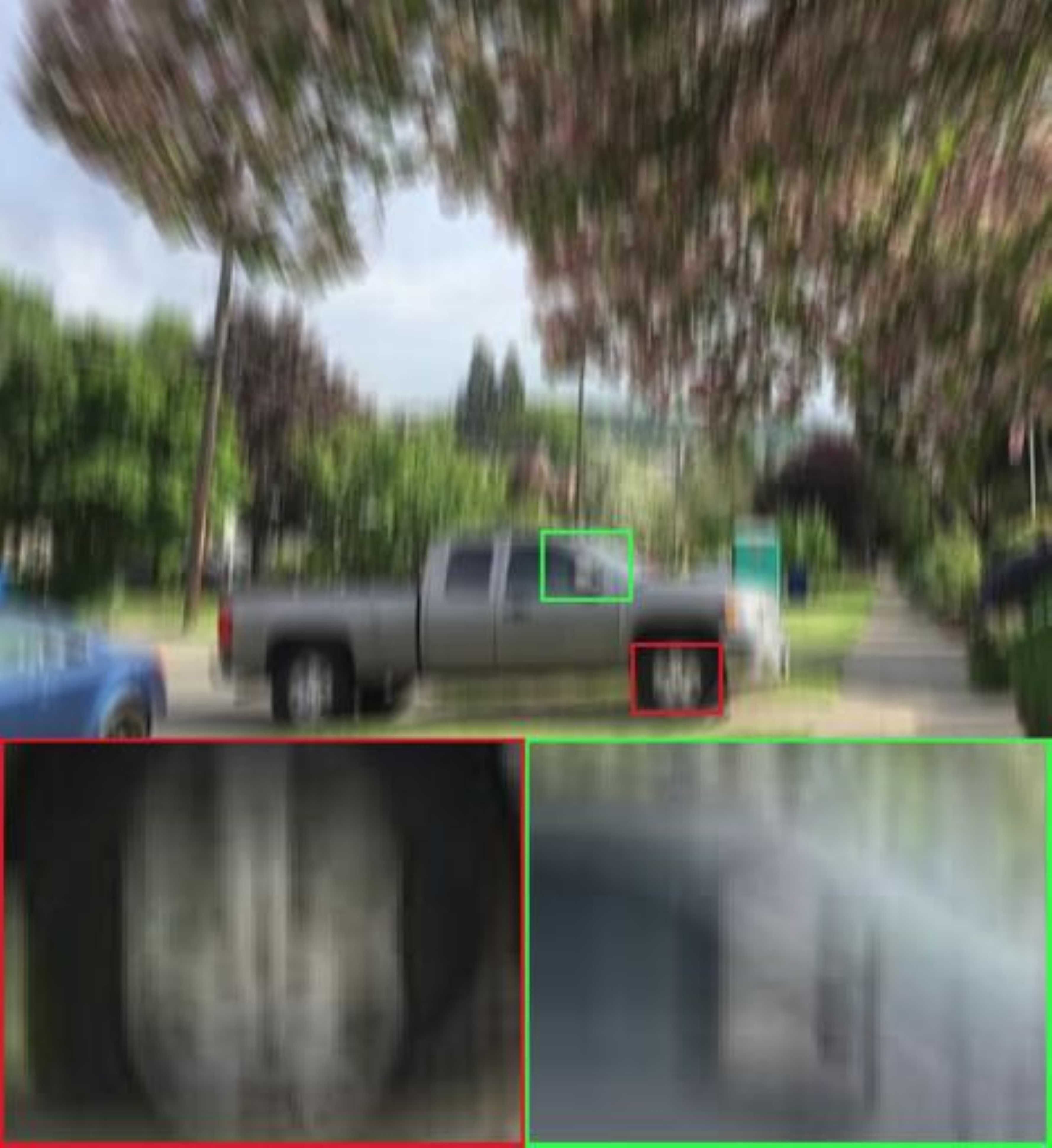} &
    		\includegraphics[width=0.16\linewidth]{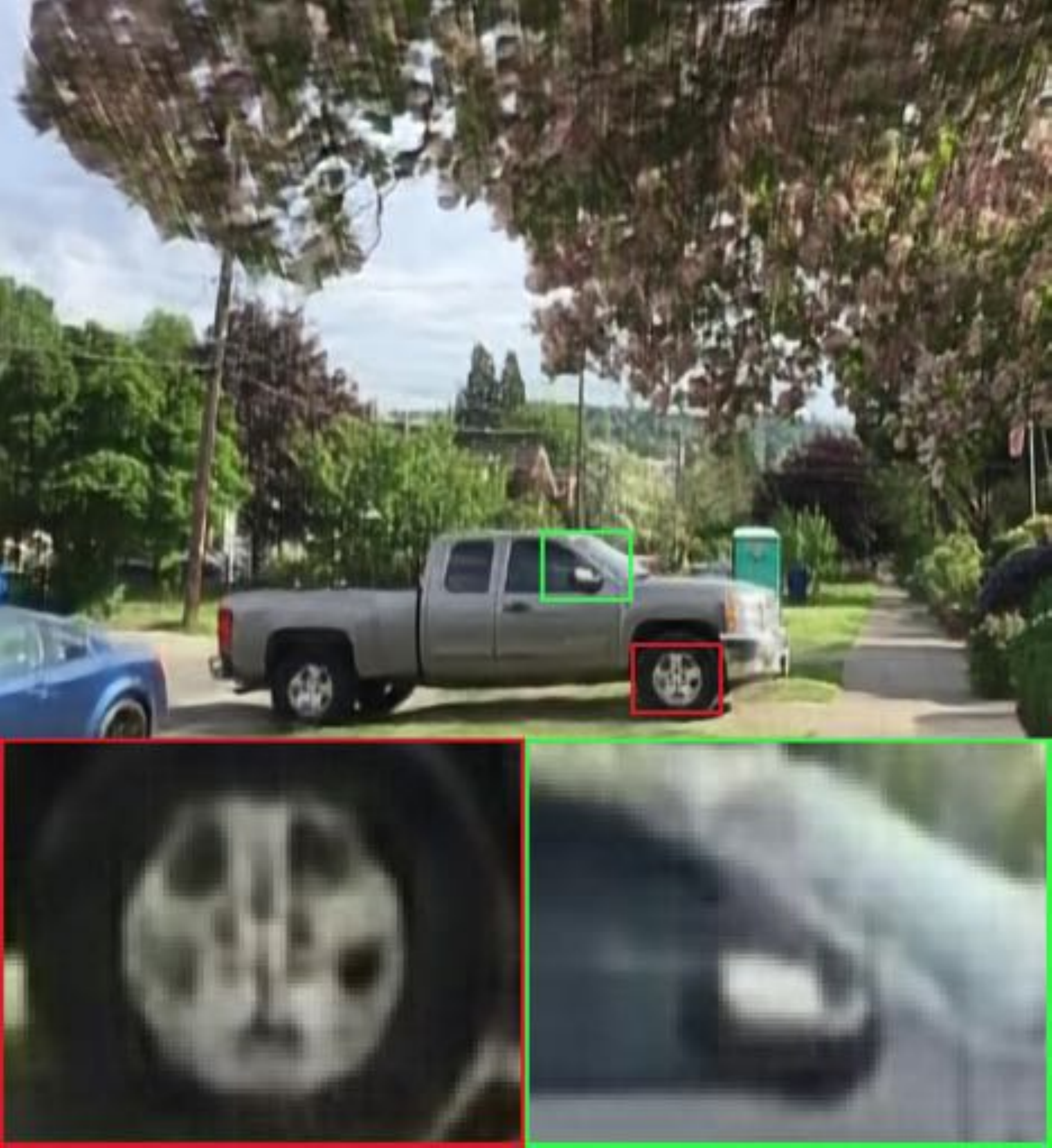} &
    		\includegraphics[width=0.16\linewidth]{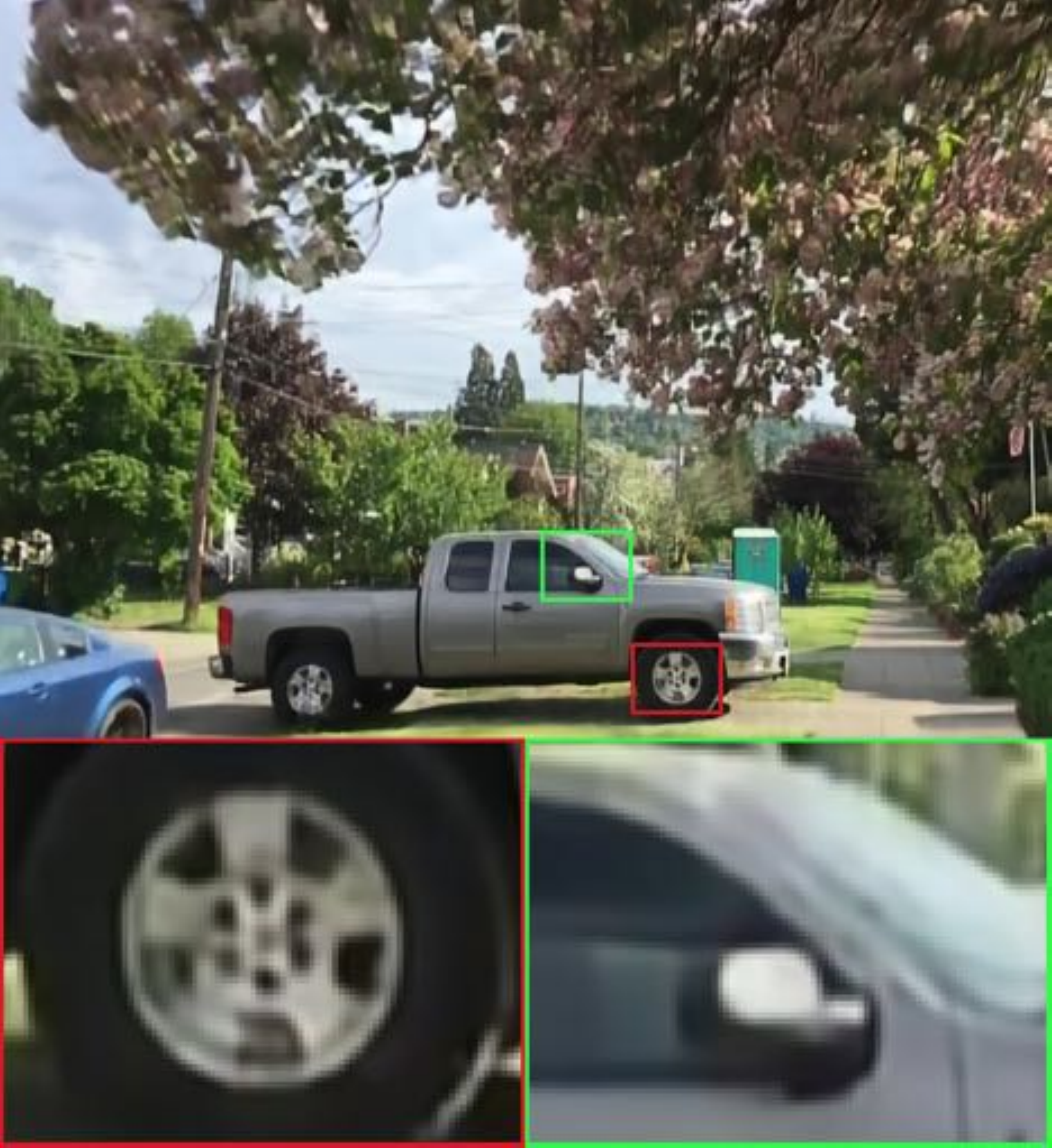} &
    		\includegraphics[width=0.16\linewidth]{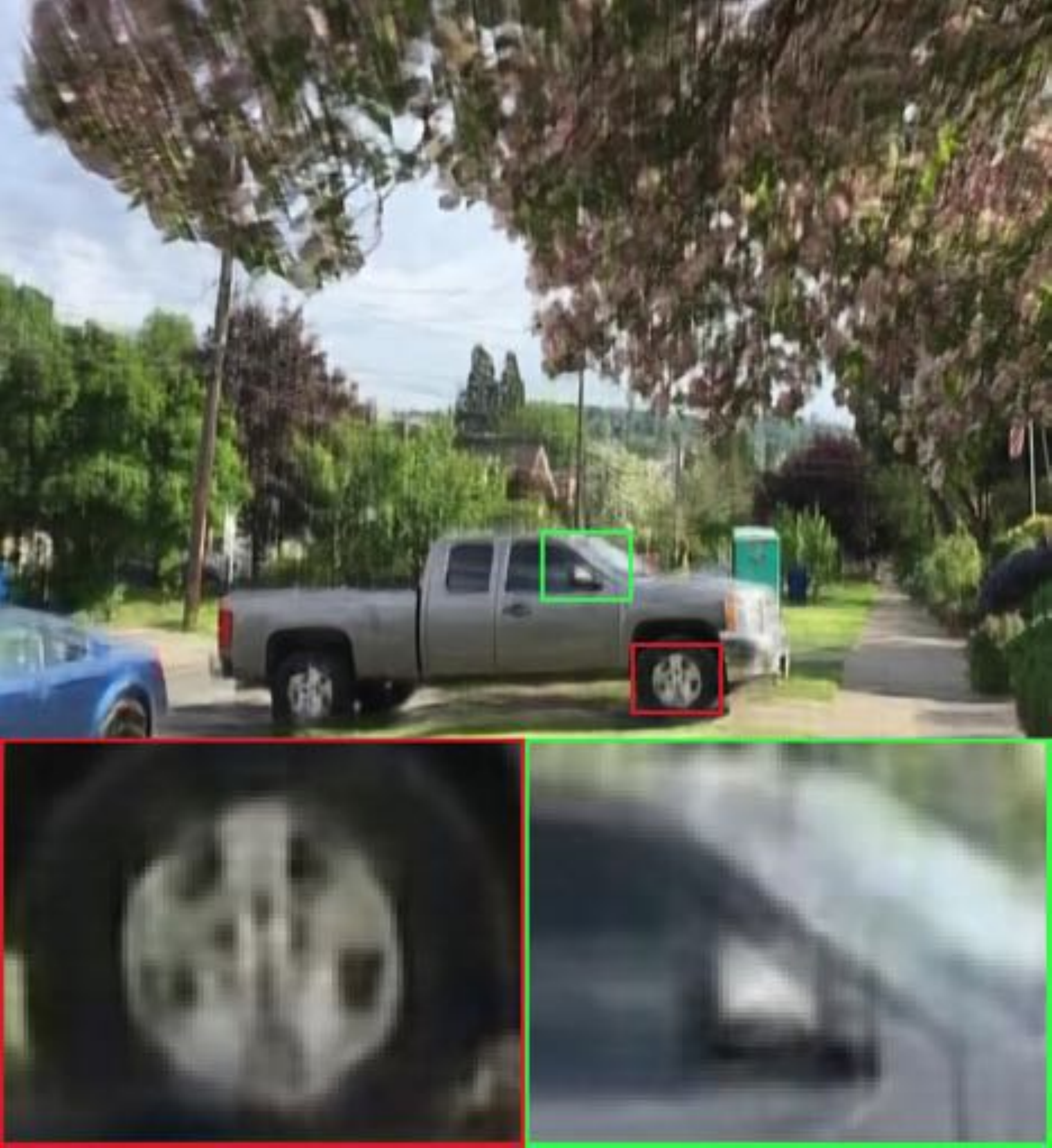} &	
    		\includegraphics[width=0.16\linewidth]{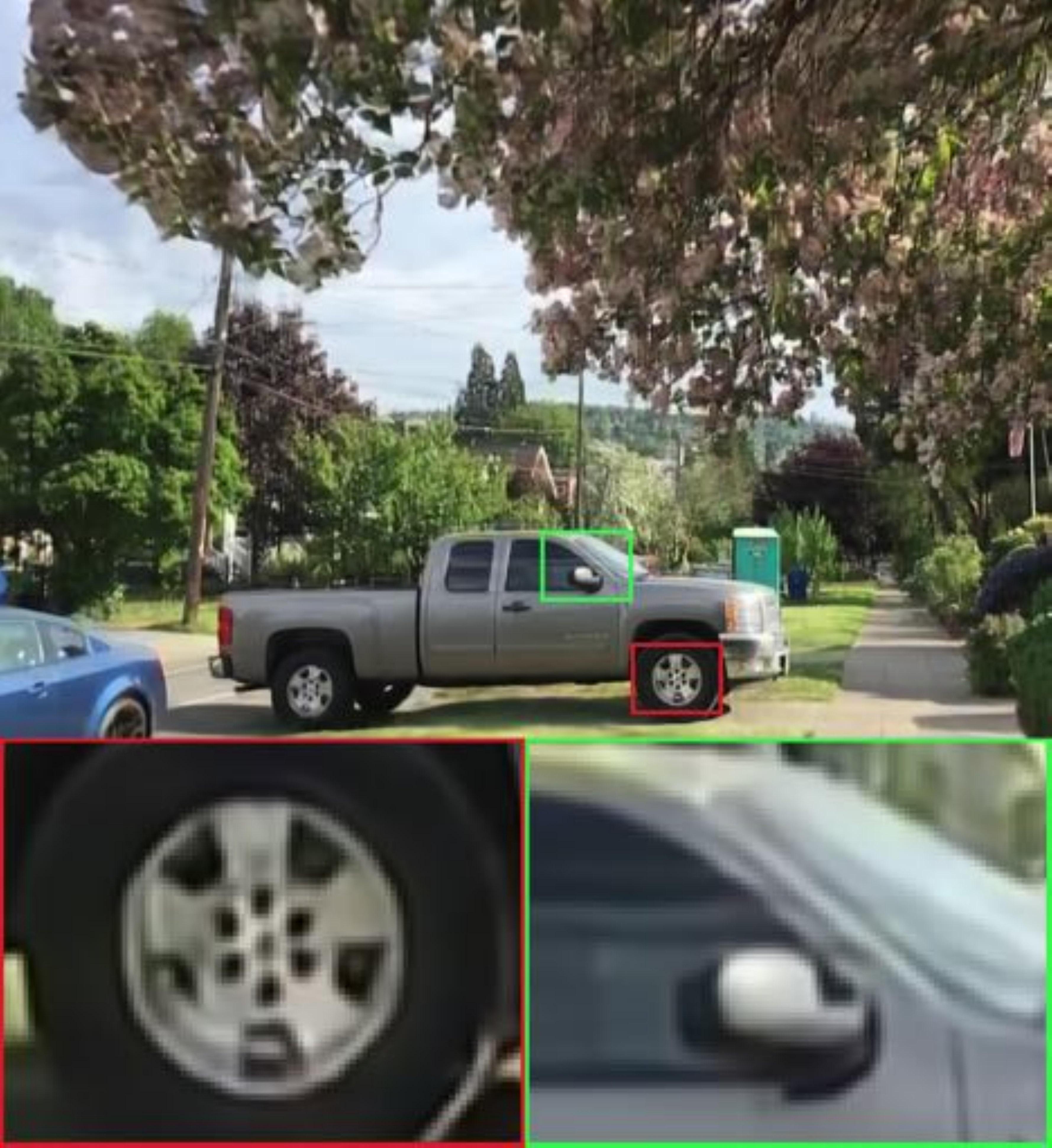} &
    		\includegraphics[width=0.16\linewidth]{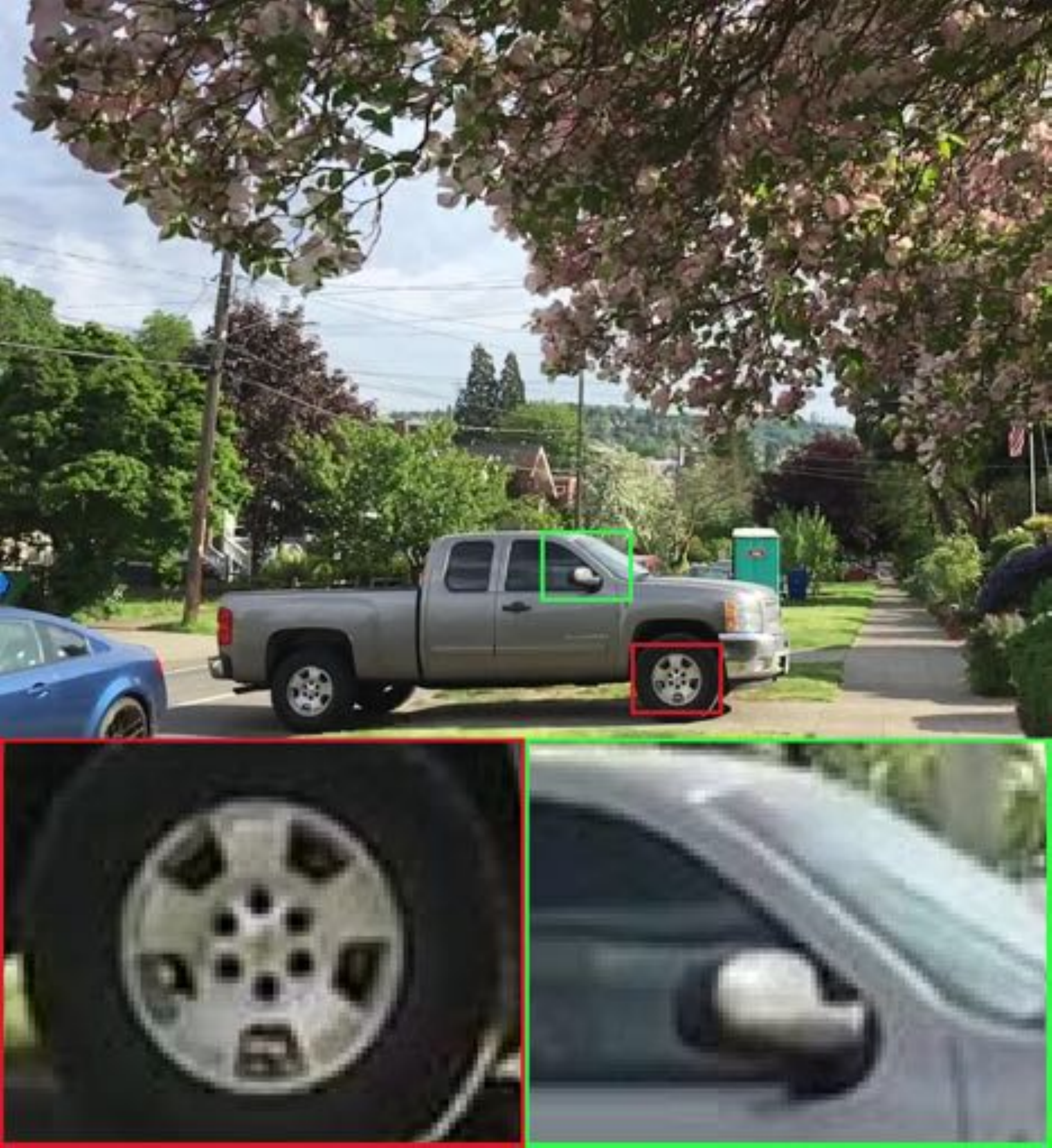}  \\		
    		\scriptsize (a) Input & \scriptsize (b) EDVR & \scriptsize (c) TSP & \scriptsize (d) PVDNet-L & \scriptsize (e) Ours & \scriptsize (f) GT\\
    	\end{tabular}
	}
    \caption{The qualitative results on the DVD dataset. Note that ``GT'' stands for ground truth.}
    \label{fig:show_dvd}
\end{figure*}

For fair comparisons, STDANet-Stack use same cascaded progressive structure like TSP~\cite{DBLP:conf/cvpr/PanBT20} and ARVo~\cite{DBLP:conf/cvpr/LiXZ0ZRSL21}. 
In the experiments, we use both peak signal-to-noise ratio (PSNR) and structural similarity (SSIM) as quantitative evaluation metrics for testing set.
Moreover, GMACs (Giga multiply-add operations per second) is used to evaluate the computational complexity. 

\subsection{Implementation Details}
To achieve better trade-off between video deblurring quality and computational efficiency, the $M, K, T$ are set as $4$, $12$ and $3$, respectively.
$\gamma$ is set to $0.05$.
The network is implemented with PyTorch~\cite{DBLP:conf/nips/PaszkeGMLBCKLGA19}~\footnote{The source code is available at \url{https://github.com/huicongzhang/STDAN}}.
The network is trained with a batch size of 8 on four NVIDIA Geforce RTX 2080 Ti GPUs.
The initial learning rate is set to $10^{-4}$.
The network is optimized using Adam optimizer~\cite{DBLP:journals/corr/KingmaB14} with $\beta_1 = 0.9$ and $\beta_2 = 0.999$. 
We randomly crop the input images into patches with resolutions of $256 \times 256$, along with random flipping or rotation during training.

\begin{table*}[!t]
\caption{The quantitative results on the GoPro dataset. Note that ``Ours$^*$'' denotes STDANet-Stack.}
\resizebox{\linewidth}{!} {
	\begin{tabular}{lccccccccc}
		\toprule
		Method
		& SRN
		& IFI-RNN-L
		& STFAN
		& EDVR
		& TSP
		& PVDNet
		& PVDNet-L
		& \textbf{Ours} 
		& \textbf{Ours$^{*}$} \\
		\midrule
		PSNR     & 30.61      & 31.05      & 28.59      & 31.54      & 31.67     & 31.52  & 31.98 &  \bf{32.29}  & \bf{32.62} \\
		SSIM      & 0.9080      & 0.9110      & 0.8608      & 0.9260    & 0.9279     & 0.9210   & 0.9280  &  \bf{0.9313} & \bf{0.9375}  \\
		GMACs & 1175      & 1,425      & 504      & 2739    & 6450      & 1004   & 1755  &  1677 & 6000  \\
		\bottomrule
	\end{tabular}
}
\label{tab:psnr_GoPro}
\end{table*}

\begin{figure*}[!t]
    \renewcommand{\tabcolsep}{0.5pt}
    \renewcommand{\arraystretch}{1}
    \resizebox{\linewidth}{!}{
    	\begin{tabular}{cccccc}
    		\includegraphics[width=0.195\linewidth]{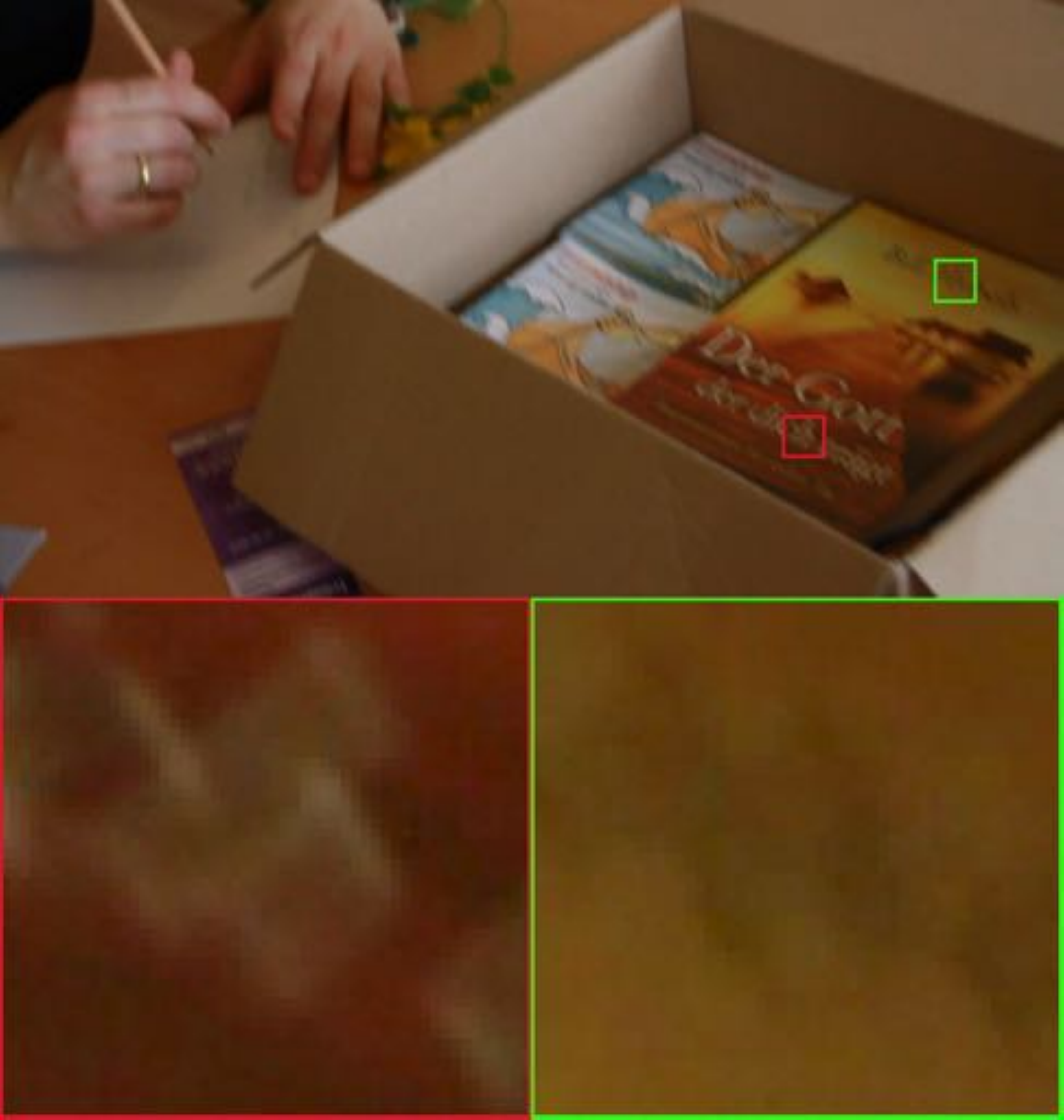} &
    		\includegraphics[width=0.195\linewidth]{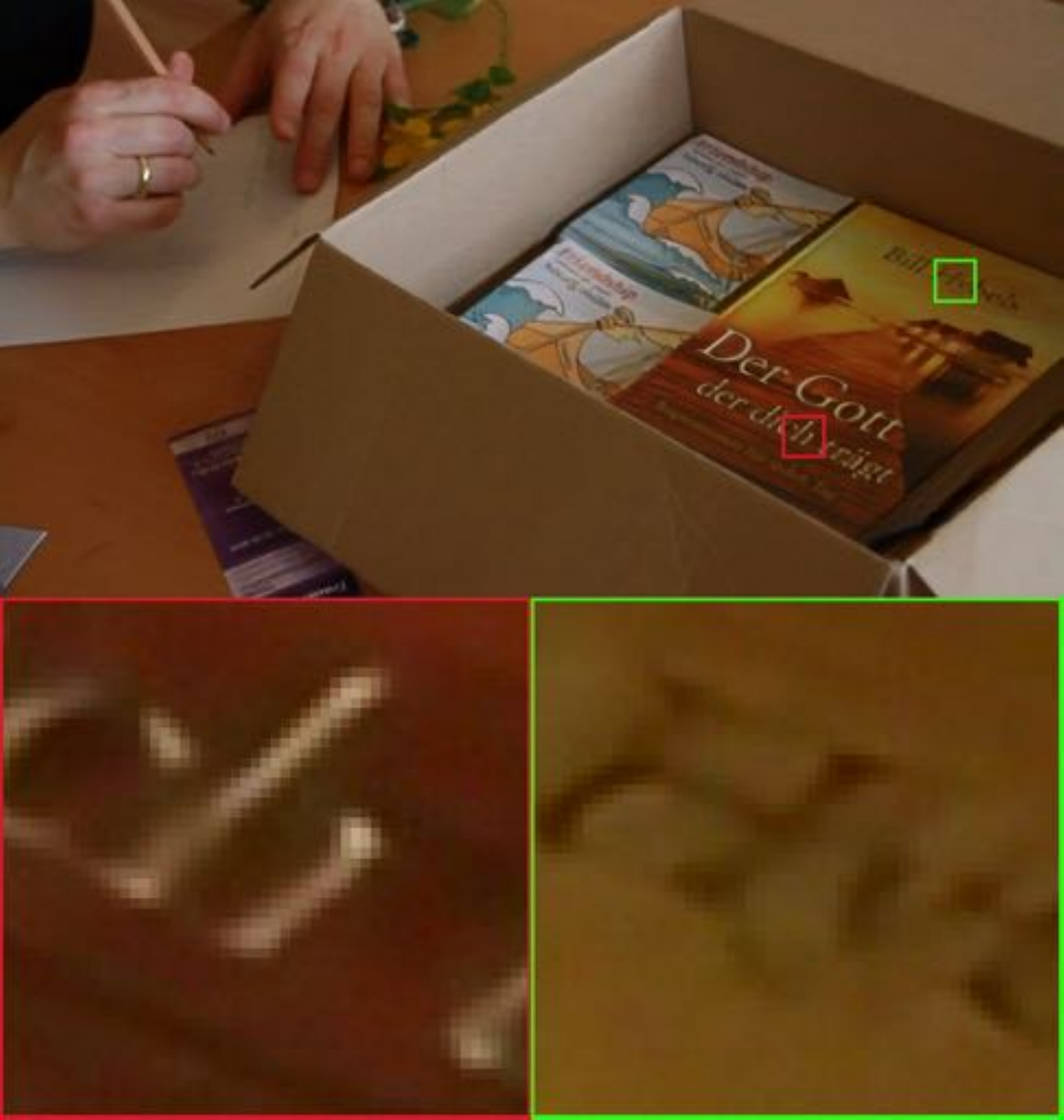} &
    		\includegraphics[width=0.195\linewidth]{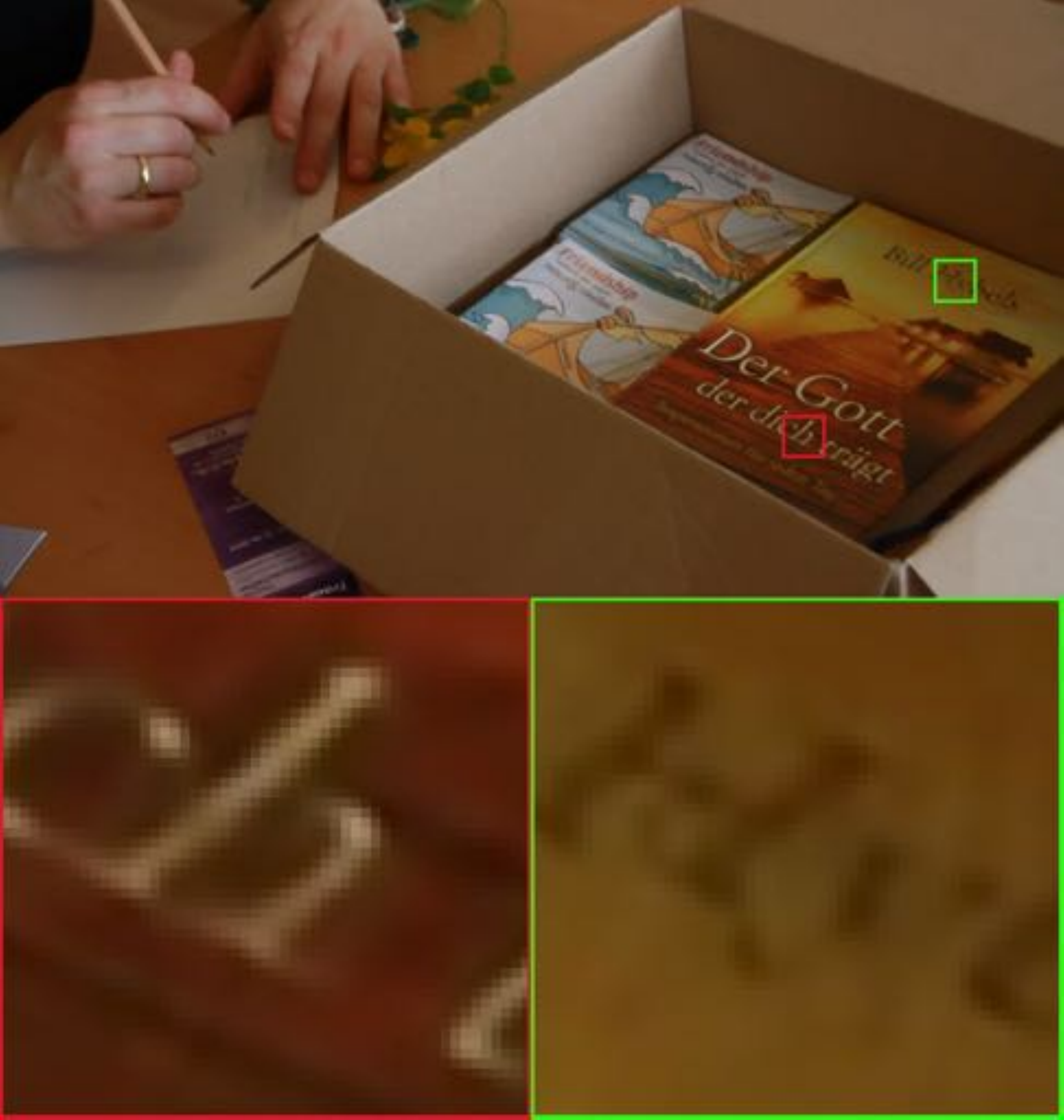} &
    		\includegraphics[width=0.195\linewidth]{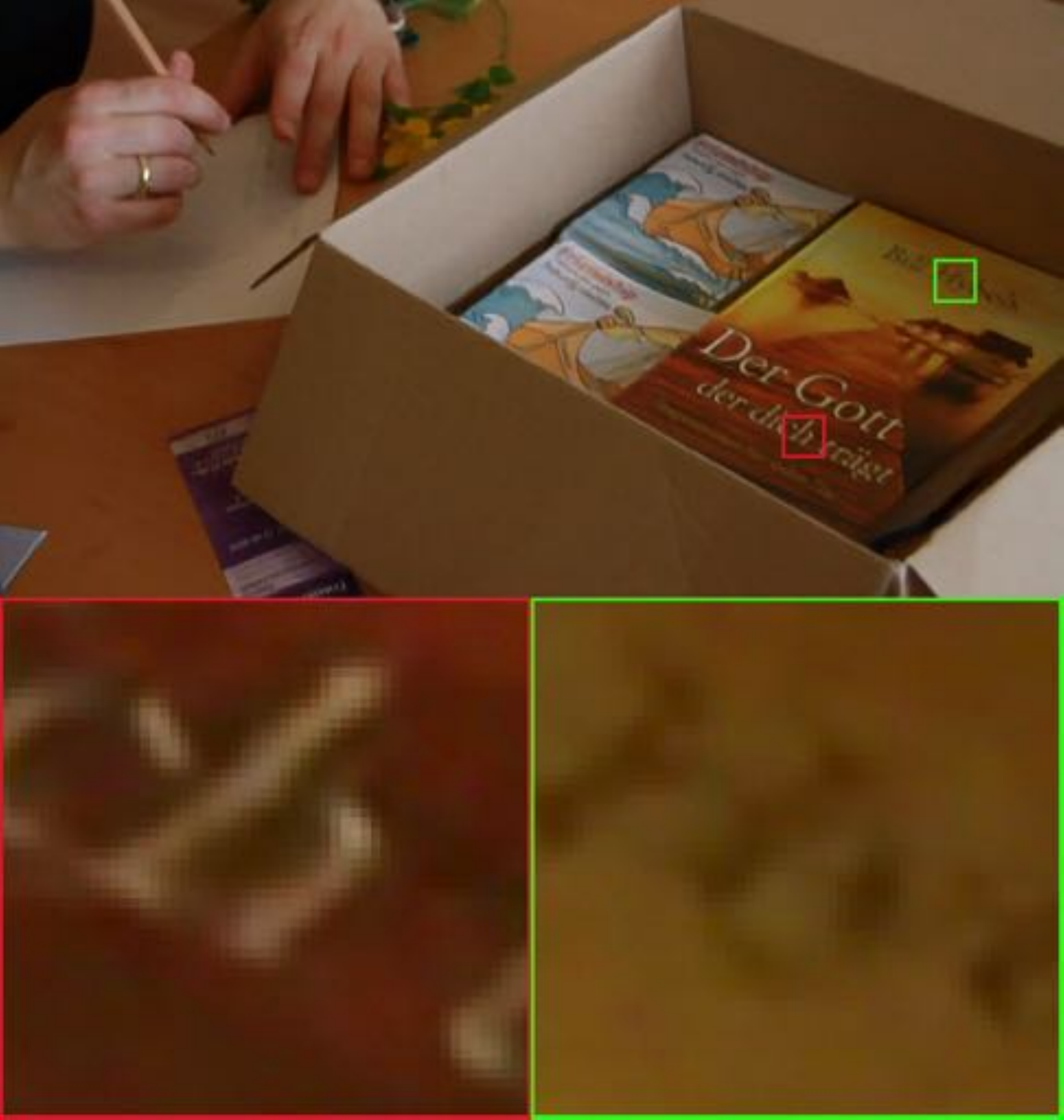} &	
    		\includegraphics[width=0.195\linewidth]{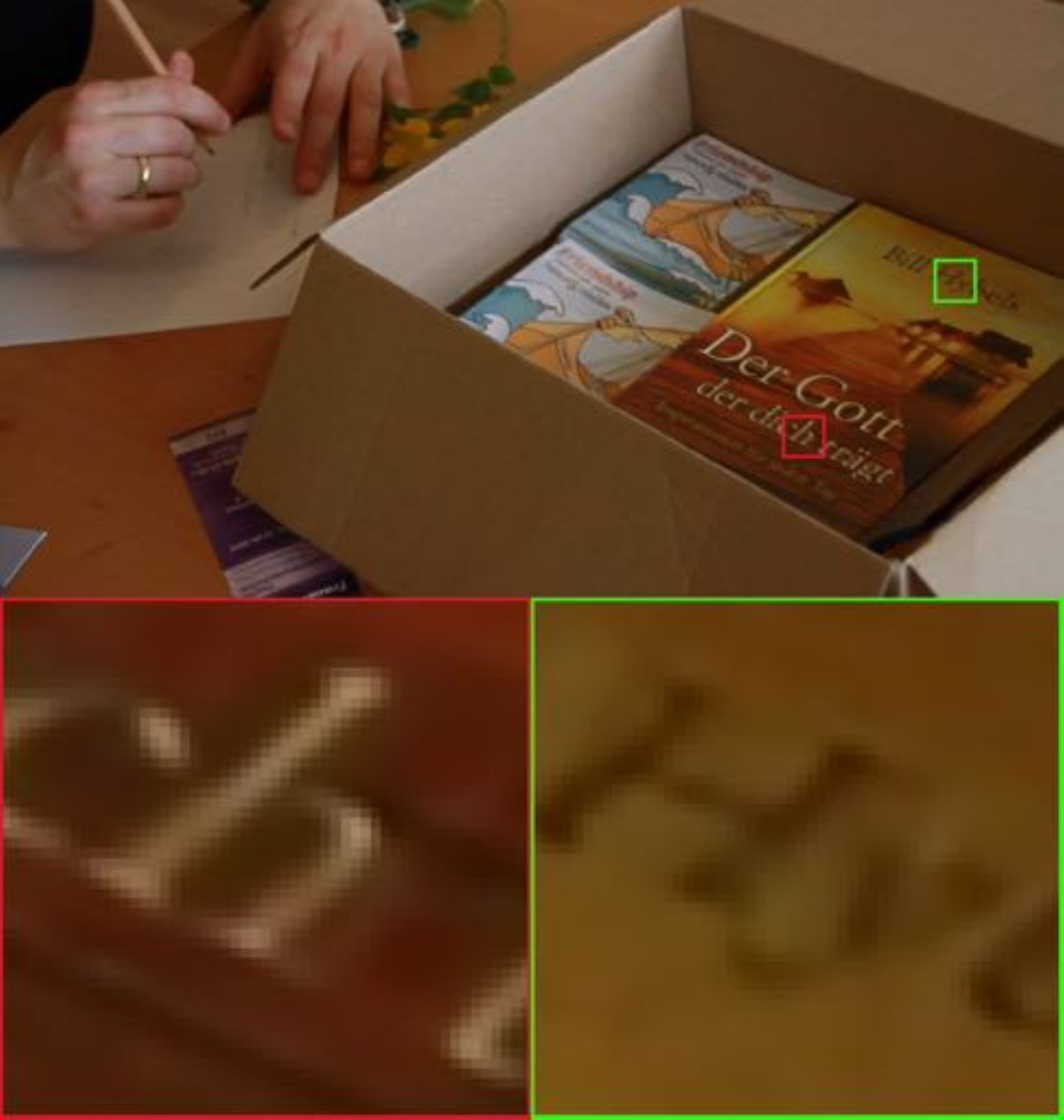}  \\		
    		 \scriptsize (a) Input & \scriptsize (b) EDVR & \scriptsize (c) TSP & \scriptsize (d) PVDNet  & \scriptsize (e) Ours \\
    	\end{tabular}
	}
    \caption{The qualitative results of real blur images from the DVD dataset. There are no corresponding ground truth for the real blur images.}
    \label{fig:show_dvd_real}
\end{figure*}

\begin{figure*}[!t]
    \renewcommand{\tabcolsep}{0.5pt}
    \renewcommand{\arraystretch}{1}
    \resizebox{\linewidth}{!}{
    	\begin{tabular}{cccccc}
    		\includegraphics[width=0.16\linewidth]{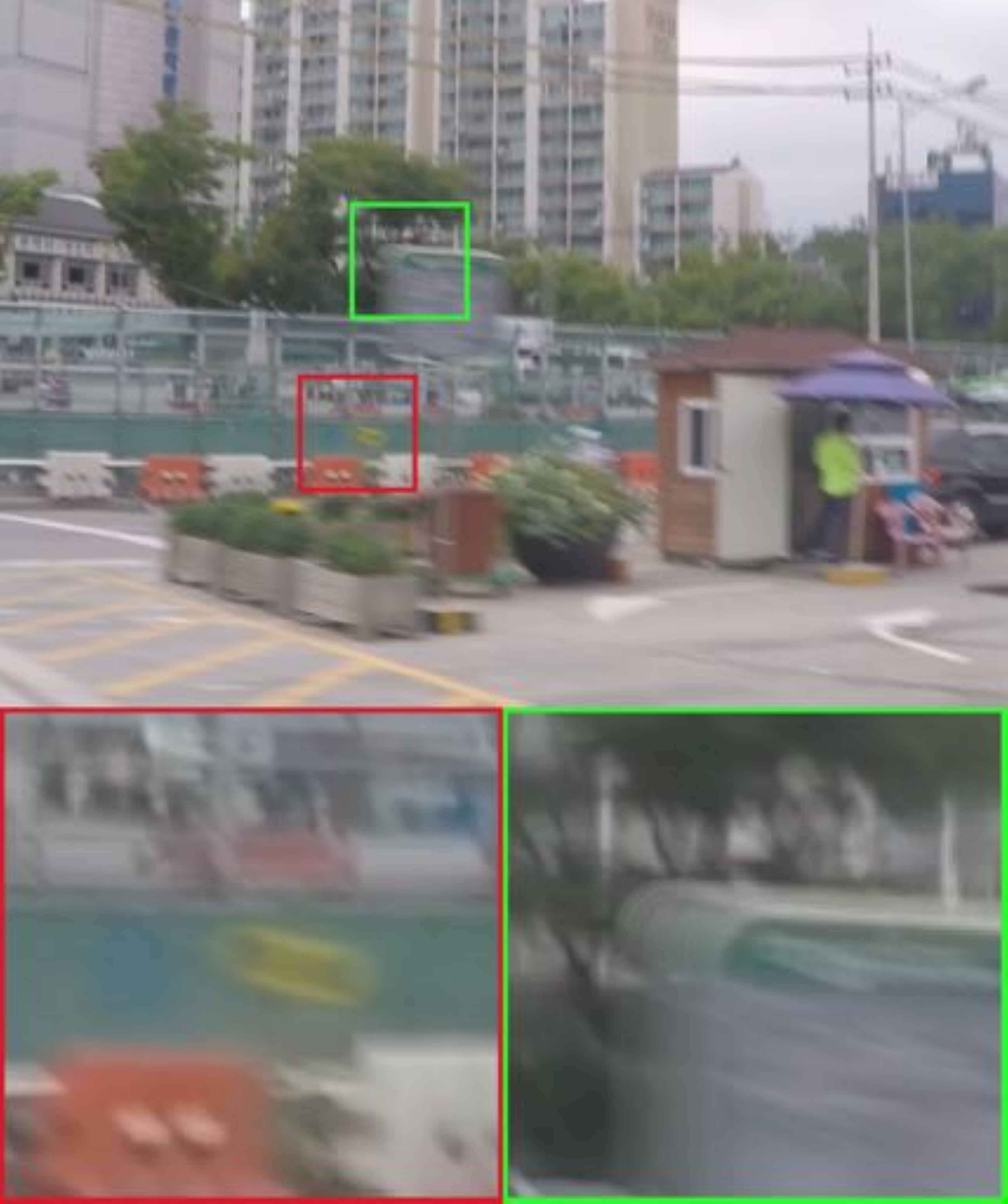} &
    		\includegraphics[width=0.16\linewidth]{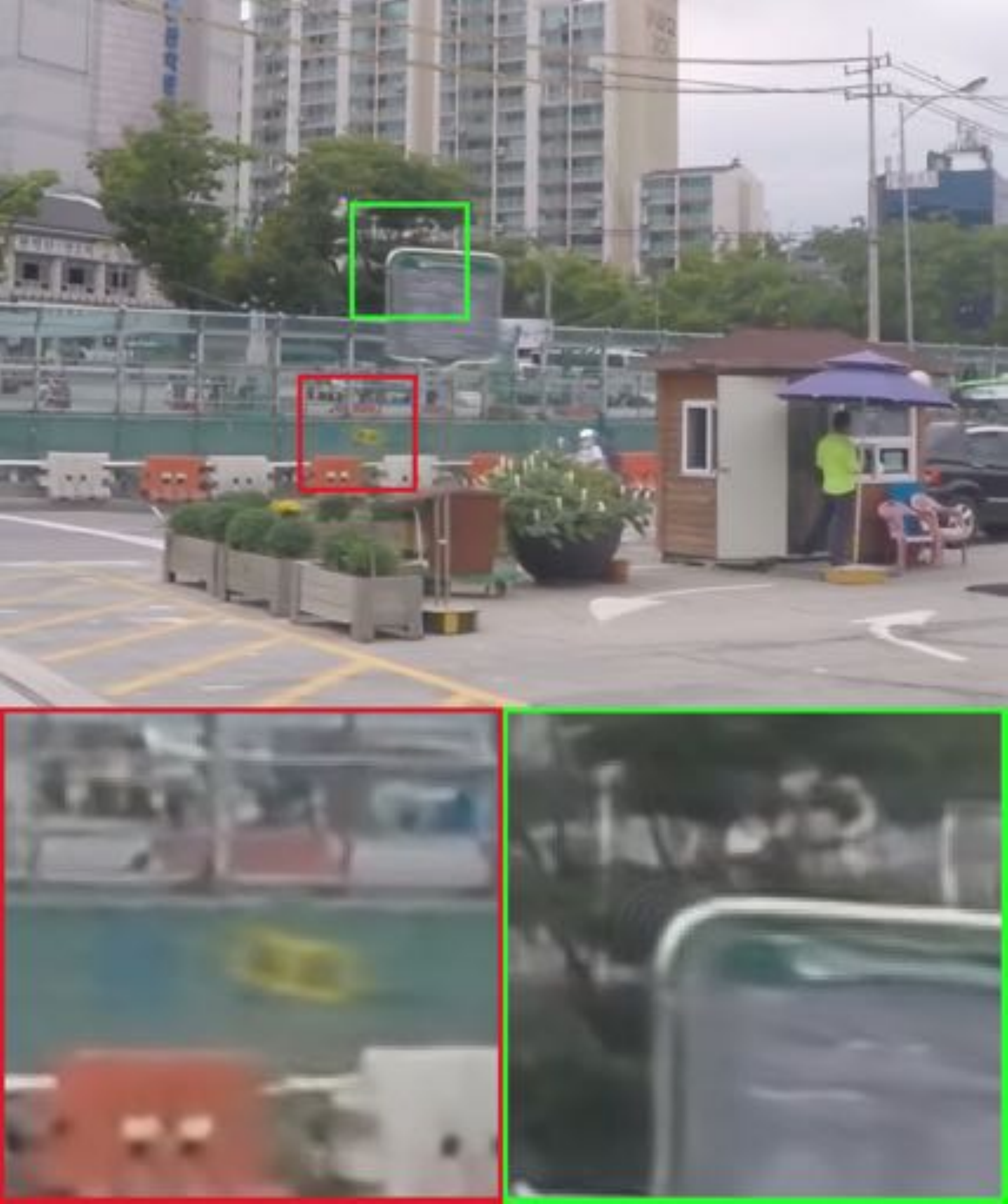} &
    		\includegraphics[width=0.16\linewidth]{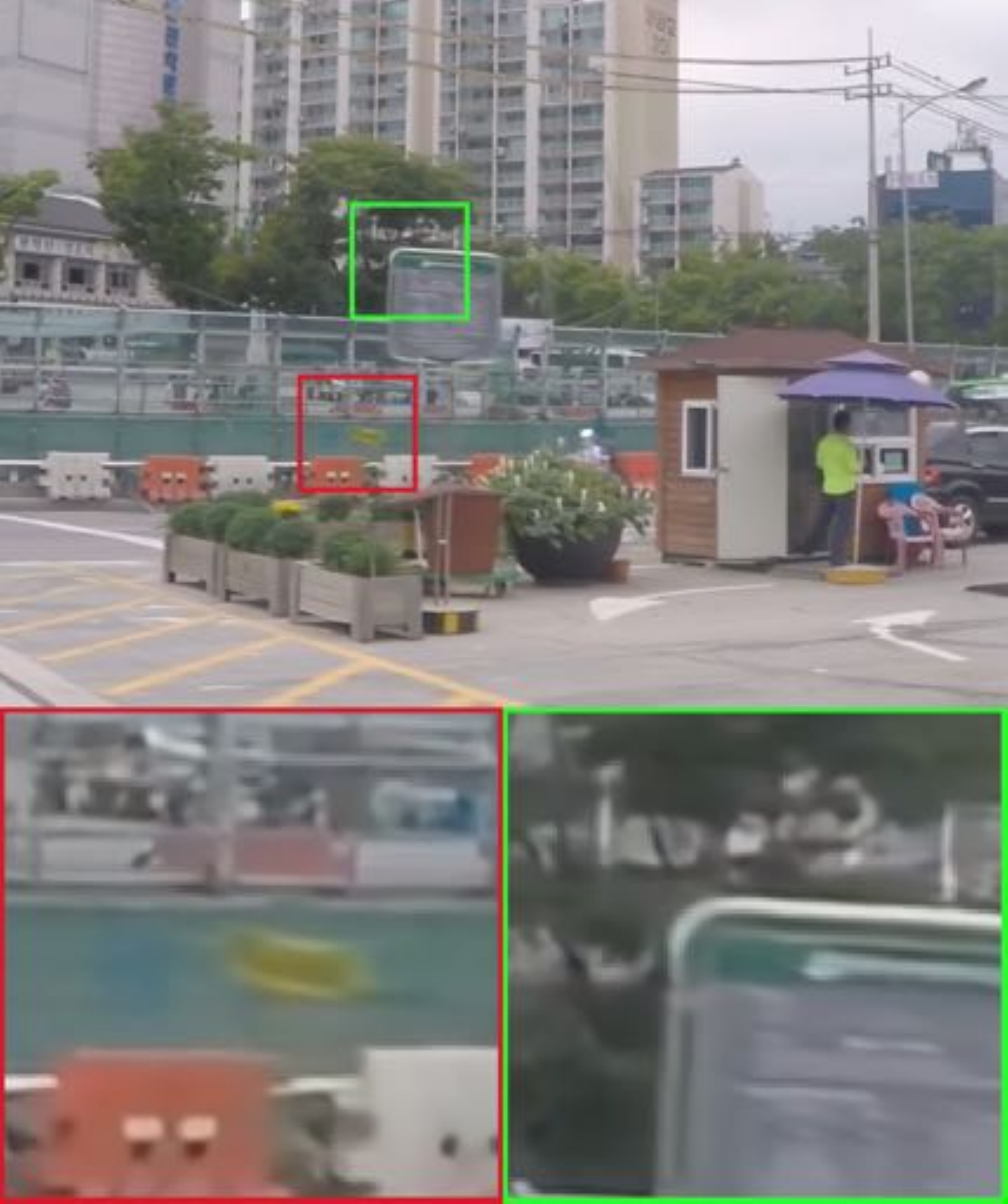} &
    		\includegraphics[width=0.16\linewidth]{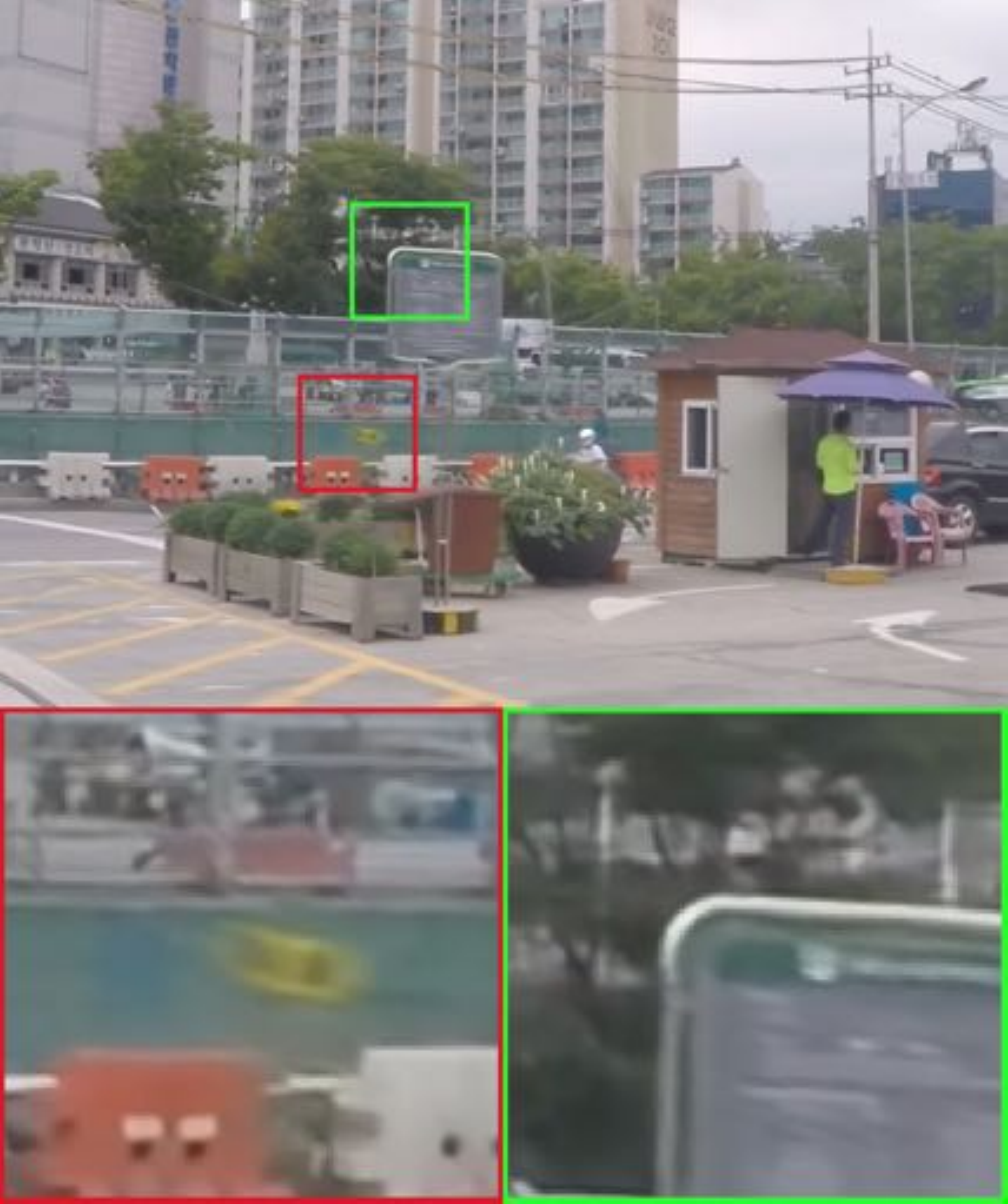} &	
    		\includegraphics[width=0.16\linewidth]{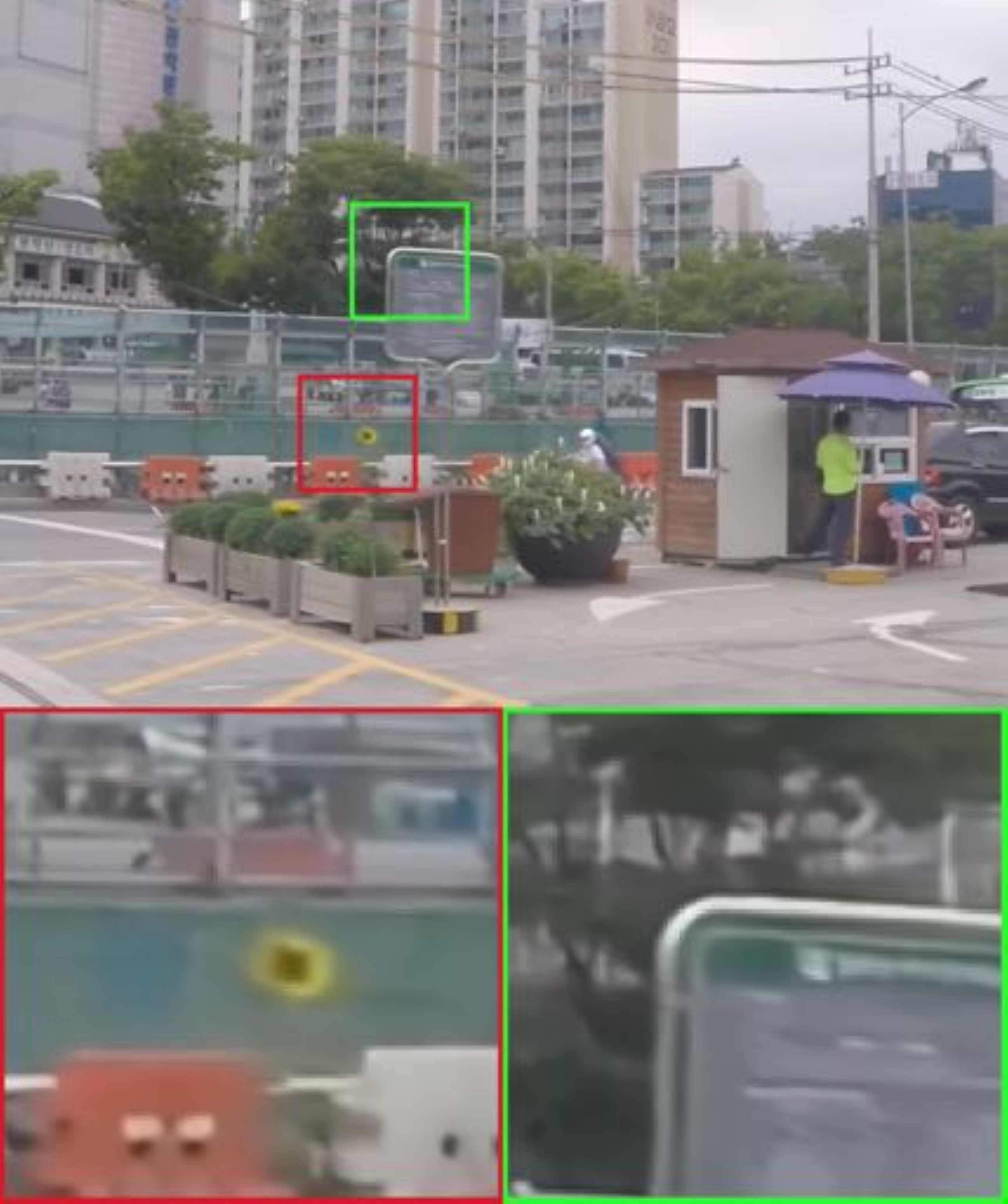} &
    		\includegraphics[width=0.16\linewidth]{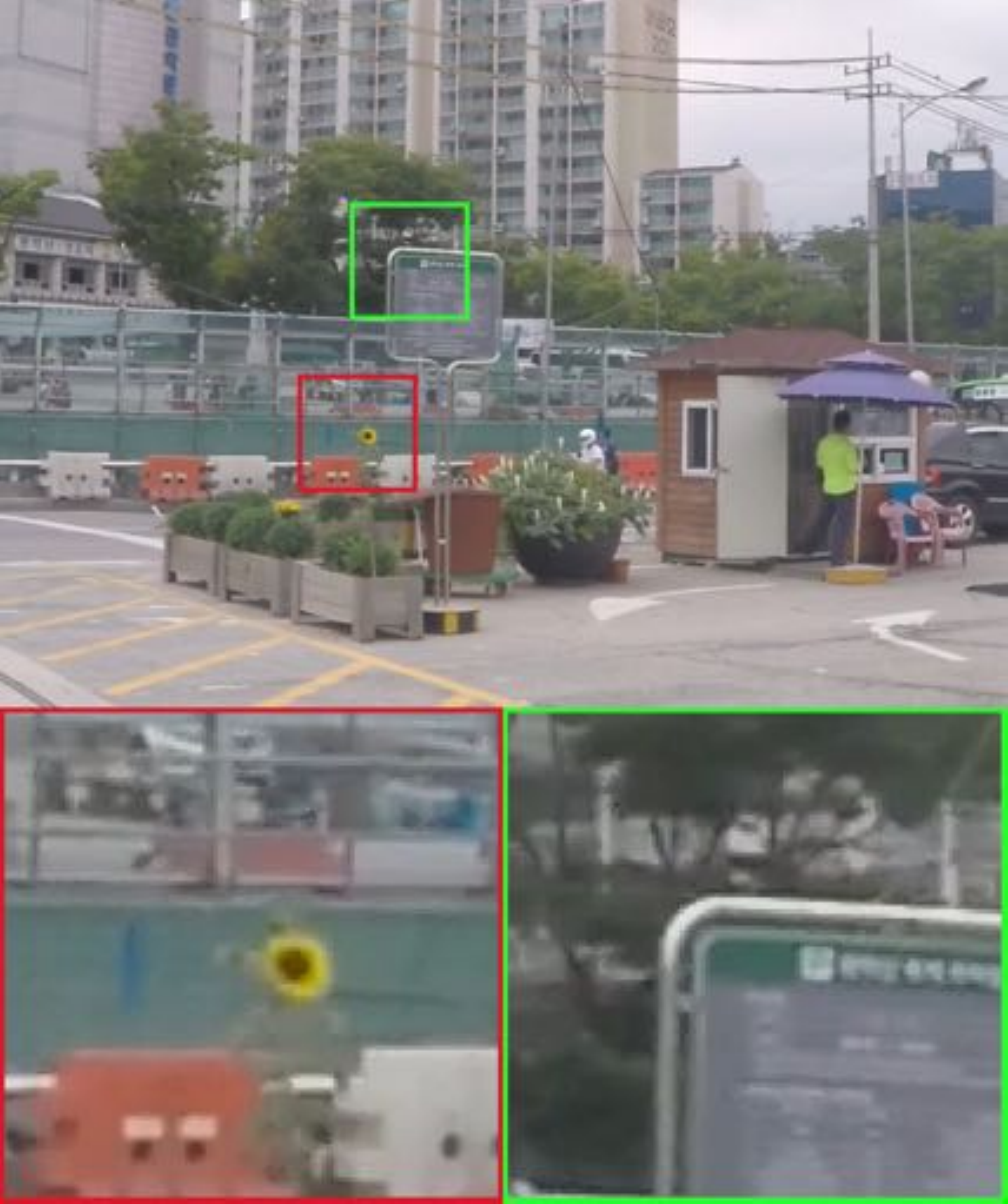} 
    		%
    		\\
    		\includegraphics[width=0.16\linewidth]{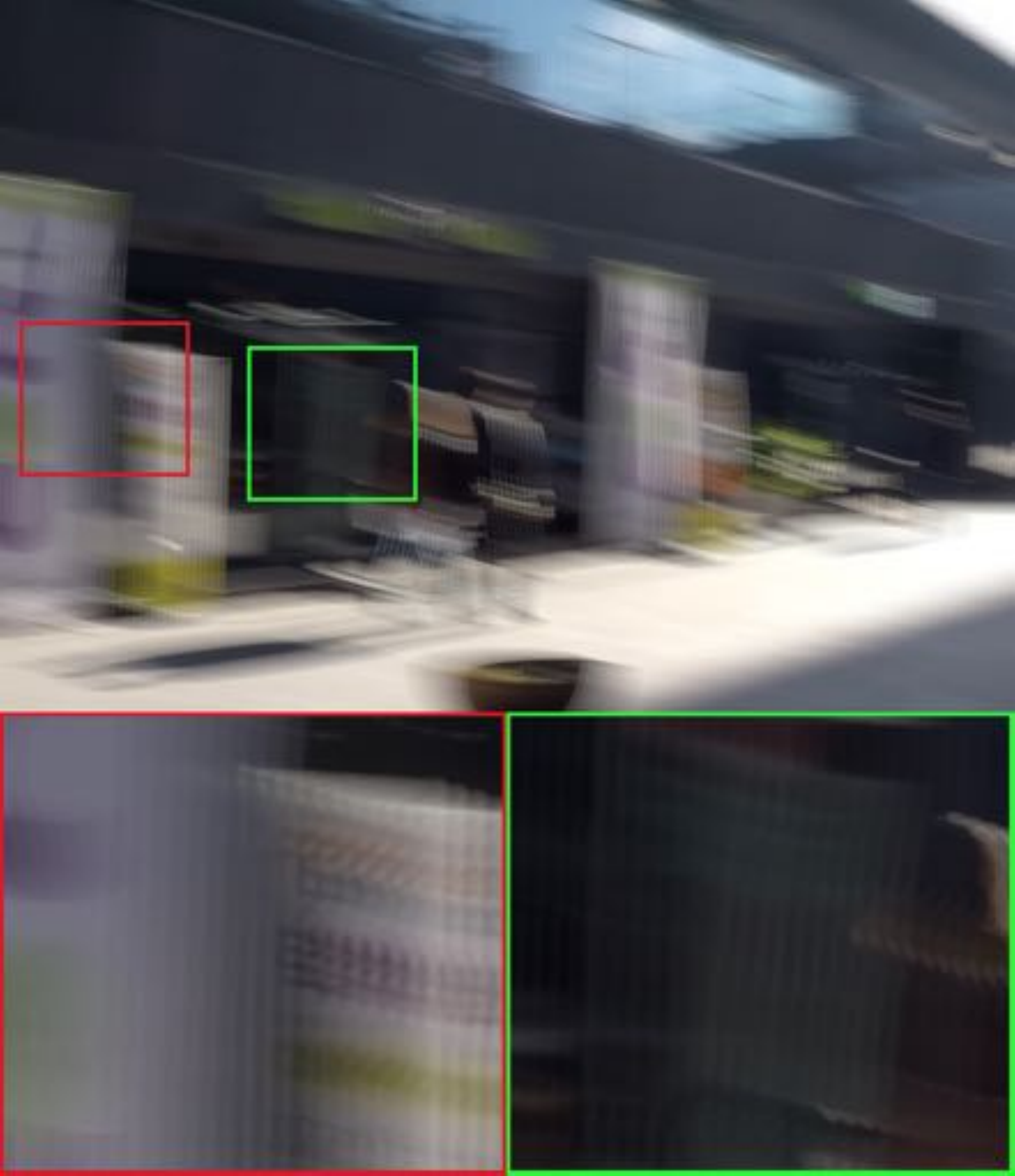} &
    		\includegraphics[width=0.16\linewidth]{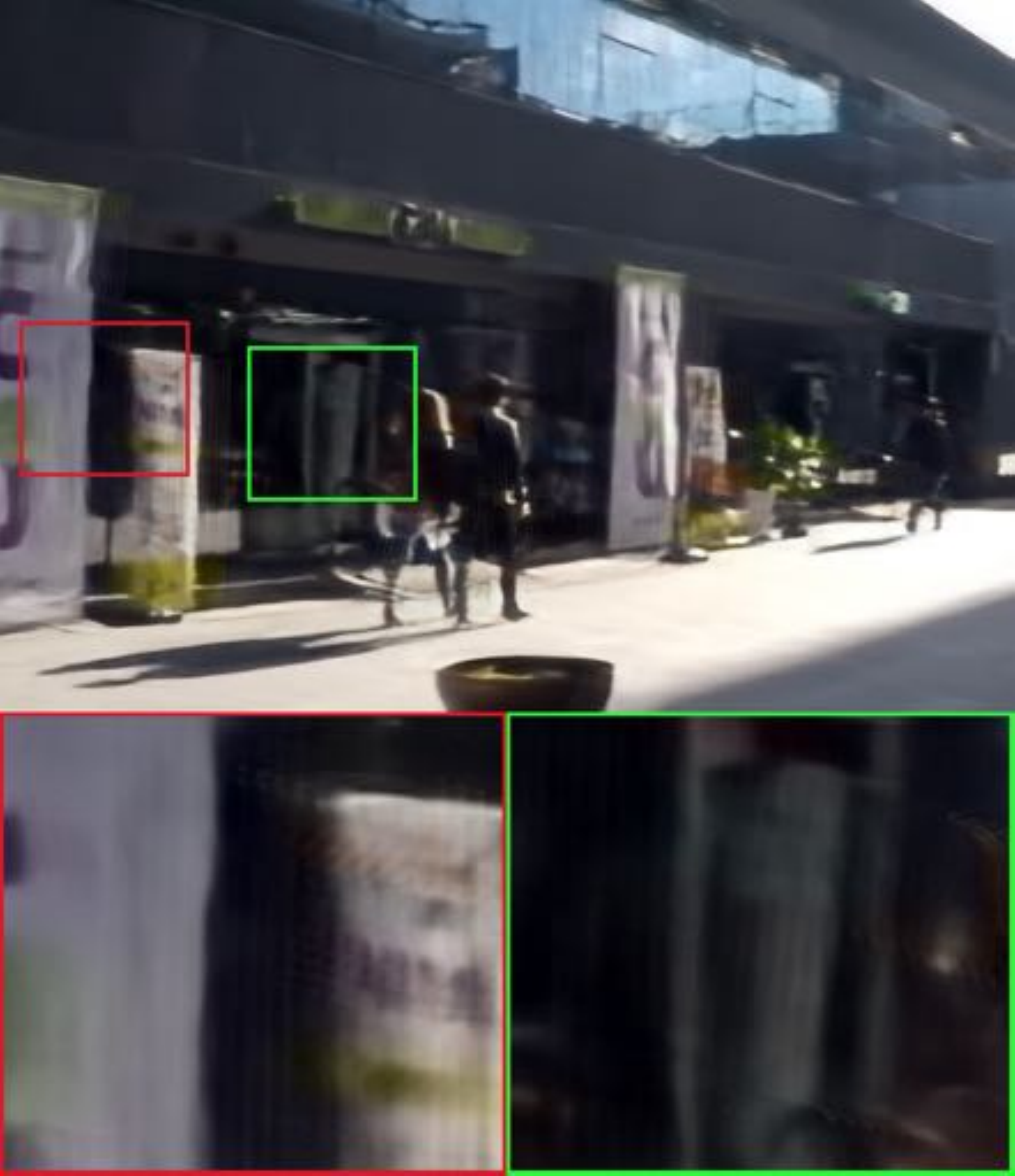} &
    		\includegraphics[width=0.16\linewidth]{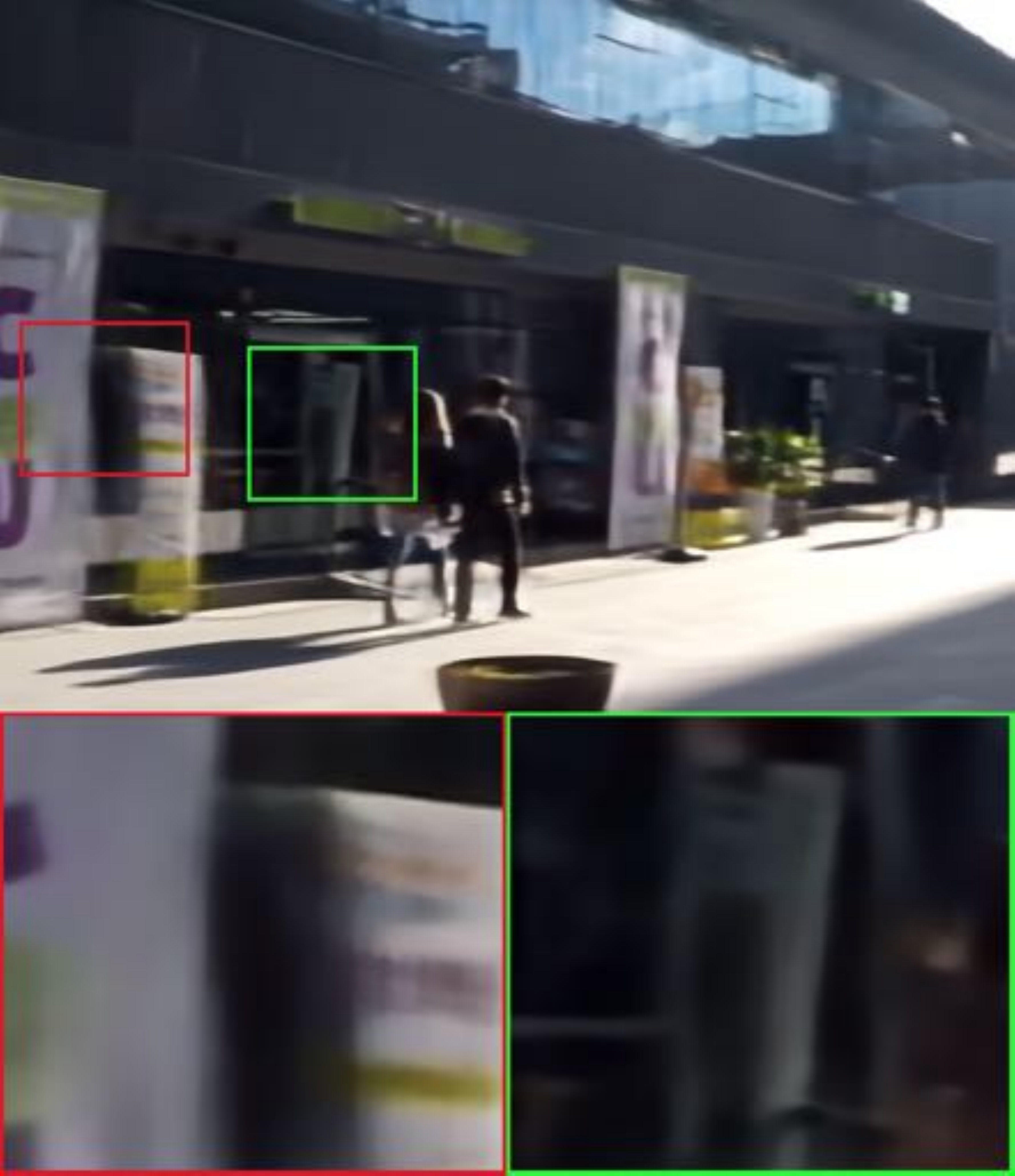} &
    		\includegraphics[width=0.16\linewidth]{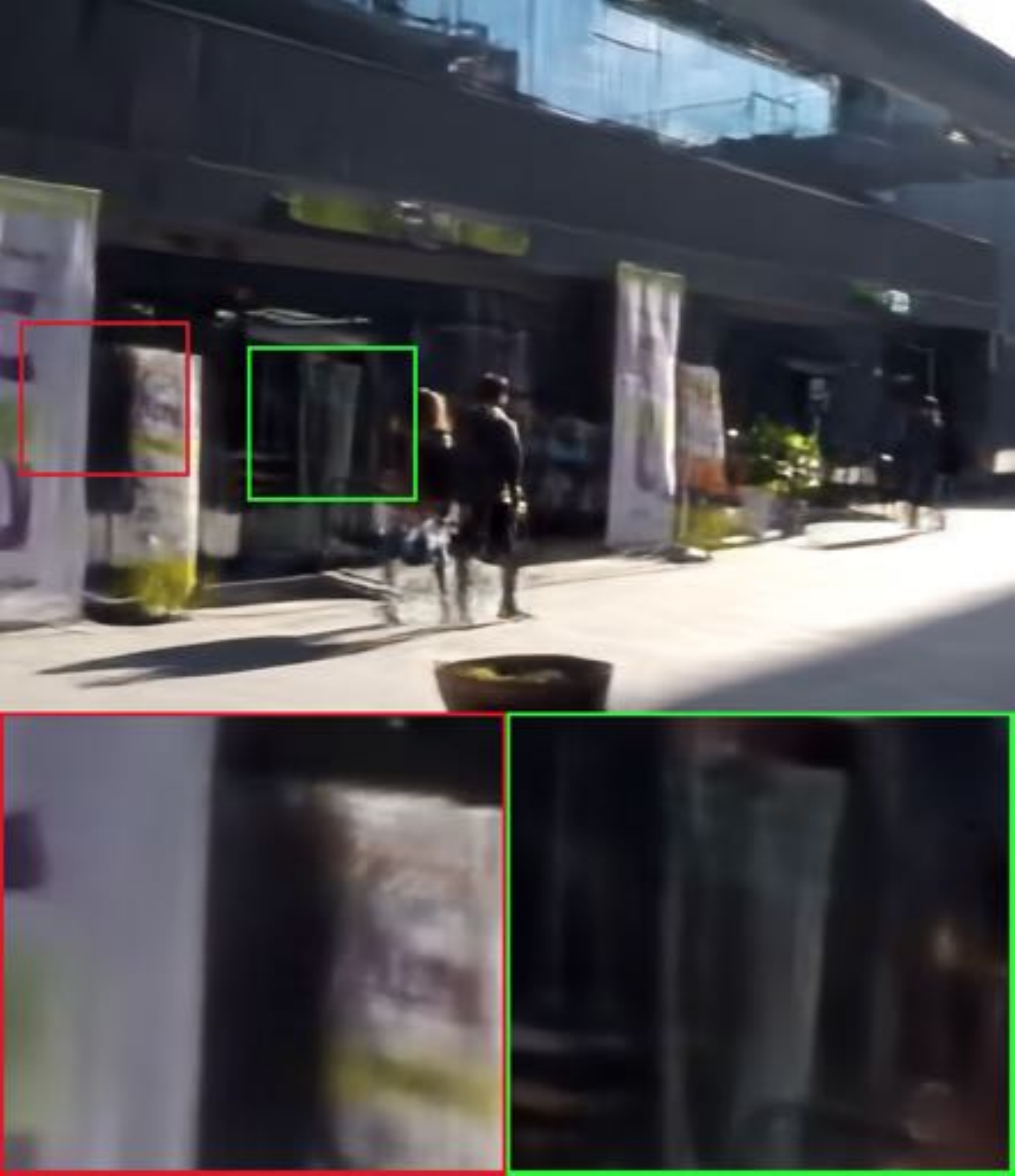} &	
    		\includegraphics[width=0.16\linewidth]{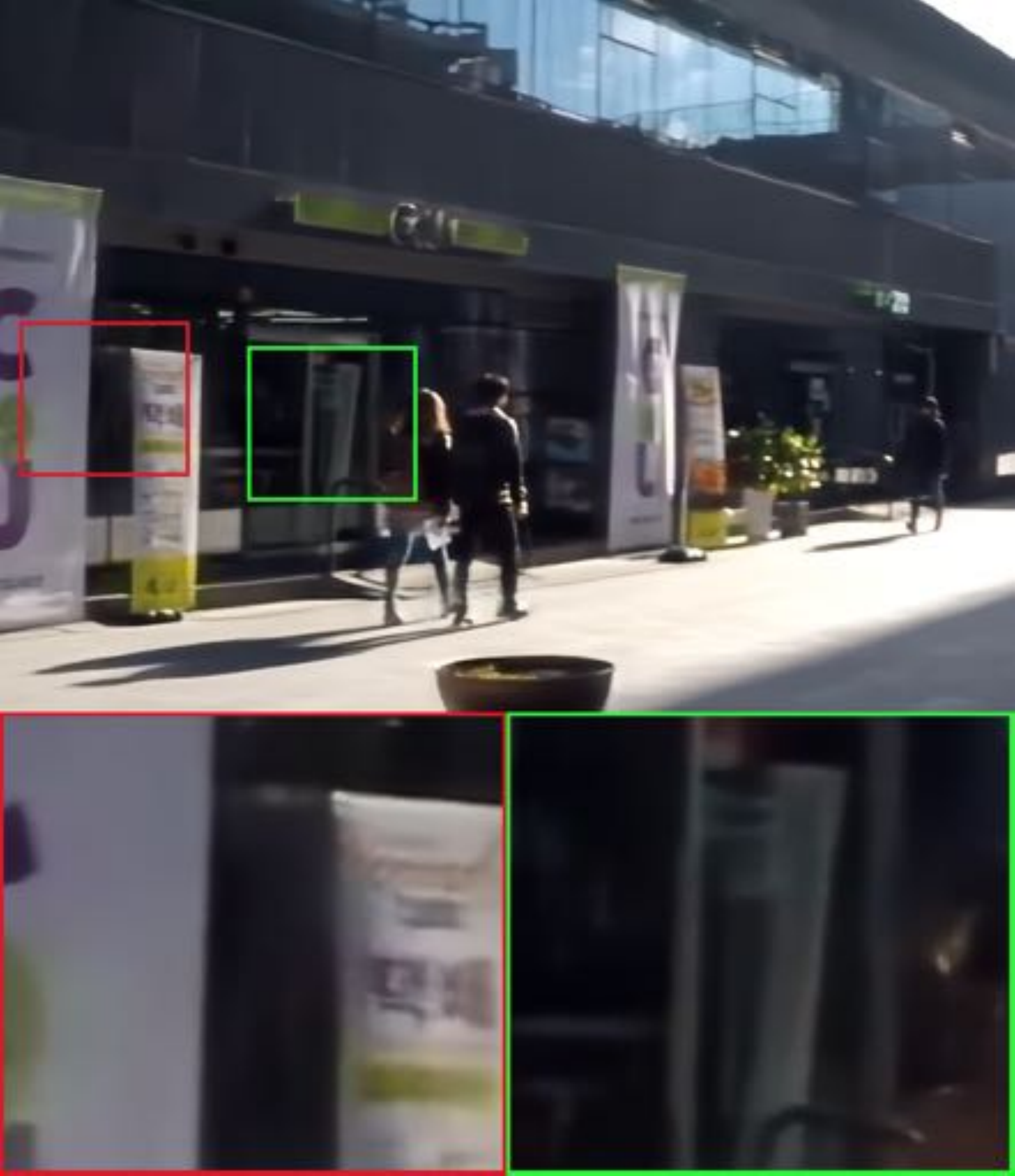} &
    		\includegraphics[width=0.16\linewidth]{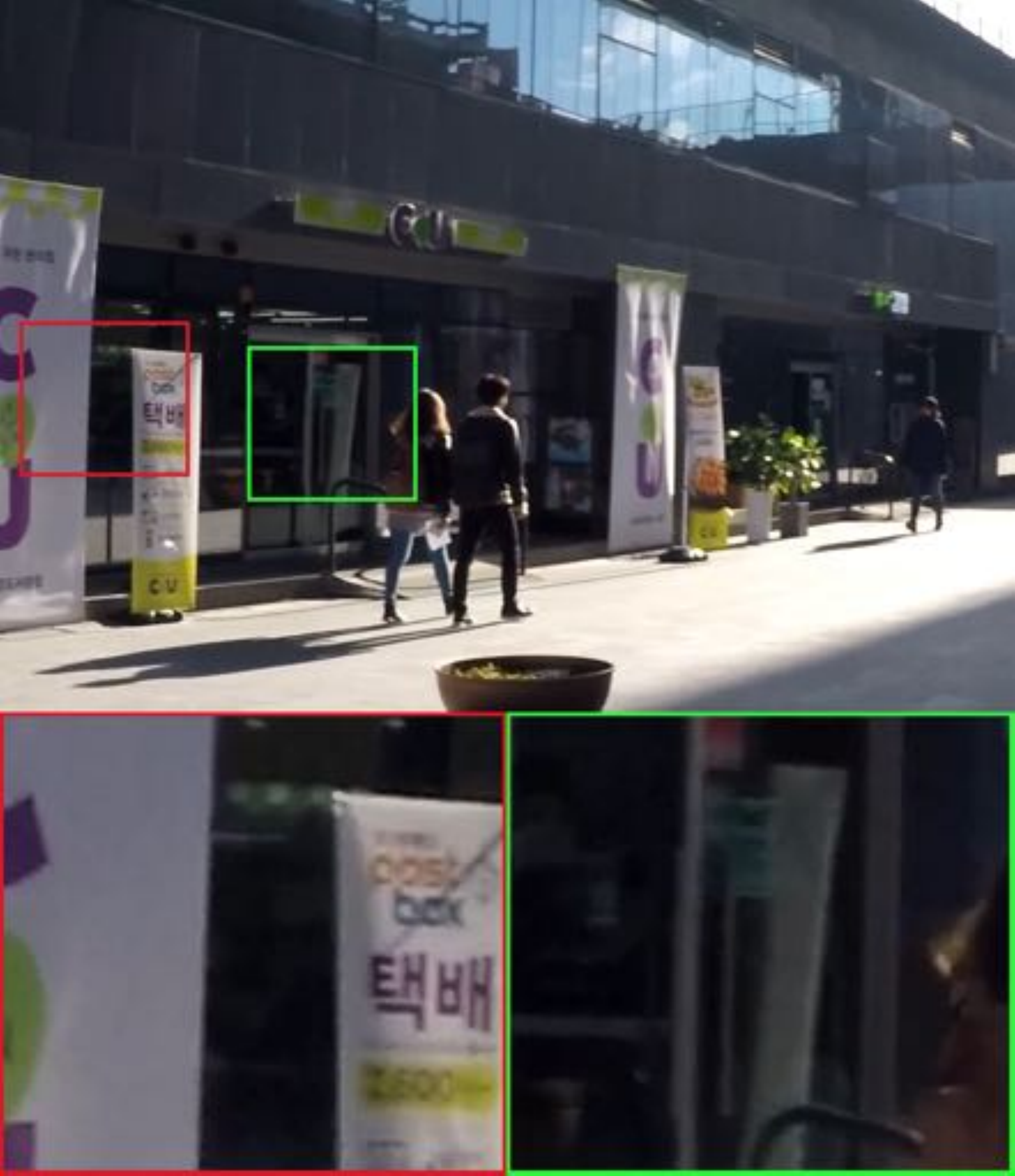}  \\		
    		 \scriptsize (a) Input & \scriptsize (b) EDVR & \scriptsize (c) TSP & \scriptsize (d) PVDNet-L & \scriptsize (e) Ours & \scriptsize (f) GT \\
    	\end{tabular}
	}
    \caption{The qualitative results on the GoPro dataset. Note that ``GT'' stands for ground truth.}
    \label{fig:show_gopro}
\end{figure*}

\subsection{Experimental Results}

\noindent \textbf{The DVD dataset.} 
To evaluate the performance of the proposed method, we compare it with the state-of-the-art methods.  
Table~\ref{tab:psnr_dvd} shows the quantitative results on the DVD dataset~\cite{DBLP:conf/cvpr/SuDWSHW17}, where IFI-RNN-L~\cite{DBLP:journals/tog/SonLLCL21} is larger IFI-RNN~\cite{DBLP:conf/cvpr/NahSL19}. 
The proposed method outperforms the state-of-the-art methods in term of PSNR and SSIM. 
Compared to the best state-of-the-art method ARVo, the proposed STDANet-Stack improves the PSNR and SSIM by 0.25dB and 0.0022, respectively. 
Figure~\ref{fig:show_dvd} shows several examples in the testing set, which indicates that existing state-of-the-art methods are less effective when the inputs contain heavy blur. 
We further compare the proposed method with state-of-the-art methods on the real blur images from the DVD dataset.
Figure~\ref{fig:show_dvd_real} shows that the proposed method generates sharper images with more visual details, which demonstrates the superiority of removing the unknown real blur in dynamic scenes robustly. 

\begin{table}[!t]
    \caption{The quantitative results on the BSD dataset. Note that ``Ours$^*$'' denotes STDANet-Stack.}
	\label{table:bsd}
	\centering
	\begin{tabularx}{\linewidth}{lYYYYYY}
		\toprule
		\multirow{2}{*}{Method} & 
		\multicolumn{2}{c}{1ms--8ms} & 
		\multicolumn{2}{c}{2ms--16ms} & 
		\multicolumn{2}{c}{3ms--24ms} \\
		\cline{2-3} \cline{4-5} \cline{6-7}
		& PSNR & SSIM & PSNR & SSIM & PSNR & SSIM\\
		\midrule
		IFIRNN &33.00 &  0.9330&  31.53& 0.9190& 30.89& 0.9170 \\
		ESTRNN~ &33.36 & 0.9370& 31.95& 0.9250& 31.39& 0.9260 \\
		EDVR &32.79  &0.9264  &31.99 &0.9129 &31.53  & 0.9192 \\
		TSP & 33.62&  0.9419&  32.19& 0.9285& 31.68& 0.9266\\
		PVDNet-L &  33.93& 0.9392&  32.46& 0.9290& 31.87& 0.9293\\
		\midrule
		\textbf{Ours} &  \textbf{34.21}& \textbf{0.9446}&  \textbf{33.13}& \textbf{0.9388}& \textbf{32.65}& \textbf{0.9409}\\
		\textbf{Ours$^{*}$} &  \textbf{34.32}& \textbf{0.9456}&  \textbf{33.27}& \textbf{0.9420}& \textbf{32.83}& \textbf{0.9443}\\
		\bottomrule
	\end{tabularx}
\end{table} 

\begin{table}[!t]
	\centering
	\caption{The quantitative results on the GoPro dataset in terms of PSNR and SSIM when replacing MMA and MSA layers with the concatenation operation.}
	\begin{tabularx}{\linewidth}{Y|YYYY}
		\toprule
		MMA &  &            & \checkmark & \checkmark \\
		MSA &  & \checkmark &            & \checkmark \\
		\midrule
		PSNR          & 30.12       & 31.15       &  31.18          & \bf{32.29}      \\
		SSIM           & 0.8950       & 0.9146       & 0.9152          & \bf{0.9313}    \\
		\bottomrule
	\end{tabularx}
	\label{tab:structure}
\end{table}

\begin{figure*}[!t]
    \renewcommand{\tabcolsep}{0.5pt}
    \renewcommand{\arraystretch}{1}
    \resizebox{\linewidth}{!}{
    	\begin{tabular}{cccccc}
    		\includegraphics[width=0.16\linewidth]{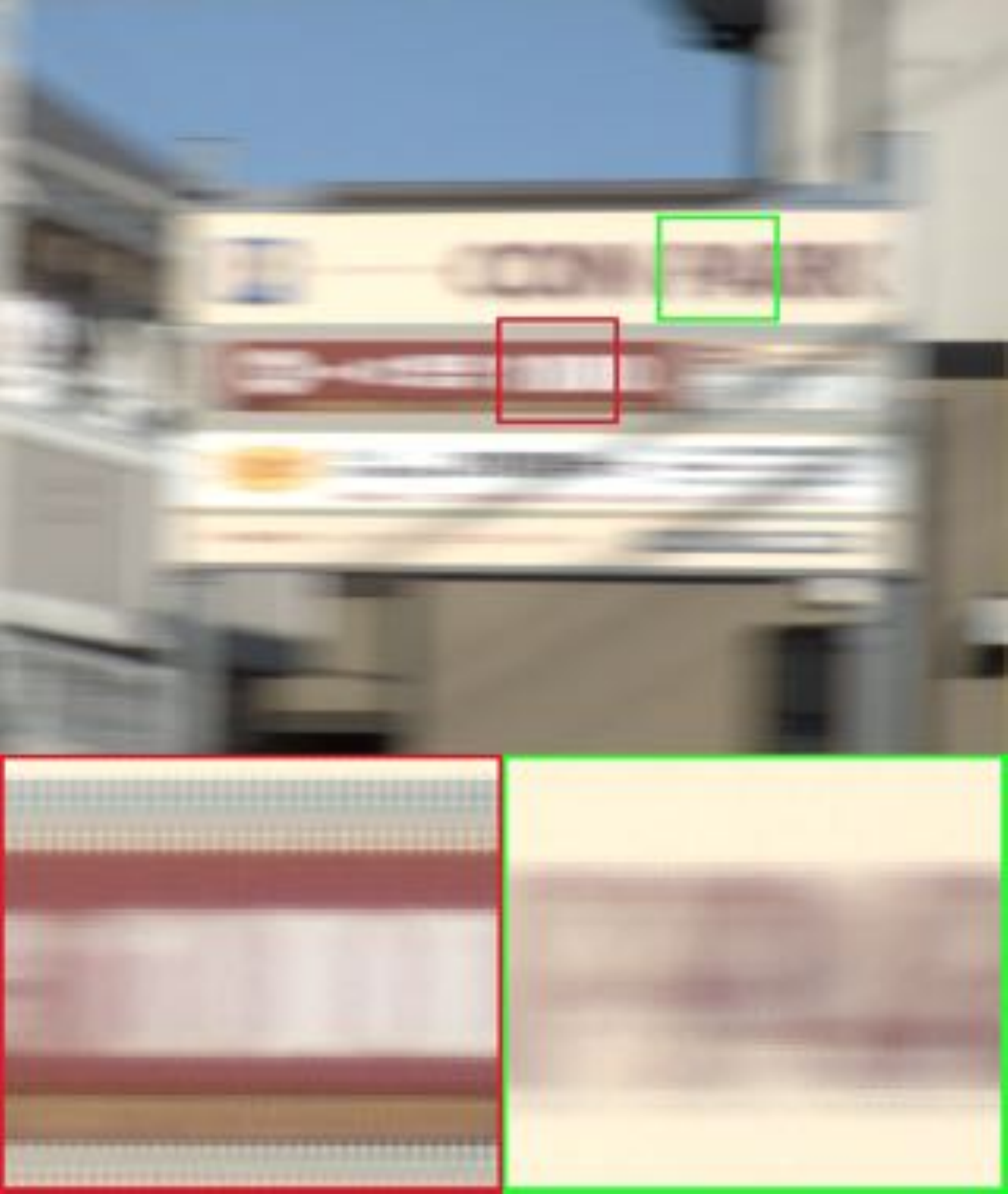} &
    		\includegraphics[width=0.16\linewidth]{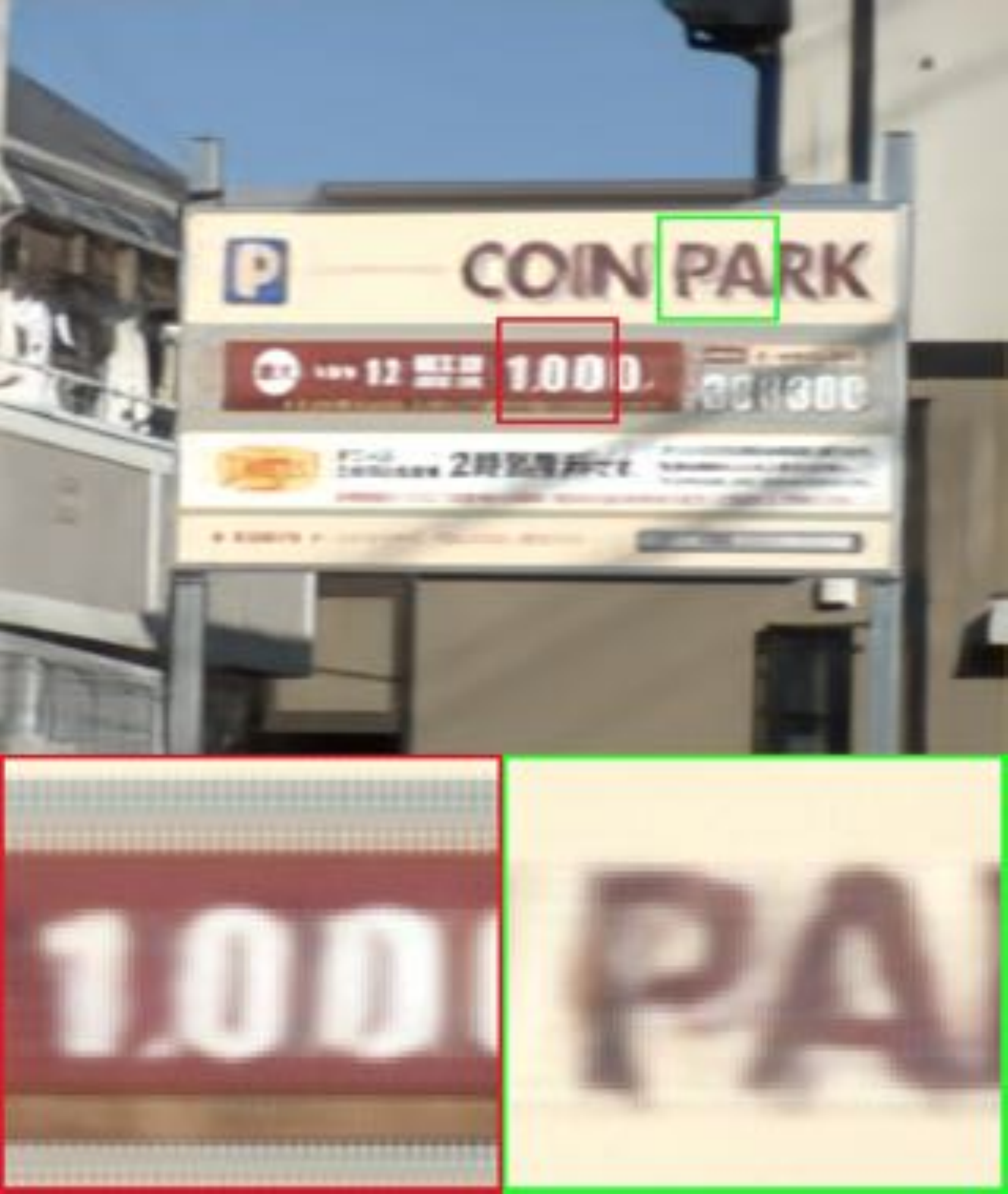} &
    		\includegraphics[width=0.16\linewidth]{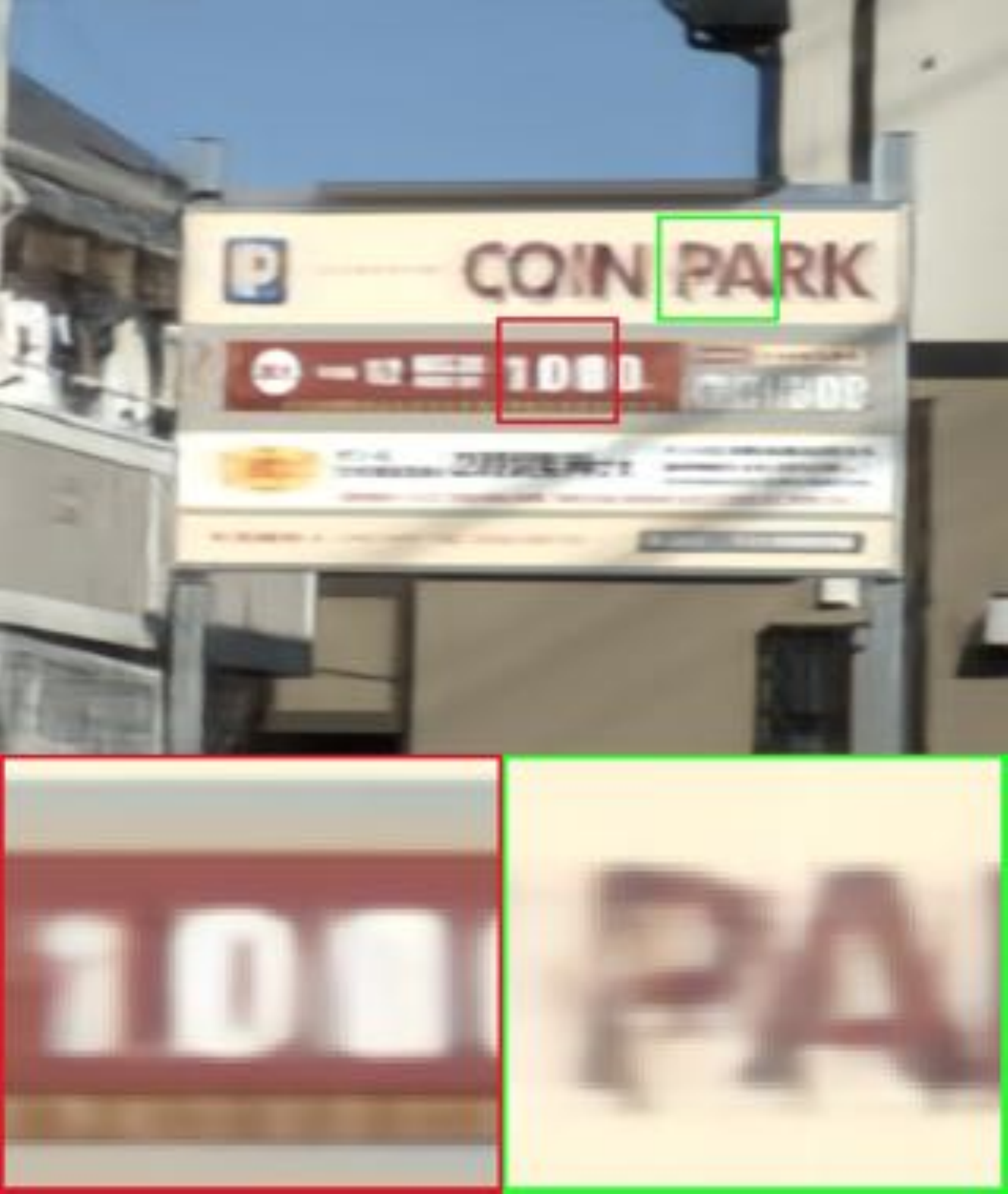} &
    		\includegraphics[width=0.16\linewidth]{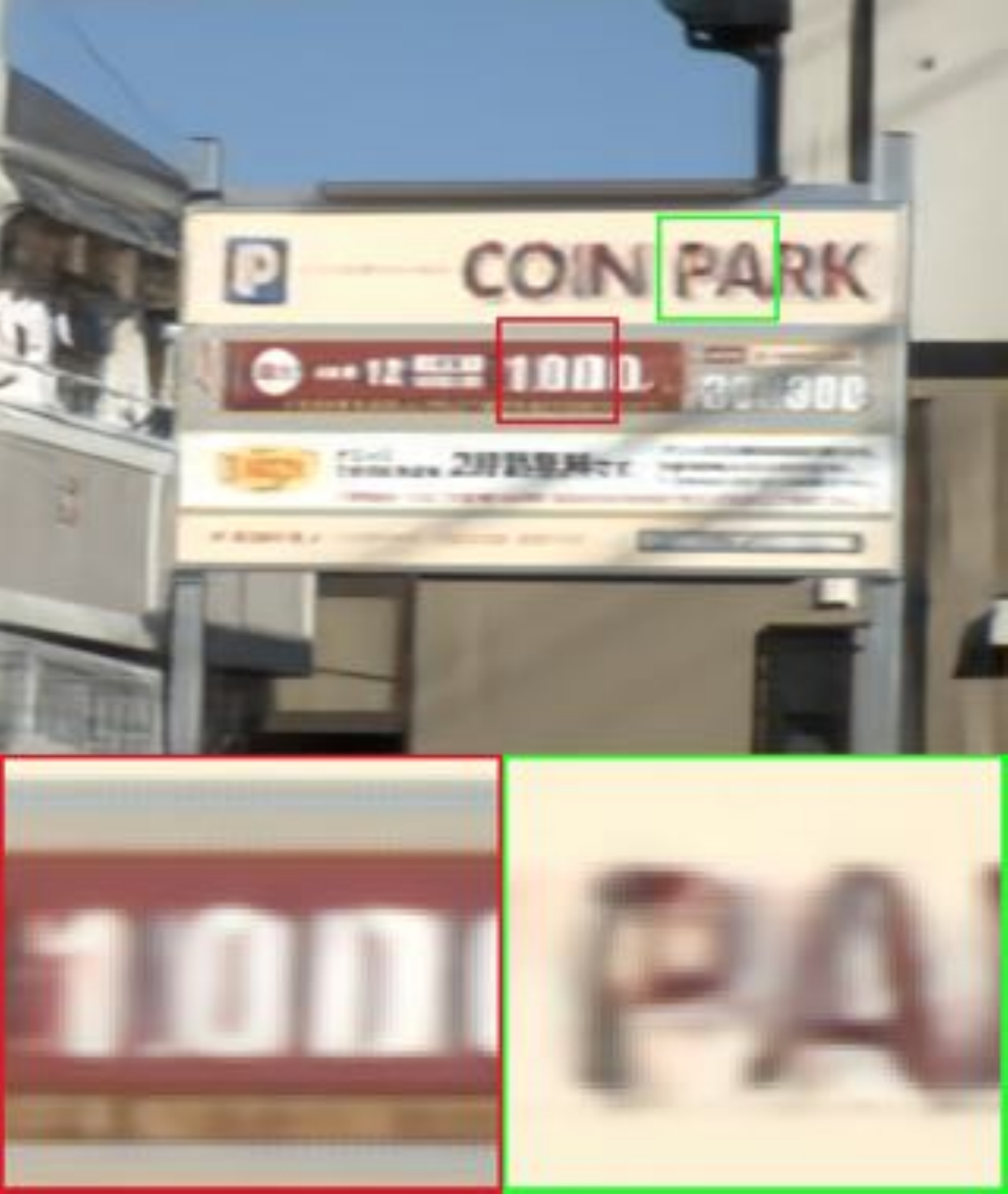}
    		&	
    		\includegraphics[width=0.16\linewidth]{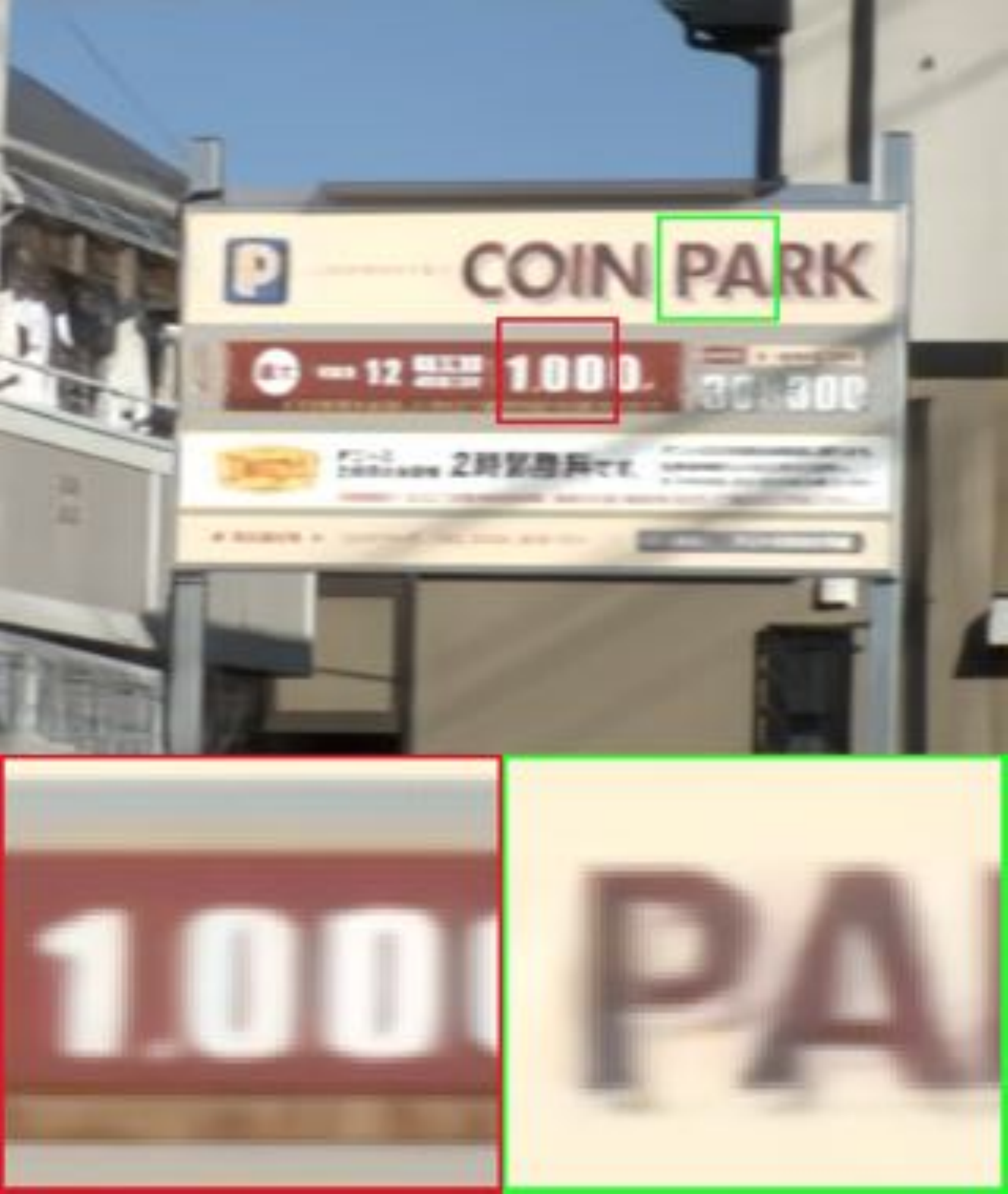} &
    		\includegraphics[width=0.16\linewidth]{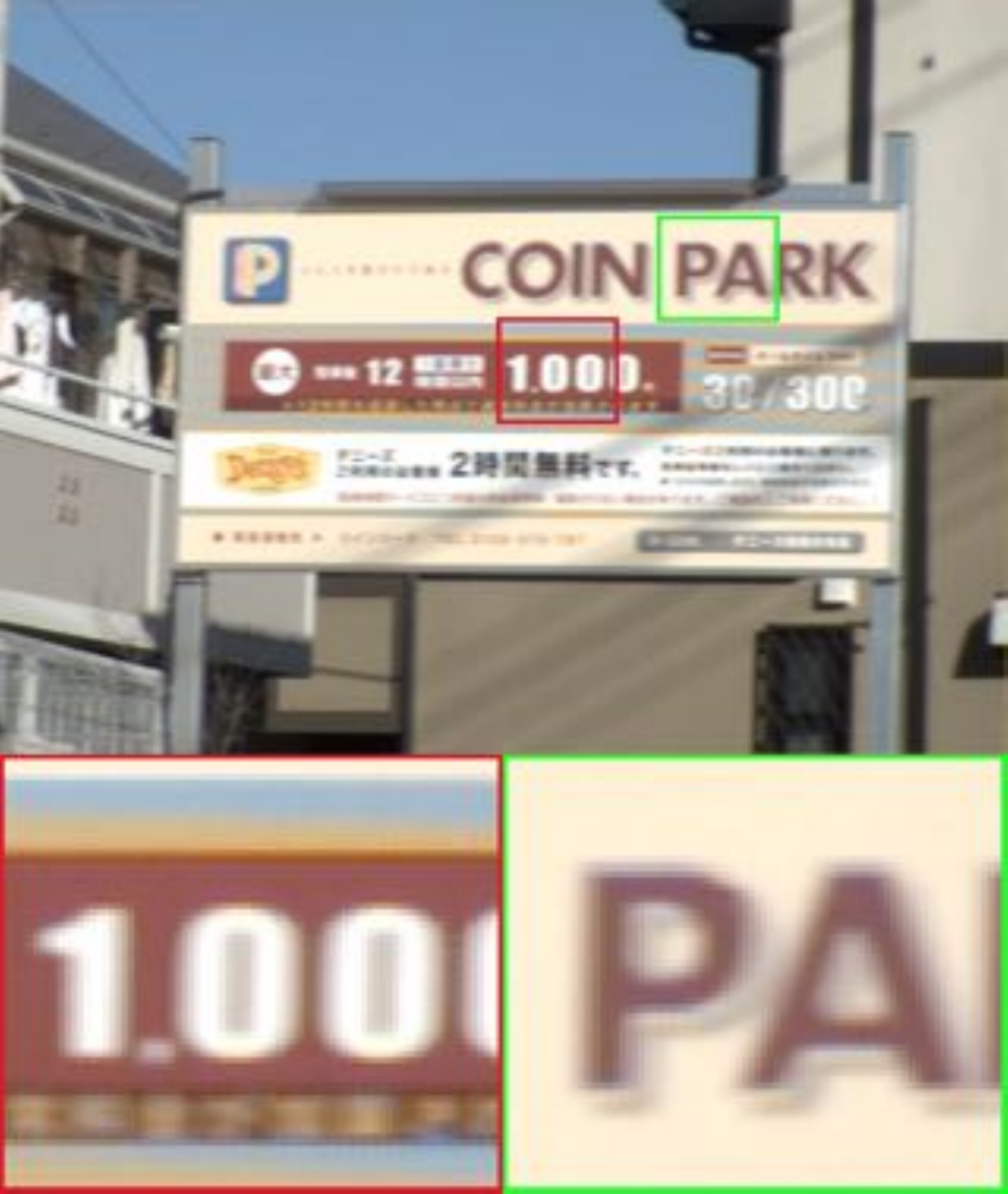} 
    		%
    	    \\
    		\includegraphics[width=0.16\linewidth]{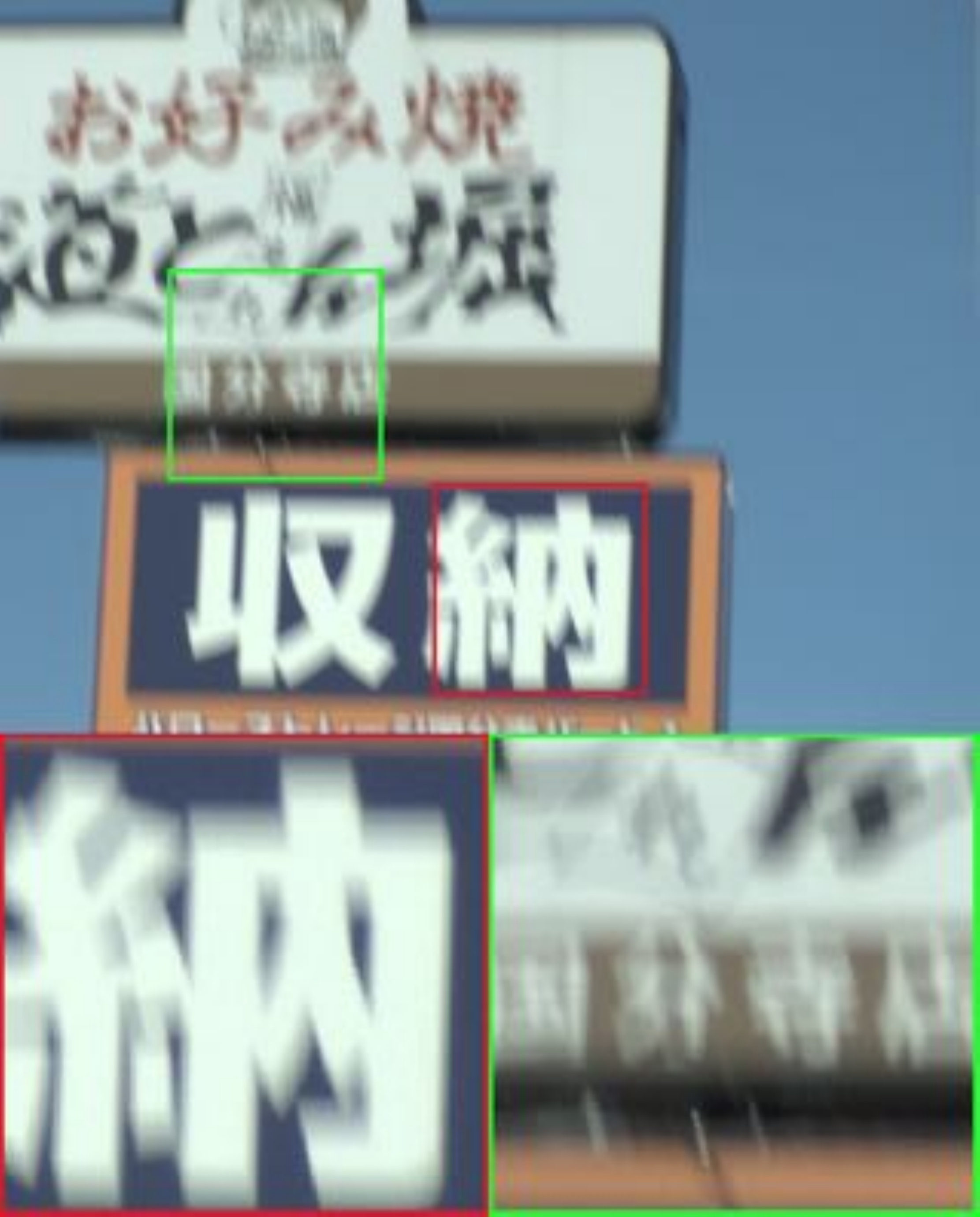} &
    		\includegraphics[width=0.16\linewidth]{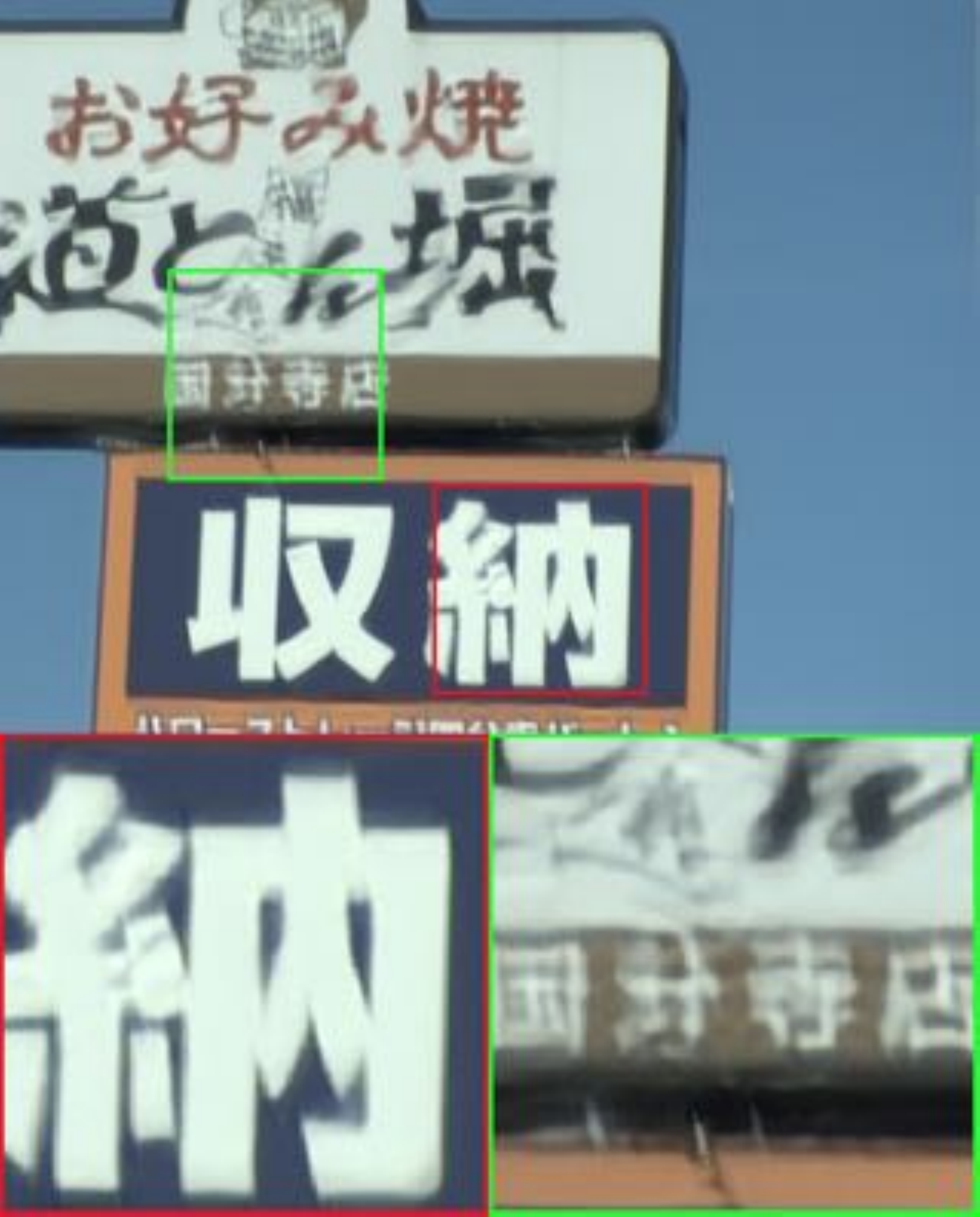} &
    		\includegraphics[width=0.16\linewidth]{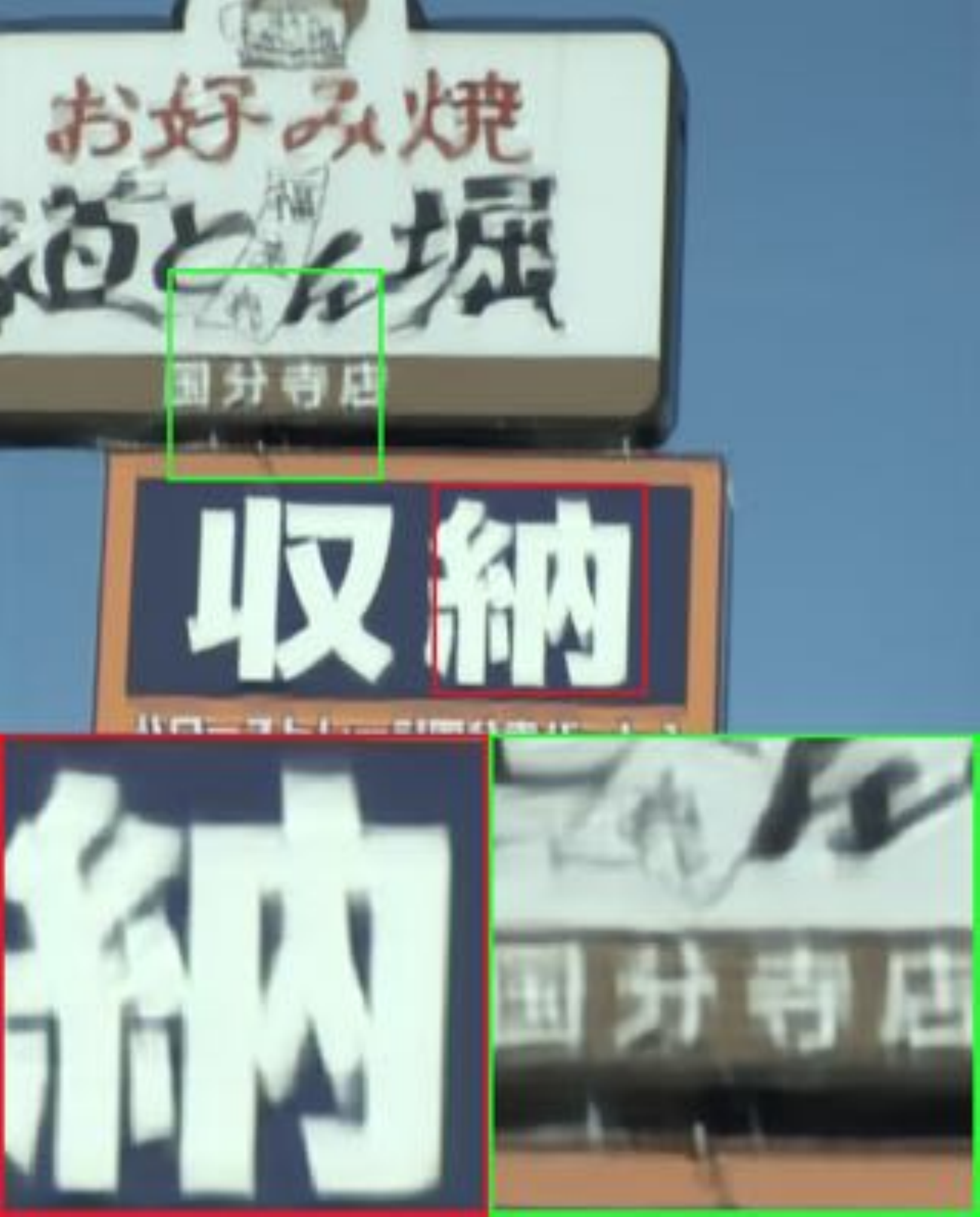} &
    		\includegraphics[width=0.16\linewidth]{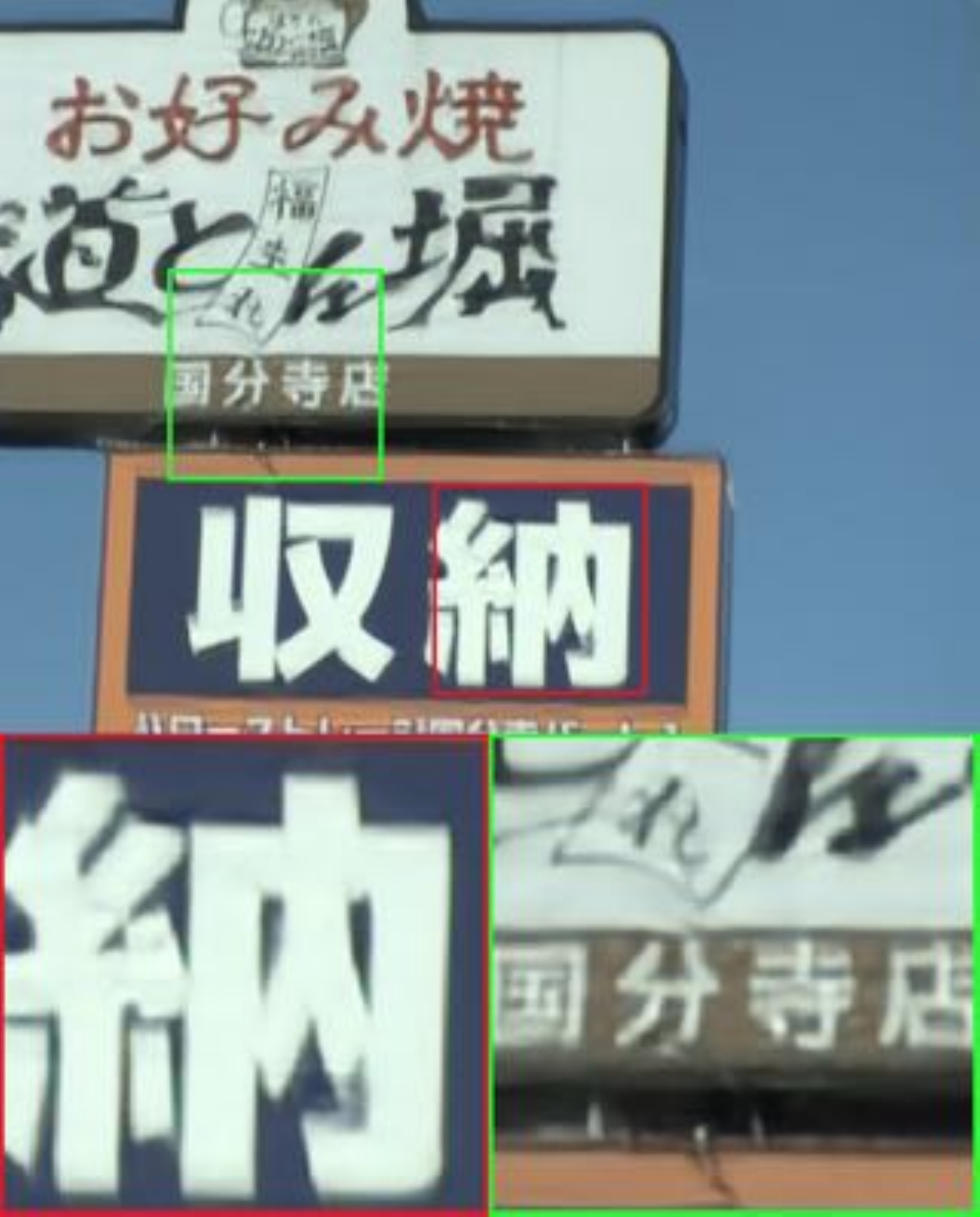} &	
    		\includegraphics[width=0.16\linewidth]{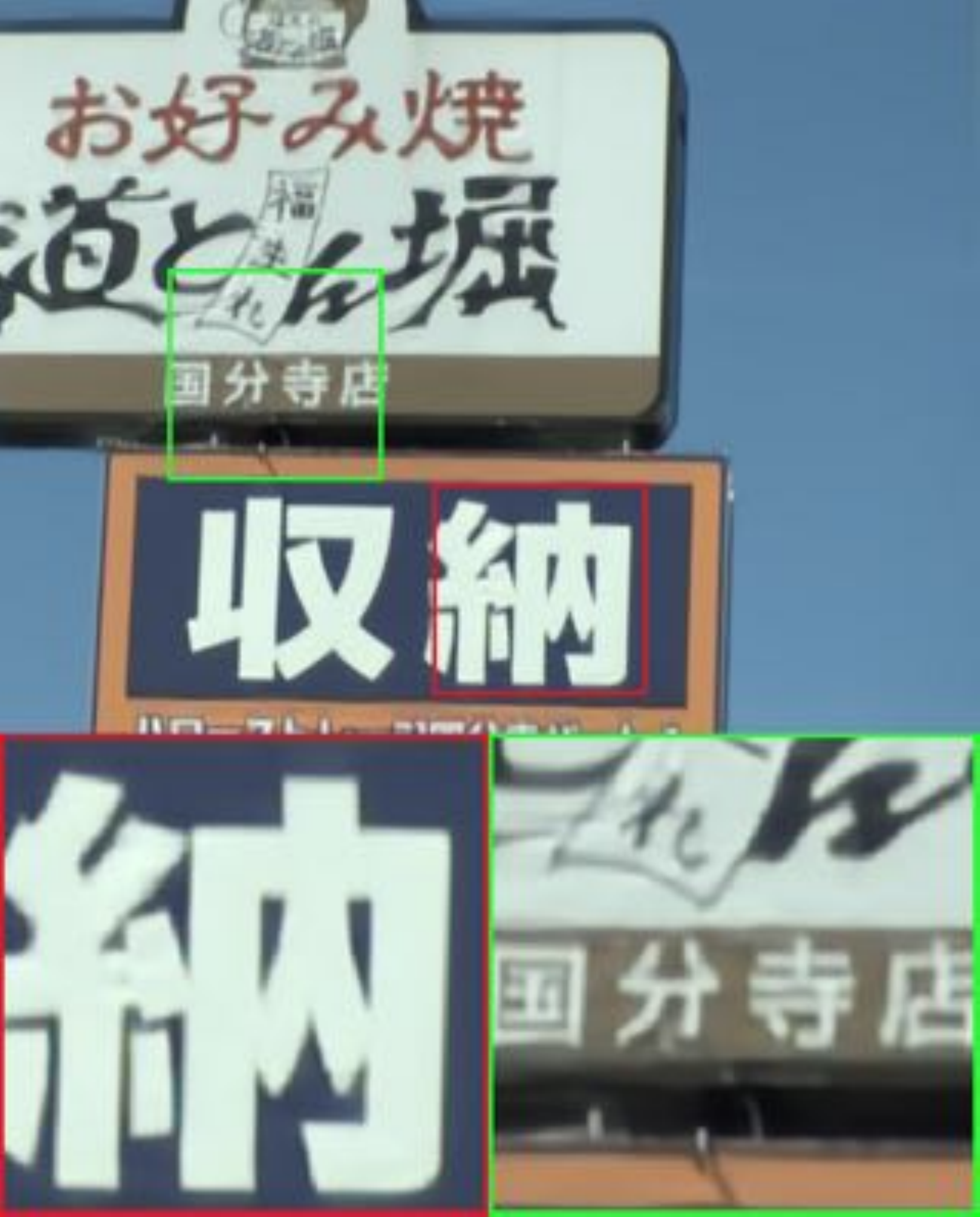} &
    		\includegraphics[width=0.16\linewidth]{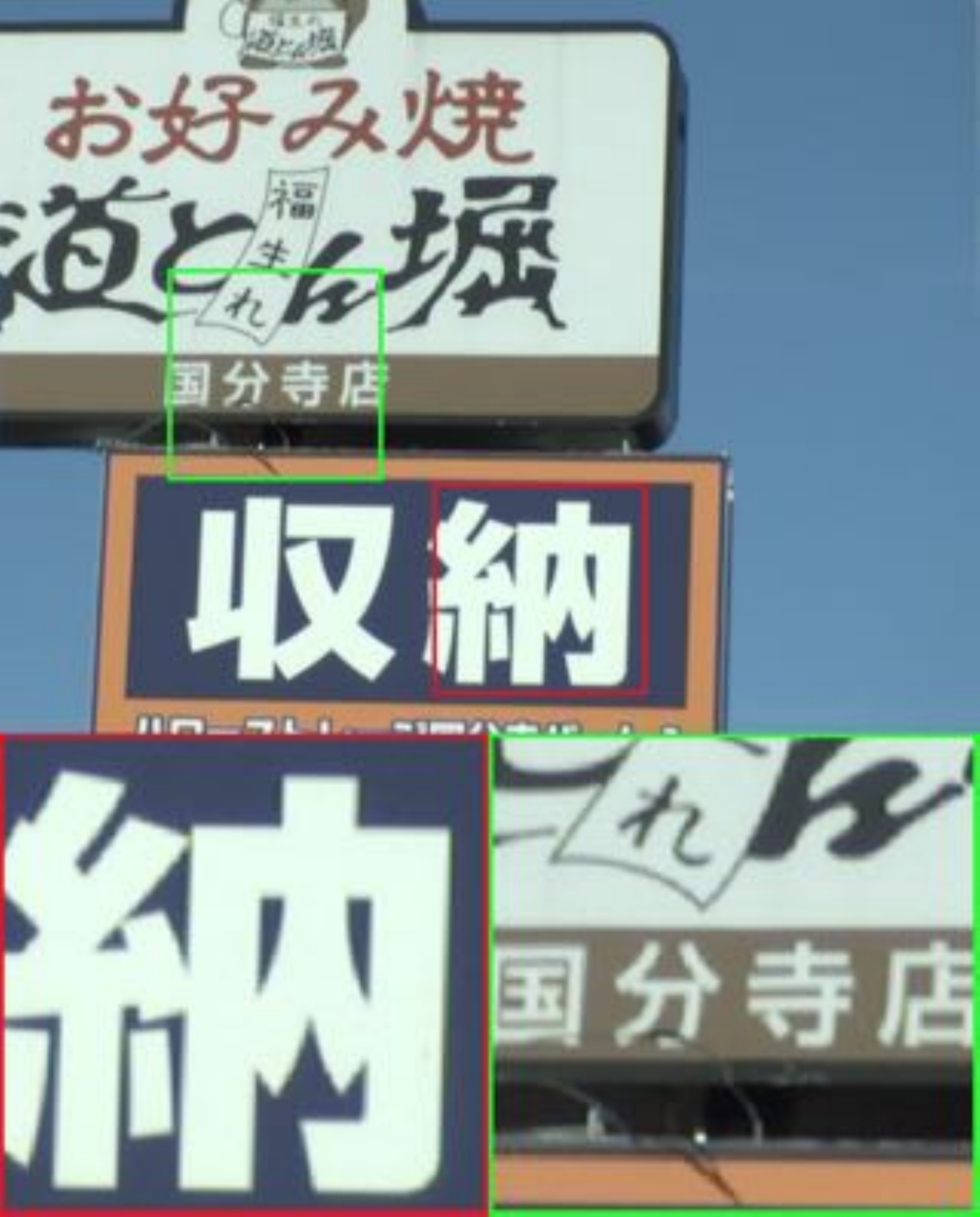}  \\		
    		\scriptsize (a) Input & \scriptsize (b) EDVR & \scriptsize (c) TSP & \scriptsize (d) PVDNet-L & \scriptsize (e) Ours & \scriptsize (f) GT\\
    	\end{tabular}
    }
    \caption{The qualitative results on the 2ms-16ms subset of the BSD dataset. Note that ``GT'' stands for ground truth.}
    \label{fig:show_bsd}
\end{figure*}

\noindent \textbf{The GoPro dataset.}
We compare STDANet to the state-of-the-art video deblurring methods on the GoPro dataset~\cite{DBLP:conf/cvpr/NahKL17}. 
As show in Table~\ref{tab:psnr_GoPro}, the proposed STDANet and STDANet-Stack perform favorably against the state-of-the-art methods in terms of PSNR and SSIM. 
Compared to the PVDNet-L~\cite{DBLP:journals/tog/SonLLCL21}, STDANet achieves higher PSNR and SSIM with lower computational complexity. 
STDANet-Stack achieves 0.95dB higher PSNR than TSP~\cite{DBLP:conf/cvpr/PanBT20} with lower computational complexity, where the STDANet-Stack use the same cascaded progressive structure as TSP~\cite{DBLP:conf/cvpr/PanBT20}. 
As shown in Figure~\ref{fig:show_gopro}, the proposed method restores better image details and structures. 

\noindent \textbf{The BSD dataset.}
We compared the our method to the state-of-the-art methods on BSD dataset~\cite{DBLP:conf/eccv/ZhongGZZ20}. 
For a fair comparison, the EDVR~\cite{DBLP:conf/cvpr/WangCYDL19}, TSP~\cite{DBLP:conf/cvpr/PanBT20}, and PVDNet-L~\cite{DBLP:journals/tog/SonLLCL21} are trained with their open-sourced implementations. 
In Table~\ref{table:bsd}, our method achieves the best results on all the three subsets in terms of PSNR and SSIM. 
The qualitative results are shown in Figure~\ref{fig:show_bsd}, which indicate that our method restores much sharper images. 

\begin{figure*}[!t]
\centering
\renewcommand{\tabcolsep}{0.5pt}
\renewcommand{\arraystretch}{1}
	\begin{tabular}{cccccc}
		\includegraphics[width=0.16\linewidth]{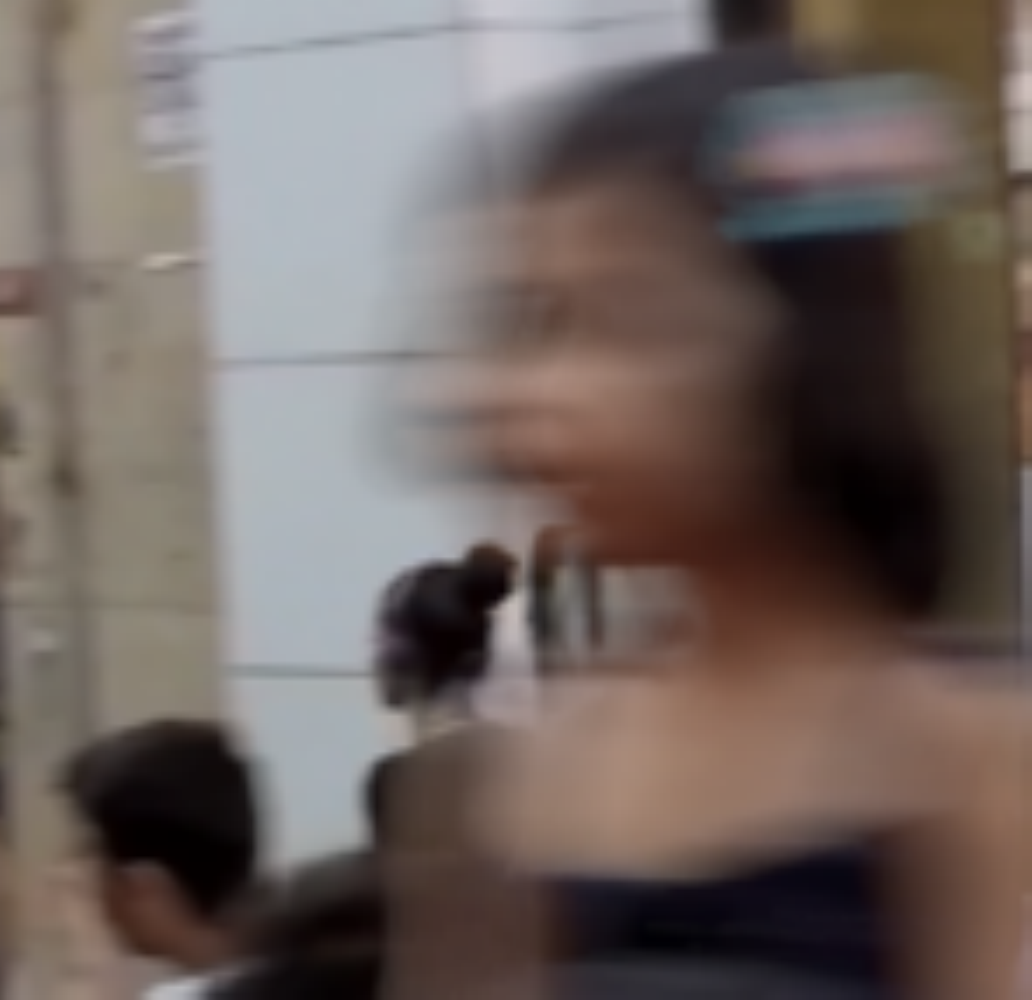}&
		\includegraphics[width=0.16\linewidth]{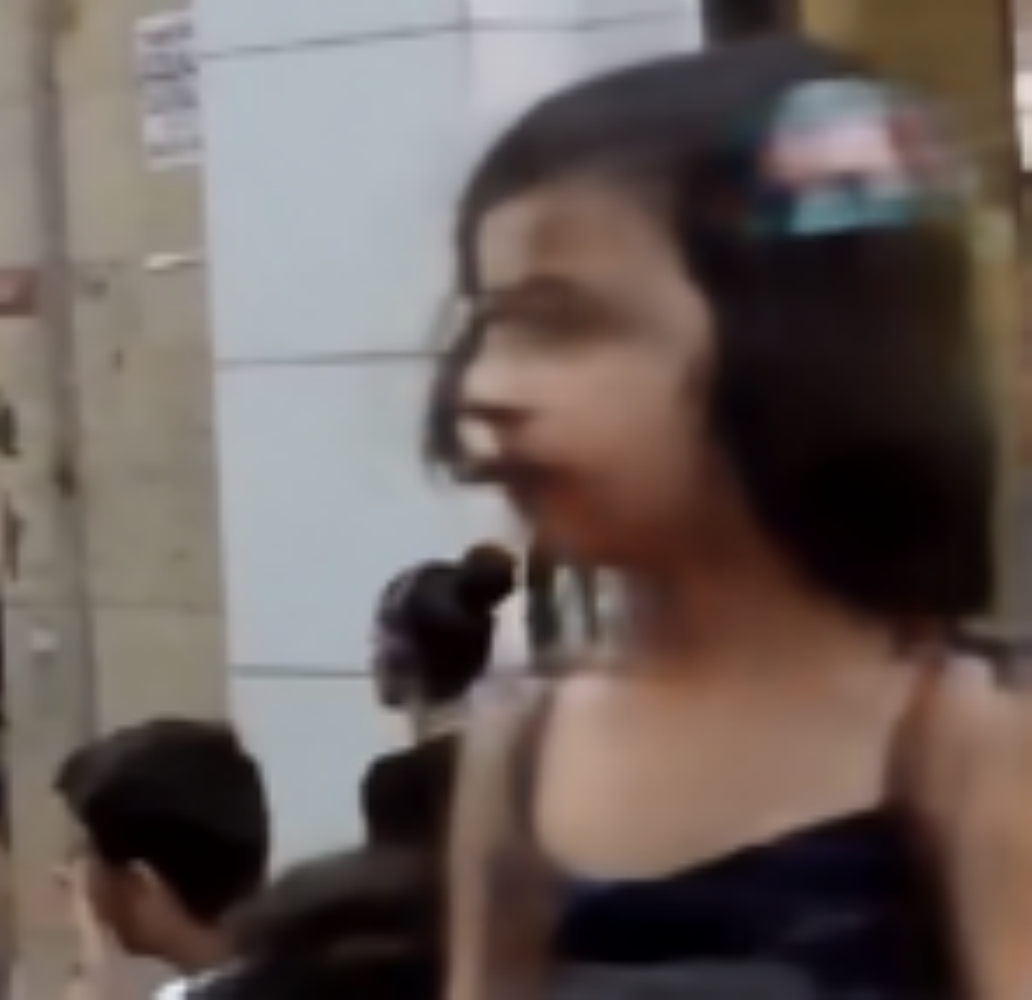} &
		\includegraphics[width=0.16\linewidth]{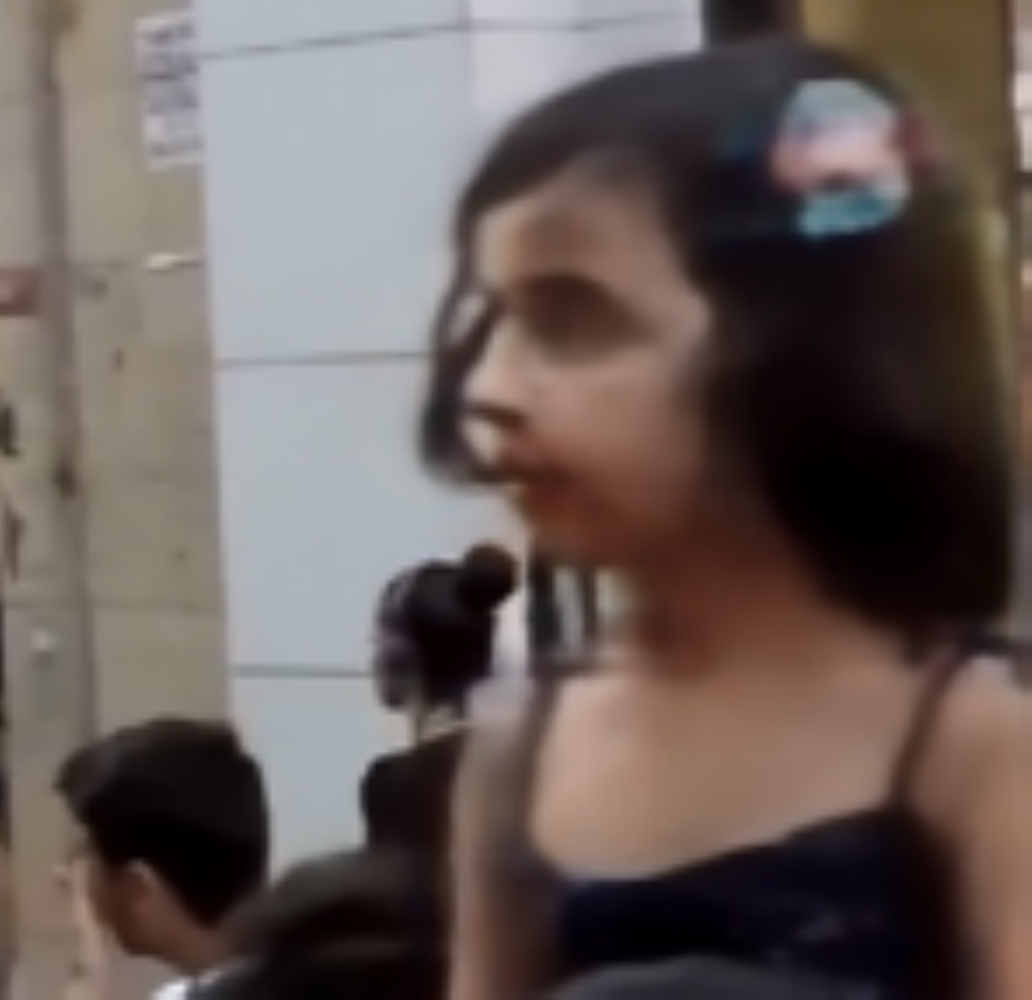} &
		
		\includegraphics[width=0.16\linewidth]{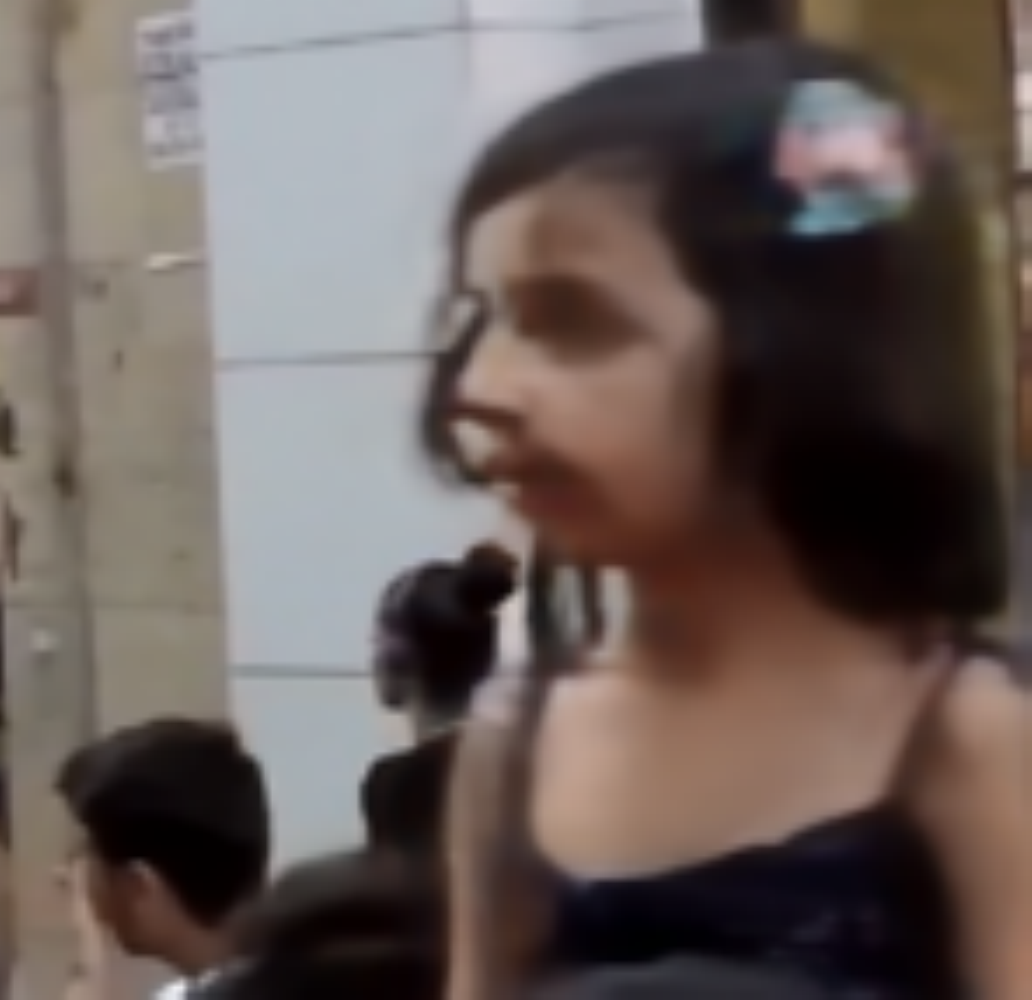}
		&	
		\includegraphics[width=0.16\linewidth]{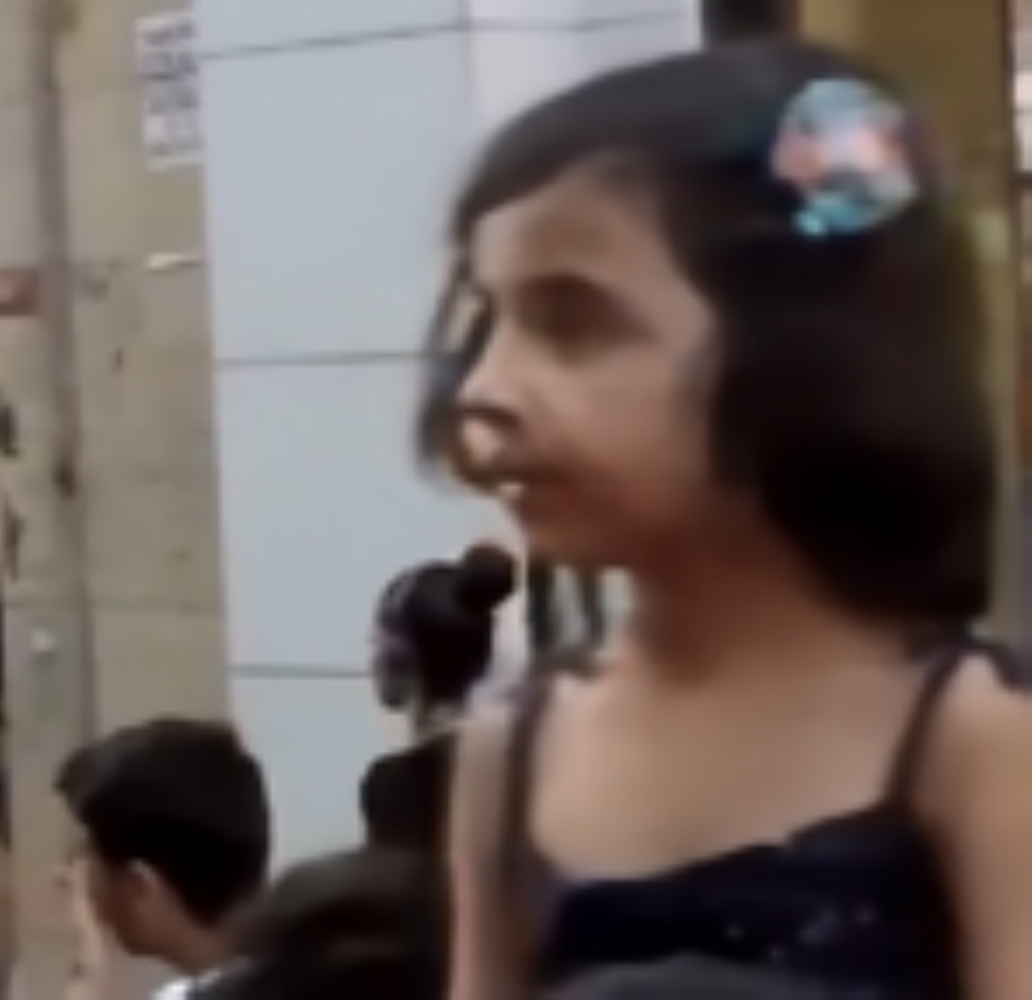}&
		\includegraphics[width=0.16\linewidth]{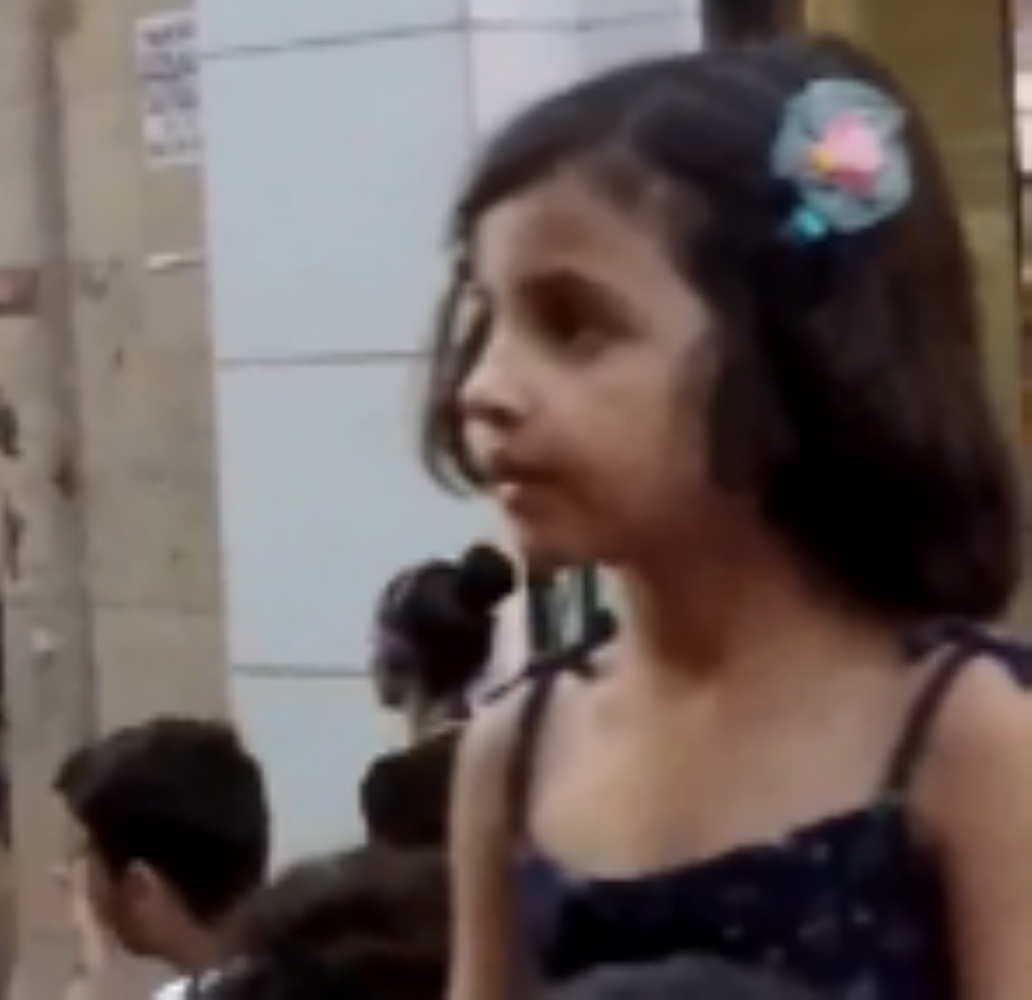} \\
	   \tiny (a) Input & \tiny(b) (-MMA,-MSA) & \tiny (c) (-MMA,+MSA)& \tiny (d) (+MMA,-MSA) & \tiny (e) (+MMA,+MSA) & \tiny(f) GT\\
	\end{tabular}
    \caption{The qualitative results when replacing MMA and MSA layers with the concatenation operation. Note that ``+'' and ``-'' denote ``with'' and ``without'', respectively. ``GT'' stands for ground truth.}
    \label{fig:show_viusalmmamsa}
\end{figure*}


\section{Analysis and Discussions}

\subsection{Effectiveness of the STDA Module}

\begin{figure*}[!t]
    \centering
    \includegraphics[width=0.8\textwidth]{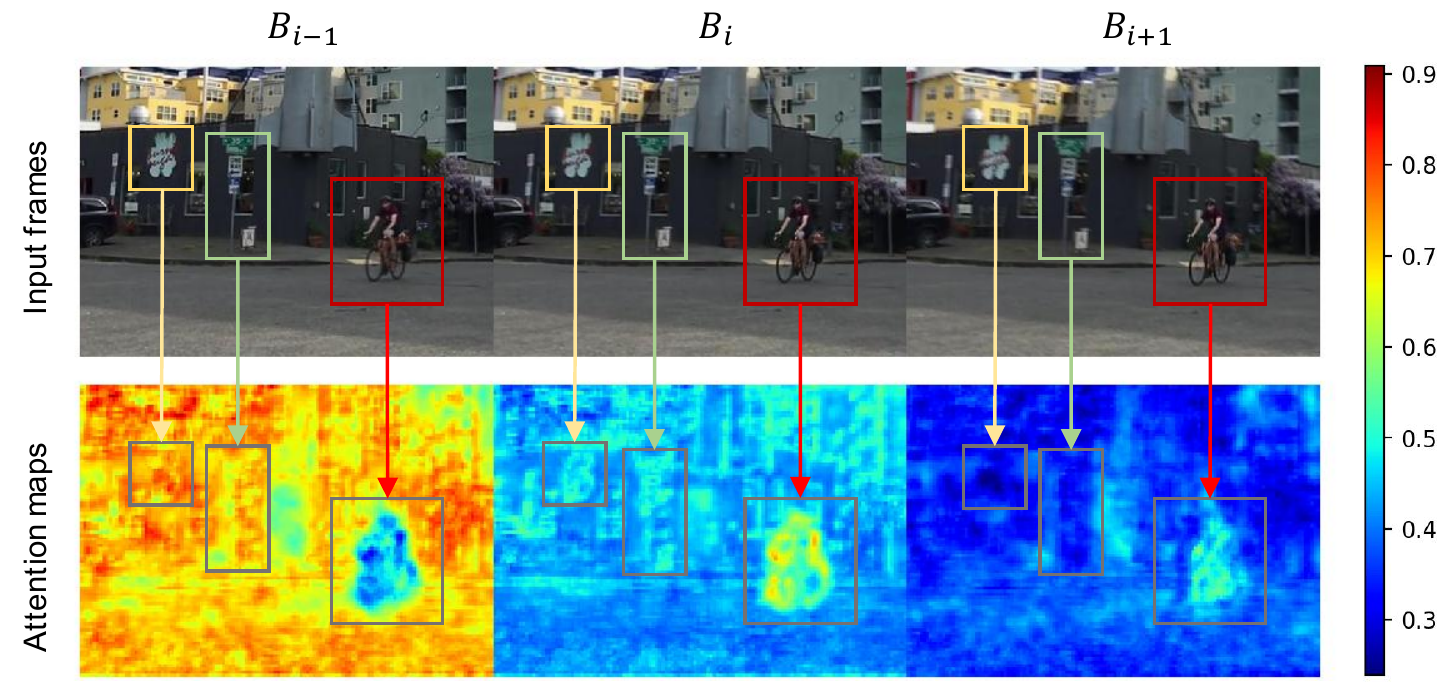}
    \caption{The visualization of the attention maps in the MSA layer. Sharper pixels have larger attention weights. \textbf{(zoom in for best view)}.}
    \label{fig:visual}
\end{figure*}

\noindent \textbf{MMA and MSA layers.} 
The STDA Module contains two main components: the MMA and MSA layers, which aggregates information of sharp pixels from adjacent frames.
To validate the effectiveness of the STDA module, the MMA and MSA layers are replaced with the concatenate operation.
In the concatenate operation, the information from all pixels are introduced from adjacent frames.
Table~\ref{tab:structure} shows the qualitative comparison when the MMA or MSA layer is removed.
Specially, when both MMA or MSA layers are removed, the estimated  optical flows are used to align the features from adjacent frames.
The experimental results shows that the networks perform worse without the help of the information of  sharp pixels extracted by the MMA and MSA layers.
Figure~\ref{fig:show_viusalmmamsa} shows the qualitative comparison on the GoPro dataset.
The network is less effective to restore sharp details when both MMA and MSA layers are removed.
Figure~\ref{fig:visual} gives the visualization of the attention maps in the MSA layer, which shows that sharper pixels have larger attention weights.
For example, the man riding a bicycle (highlighted with a red bounding box) is blurry in $B_{i - 1}$, and thus the corresponding regions are of low weights in the attention maps.
In contrast, $B_i$ have larger weights for this region.
To conclude, the proposed STDA module effectively aggregates the information of sharp pixels from adjacent frames. 

\begin{table}[!t]
	\centering
	\caption{The quantitative results on the GoPro dataset in terms of PSNR, SSIM, and GMACs with different numbers of sampling points.}
	\begin{tabularx}{\linewidth}{lYYYY}
		\toprule
		\#Sampling Points    & $K=$ 1             & $K=$ 8             & $K=$ 12       & $K=$ 16\\
		\midrule
		PSNR           & 31.64              & 32.12       & 32.29            & \bf{32.32}  \\
		SSIM           & 0.9183             & 0.9288      & 0.9313           & \bf{0.9319}  \\
		\midrule
		GMACs & \bf{1520}         &  1620        & 1677        &1735\\
		\bottomrule
	\end{tabularx}
	\label{tab:sample_points}
\end{table}

\noindent \textbf{Sampling Points.}
To investigate the effect of numbers of sampling points in the STDA module, we compare the performance with different numbers of sampling points.
As shown in Table~\ref{tab:sample_points}, larger number of sampling points leads to better restoration results but also heavier computational cost.
Specially, the STDA Module degenerates to the temporal attention when $K = 1$, which causes severe degeneration in restoration results. 
The PSNR only increases $0.03$ dB when increasing the number of sampling points from $12$ to $16$.
Therefore, we set $K = 12$ due to the trade-off between the computational cost and restoration performance. 

\noindent \textbf{Attention Heads.} 
The number of attention heads is one of the important hyperparameter in the deformable attention function.
We also compare the effect with different numbers of attention heads in Table~\ref{tab:heads}.
As the number of attention heads increases, the PSNR, SSIM, and GMACs increase.
Considering the trade-off between the computational complexity and restoration performance, we choose the number of attention heads $M = 4$.

\begin{table}[!t]
	\caption{The quantitative results on the GoPro dataset in terms of PSNR, SSIM, and GMACs with different numbers of attention heads.}
	\label{tab:heads}
	\begin{tabularx}{\linewidth}{lYYY}
		\toprule
		\#Attention Heads    & $M=$ 1             & $M=$ 4             & $M=$ 8       \\
		\midrule
		PSNR          & 32.13              & 32.29       & \bf{32.34}            \\
		SSIM           & 0.9294              & 0.9313        & \bf{0.9322}            \\
		\midrule
		GMACs & \bf{1548}         &  1677        & 1849        \\
		\bottomrule
	\end{tabularx}
\end{table} 

\begin{table}[!t]
	\centering
	\caption{The quantitative results on the GoPro dataset in terms of PSNR, SSIM, and GMACs with different optical flow estimators.}
	\begin{tabularx}{\linewidth}{lYYc}
		\toprule
		Estimator  & None & PWC-Net & Motion Estimator \\
		\midrule
		PSNR          & 31.58              & \bf{32.36}      & 32.29         \\
		SSIM           & 0.9176              & \bf{0.9326}       & 0.9313         
		\\
		\midrule
		GMACs          & \bf{1632}              & 2352       & 1677           \\
		\bottomrule
	\end{tabularx}
	\label{tab:flow}
\end{table}

\subsection{Effectiveness of the Motion Estimator} 

To evaluate the effectiveness of the motion estimator, we compare the video deblur results with different optical flow estimators.
As shown in Table~\ref{tab:flow}, removing the optical flow estimator causes considerable degeneration.
Although STDANet with PWC-Net~\cite{DBLP:conf/cvpr/SunY0K18} archives the best results, it also leads to high computational cost.
STDANet with the proposed motion estimator archives the best trade-off between the deblur results and computational complexity.

\section{Conclusions}

In this paper, we propose STDANet for video deblurring.
The main motivation of this work is that not all the pixels in the video frames are sharp and beneficial for deblurring.
Therefore, the proposed STDANet extracts the information of sharp pixels by considering the pixel-wise blur levels of the video frames.
Different from mainstream video debulr methods that requires accurate optical flows to align two adjacent frames to the mid-frame, 
the coarse optical flows are estimated by a lightweight motion estimator and are used as the base offsets to find the corresponding sharp pixels in the adjacent frames.
Experimental results indicate that the proposed STDANet performs favorably against state-of-the-art methods on the GoPro, DVD, and BSD datasets. 

\noindent \textbf{Acknowledgement.}
This work is supported by the National Key R$\&$D Program of China (No. 2021ZD0110901). 

\clearpage
%
%
\bibliographystyle{splncs04}
\bibliography{egbib}

\end{document}